\documentclass[11pt]{style/hoh-arxiv}

\makeatletter
\newcommand{\longdash}[1][2em]{%
  \makebox[#1]{$\m@th\smash-\mkern-7mu\cleaders\hbox{$\mkern-2mu\smash-\mkern-2mu$}\hfill\mkern-7mu\smash-$}}
\makeatother
\newcommand{\omitskip}{\kern-\arraycolsep}

\usepackage{algorithm}
\usepackage{algorithmic}
\usepackage{fontawesome5}
\usepackage{wrapfig}
\usepackage{needspace}

\newcolumntype{L}[1]{>{\raggedright\arraybackslash}p{#1}}
\colorlet{headerrow}{accent_color!10}
\colorlet{finalcol}{accent_color!8}
\definecolor{actionrow}{HTML}{EAF2FB}
\definecolor{timingrow}{HTML}{F8F0DE}
\definecolor{strategyrow}{HTML}{F1ECF8}
\definecolor{simulationrow}{HTML}{E8F3EC}
\definecolor{adventurerow}{HTML}{F8EDEF}
\colorlet{implementationrow}{actionrow}
\colorlet{performancerow}{simulationrow}
\colorlet{researchrow}{strategyrow}
\newtcolorbox{plannerprompt}[1]{
  enhanced, breakable,
  colback=hohBlue!5, colframe=hohBlue!70!black,
  coltitle=white, title={#1}, title after break={#1 (continued)},
  fonttitle=\sffamily\bfseries, boxrule=0.7pt, arc=2pt,
  left=7pt, right=7pt, top=6pt, bottom=6pt, before upper=\raggedright
}
\newtcolorbox{developerprompt}[1]{
  enhanced, breakable,
  colback=hohTeal!5, colframe=hohTeal!75!black,
  coltitle=white, title={#1}, title after break={#1 (continued)},
  fonttitle=\sffamily\bfseries, boxrule=0.7pt, arc=2pt,
  left=7pt, right=7pt, top=6pt, bottom=6pt, before upper=\raggedright
}
\newtcolorbox{testprompt}[1]{
  enhanced, breakable,
  colback=hohAmber!6, colframe=hohAmber!75!black,
  coltitle=white, title={#1}, title after break={#1 (continued)},
  fonttitle=\sffamily\bfseries, boxrule=0.7pt, arc=2pt,
  left=7pt, right=7pt, top=6pt, bottom=6pt, before upper=\raggedright
}
\newtcolorbox{controlprompt}[1]{
  enhanced, breakable,
  colback=hohRose!5, colframe=hohRose!70!black,
  coltitle=white, title={#1}, fonttitle=\sffamily\bfseries,
  boxrule=0.7pt, arc=2pt, left=7pt, right=7pt, top=6pt, bottom=6pt
}
\newcommand{\promptslot}[1]{%
  \textcolor{hohRose!85!black}{\texttt{\{\{#1\}\}}}}
\newcommand{\promptpath}[1]{%
  \textcolor{hohTeal!75!black}{\texttt{#1}}
}
\newcommand{\promptmodule}[1]{%
  \textcolor{hohAmber!80!black}{\texttt{[#1]}}
}
\floatstyle{ruled}
\newfloat{listing}{tb}{lst}{}
\floatname{listing}{Listing}
\hohsettheme{hohRose}

\title{Harness-of-Harness: Multi-Day Autonomous Software Development with Continual Improvement}
\author{Haoyang Yan\textsuperscript{\textdagger}}
\author{Min-Le Su\textsuperscript{\textdagger}}
\author{Hangfan Zhang\textsuperscript{\textdagger}}
\author{Zhanhao Li\textsuperscript{\textdagger}}
\author{Chen Zhang}
\author{Shao Zhang}
\author{Yang Chen}
\author{Lei Bai}
\author{Shuyue Hu}
\affiliation{Shanghai Artificial Intelligence Laboratory}

\abstract{
\small
This paper studies autonomous software development, in which LLM-based coding agents transform high-level requirements into complete, functional, and usable software systems without human intervention. We introduce Harness-of-Harness (HoH), a framework that enables coding agents to continually improve software during autonomous development.
HoH operates on existing coding-agent harnesses, and organizes their executions into iterative planning–coding–testing loops. To sustain improvement across loops, HoH balances repair with capability growth, scopes development into small and verifiable increments, separates implementation-time testing from independent evaluation, and constrains verifiable outputs rather than prescribing agent workflows.
It progressively exposes deliverables, role-specific tools, and skills, encourages reuse rather than recreation, and maintains versioned project histories. 
On GameCraft-Bench, FrontierSWE, and ProgramBench, three harness–model pairs (Codex with GPT-5.5, OpenCode with DeepSeek-V4-Pro, and Pi with MiniMax-M3), HoH consistently outperforms the corresponding standalone harnesses, achieving an average relative gain of 52.25\% and a maximum gain of 82.86\% after three iterations. In a multi-day deployment with more than 70 iterations, HoH autonomously develops a first-person-shooter game, featuring a coherent storyline, fully implemented core mechanics, human-playable experience, polished visuals and integrated audio.}

\newcommand{\projectpagelink}{}
\renewcommand{\website}[2]{%
  \def\projectpagelink{{\textcolor{accent_color}{\faGlobe}\enspace
    \textbf{Project Page:} \href{#1}{\color{accent_color}#2}}}%
}
\renewcommand{\code}[2]{%
  \def\codelist{{\small \sffamily
    \textcolor{accent_color}{\faGithub}\enspace\textbf{GitHub:}
    \href{#1}{\color{accent_color}#2}\hspace{1.5em}\projectpagelink}}%
  \def\websitelist{}%
}
\website{https://flesymeb.github.io/HarnessOfHarness/}{HarnessOfHarness}
\code{https://github.com/Flesymeb/HarnessOfHarness}{Flesymeb/HarnessOfHarness}

\providecommand{\keywordslist}{}
\providecommand{\websitelist}{}
\providecommand{\codelist}{}

\begin{document}

\enlargethispage{1.5cm}
\maketitle
\begingroup
\renewcommand{\thefootnote}{\textdagger}
% Match the dagger markers attached to the first four authors while giving
% hyperref a concrete footnote number for its PDF anchor.
\footnotetext[1]{Equal contribution. \{yanhaoyang, suminle, zhanghangfan, lizhanhao, hushuyue\}@pjlab.org.cn}
\endgroup

\vspace{-0.2cm}
\begin{figure}[H]
    \centering
    \includegraphics[width=\textwidth]{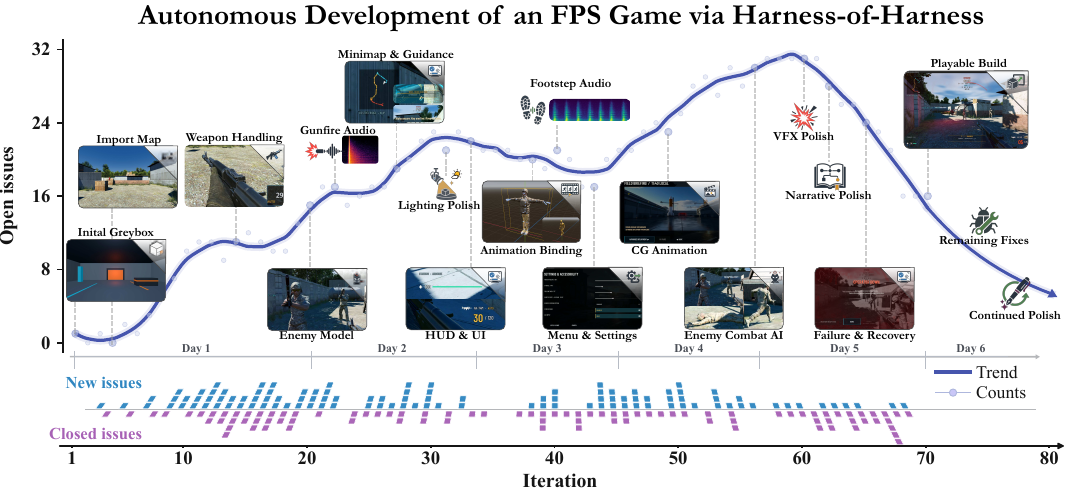}
    \captionsetup{font=small}
    \caption{Across successive iterations by Harness-of-Harness, the resulting First-Person-Shooter game features a coherent storyline, implemented combat, weapon and enemy-interaction systems, player guidance, heads-up display and menu systems, cinematic animation, and polished visual and audio presentation, yielding human-playable experience. The game, development traces, and gameplay videos are available on GitHub.}
    \label{fig:fusepoint-development-trajectory}
\end{figure}
\vspace{-0.2cm}

\section{Introduction}
Software development has become a prominent application of large language models (LLMs)~\citep{chen2021codex,wang2023codet5plus}. As LLM capabilities have advanced, LLM-based coding agents have progressed from localized assistance, such as function completion~\citep{lu2021codexglue,barke2023groundedcopilot}, to increasingly complex tasks, including navigating large codebases and resolving repository-level issues~\citep{jimenez2024swebench,yang2024sweagent,wang2025openhands,oueslati2026refagent,merrill2026terminal}. Despite their growing adoption, most coding agents still largely operate under a \textit{human-in-the-loop} setting (Figure 2a): developers must define tasks, guide intermediate decisions, review generated changes and intervene when failures occur~\citep{anthropic2025claudecodeproduct}. In this study, we pursue a more ambitious goal: \textit{autonomous software development}~(Figure 2b); given only high-level requirements as human input, coding agents start from scratch and independently transform the requirements into complete, functional, and deployable software systems, without further human guidance or intervention.

Such autonomous development poses a fundamentally longer-horizon problem than conventional agentic coding tasks~\citep{le2025sweevo,orlanski2026slopcodebench}. Building a software system from scratch requires agents not only to generate code snippets, but also to translate high-level requirements into executable plans, coordinate interdependent tasks, design and integrate components, and continuously test and debug the evolving system~\citep{hong2024metagpt,qian2024chatdev,wang2025openhands}. As these interdependent decisions and modifications accumulate, development naturally unfolds over increasingly long trajectories~\citep{le2025sweevo,orlanski2026slopcodebench}. As trajectories grow, agents may lose track of earlier requirements and design decisions, or introduce local fixes that violate constraints elsewhere~\citep{cemri2025multiagentfailures,mundler2024swtbench}. Failed attempts and suboptimal decisions may accumulate, while new evidence from testing can invalidate earlier assumptions~\citep{shinn2023reflexion,madaan2023selfrefine,chen2026failedtrajectoriesreliablellm}. Long trajectories can also lead to repetitive cycles of inspection and repair,  redundant verification of completed components, or premature declaration of completion despite missing or incorrect functionality~\citep{cemri2025multiagentfailures,huang2026prooforstop}. Together, these challenges suggest that autonomous software development is not simply a problem of longer execution; the real challenge is sustaining coherent and effective progress over time.

Here, we introduce Harness-of-Harness (HoH), a framework that equips coding agents with \emph{continual improvement} capabilities for autonomous software development. Modern coding agents operate within a harness---the surrounding system that provides tools, manages execution and mediates the LLM’s interaction with the development environment~\citep{yang2024sweagent,wang2025openhands,zhang2026selfharness}. HoH builds upon existing harnesses and organizes development into iterative planning–coding–testing loops. At each iteration, the planner synthesizes the high-level requirements and evidence from previous iterations into a development plan. Each plan must both address outstanding problems and deliver a small yet concrete new capability, following the principle of iterative and incremental development~\cite{larman2003iterative}. This helps prevent development from collapsing into repetitive local repairs, while the limited scope makes progress easier to verify and reduces the risk of uncontrolled changes. The developer then implements the plan and embeds focused testing throughout implementation, creating immediate feedback around local changes. After passing these tests, a tester independently evaluates the resulting system against both the overall requirements and the development plan, using complementary white-box and black-box tests. The tests are conducted from multiple perspectives, such as functional correctness, completeness, usability, and visual and audio quality (if any). The resulting structured test report is returned to the planner as evidence for the next iteration, closing the loop.

Throughout this process, HoH specifies the artifacts and evidence that agents must deliver, but does not prescribe a rigid workflow for producing them. Each role must return a structured artifact, and outputs that violate the required schema trigger a retry. This constrains verifiable outcomes while preserving agent autonomy over reasoning, tool use and implementation strategy. To maintain continuity without overwhelming the context window, HoH adopts progressive disclosure rather than a dedicated memory module: plans, reports, histories and other artifacts are persisted in the file system and initially exposed through a concise, categorized index, with detailed contents retrieved only when relevant. 
Tools, such as MCP servers, expert models and domain-specific algorithms, are organized by role, with lightweight Markdown-based skills providing concise, on-demand guidance for their use. Agents are encouraged to draw on existing resources rather than recreate standard capabilities, reducing redundant effort on routine engineering tasks.
Finally, HoH maintains a versioned record of project evolution at both the agent role and iteration levels. By preserving the software state together with concise accounts of how it changes, HoH can return to previously verified states after major regressions and draw on evidence from earlier attempts when similar failures recur to inform the diagnosis and resolution.

We evaluate HoH in two complementary settings: three controlled benchmarks (GameCraft-Bench~\cite{luo2026gamecraftbench}, FrontierSWE~\cite{proximal2026frontierswe}, and ProgramBench~\cite{yang2026programbench}), and open-ended game development that spans over multiple days. First, we evaluate the HoH loop under the original benchmark specifications, without additional tools, skills or version-control mechanisms. We consider three harness–model configurations: Codex with GPT-5.5 (high), OpenCode with DeepSeek-V4-Pro, and Pi with MiniMax-M3. Across all three benchmarks, HoH consistently outperforms the corresponding standalone harnesses. After three iterations, it yields absolute gains of 16.62–22.08 points on GameCraft-Bench, 19–29 points on FrontierSWE, and 6.09–16.85 points on ProgramBench. On FrontierSWE, HoH with Codex and GPT-5.5 (high) continues improving over ten iterations, from 22\% to 72.67\%. 
In our second setting, HoH autonomously builds a complex game from scratch, given only high-level product requirements, which exposes challenges that are largely absent from conventional benchmarks. Different from benchmark evaluation, we additionally implement HoH with role-specific tools and skills, supporting development engine interaction, asset acquisition and generation, reference retrieval, testing, and project-state management. Code changes and role-specific artifacts are committed to a public GitHub repository after each agent stage, making the complete development trajectory traceable. Over multiple days of autonomous development, HoH transforms the initial requirements into a complete, human-playable game with a coherent storyline, fully implemented core mechanics, polished visuals and integrated audio.

\begin{figure}[t]
    \centering
    \includegraphics[width=0.9\textwidth]{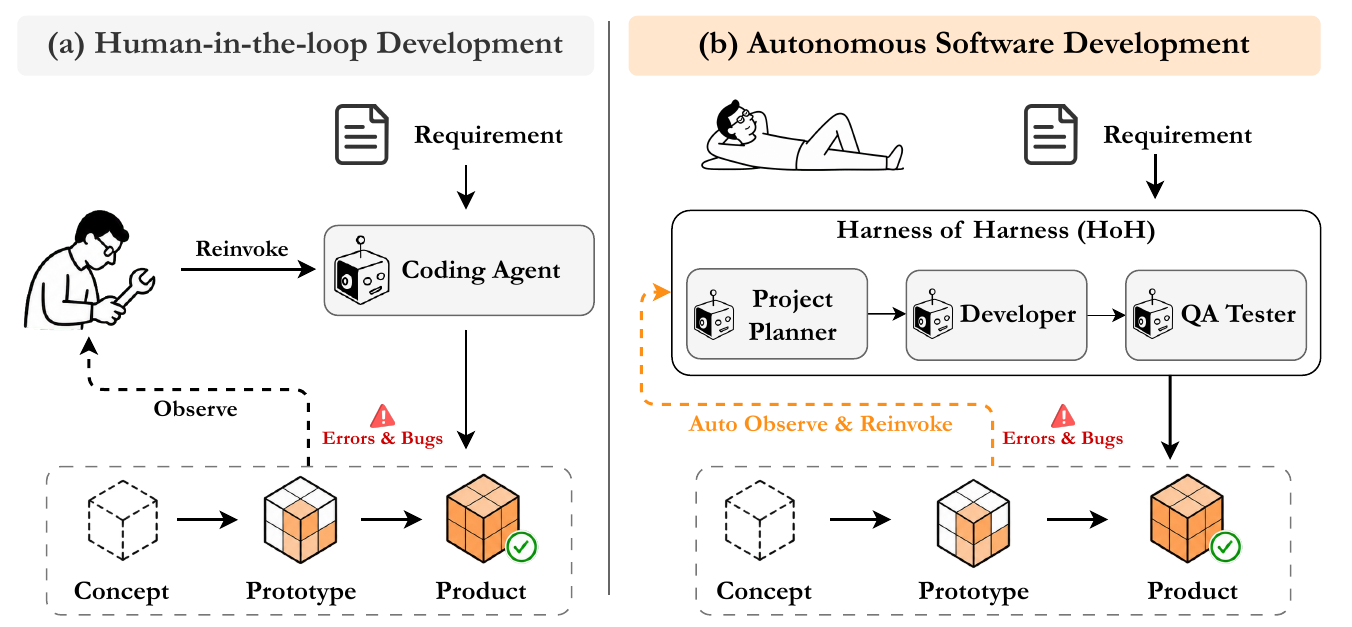}
    \caption{\textbf{Two different modes of software development.} In human-in-the-loop development, coding agents generate code under continuous human oversight, guidance, review, and intervention. In autonomous software development, agents independently transform high-level requirements into complete, functional, and deployable software systems without human guidance or intervention.}
    \label{fig:intro-motivation}
\end{figure}

\section{Related Work}

\paragraph{Agent Harnesses.}
An agent harness is the operational layer that determines what information an
LLM receives, what actions it can execute, and how execution results enter
subsequent decisions~\citep{li2026agentharness,liu2026diveclaudecodedesign}.
Many mechanisms now assembled within harnesses were developed as distinct
research directions. Prompting and context engineering shape model-facing
state~\citep{liu2023pretrain,zhang2026ace}; external memory extends the state
available across interactions~\citep{packer2024memgptllmsoperatingsystems};
ReAct couples reasoning with environment actions~\citep{yao2023react}; and
GPTSwarm represents multi-agent orchestration as an optimizable graph~\citep{pmlr-v235-zhuge24a}.
More recent work treats the harness itself as the optimization target:
AutoHarness synthesizes a code harness from environment feedback, Meta-Harness
searches over harness code, and Self-Harness iteratively diagnoses and modifies
its own harness~\citep{lou2026autoharness,lee2026metaharness,zhang2026selfharness}.
These approaches improve agent behavior by changing the operational layer.
HoH builds on existing agent harnesses and iteratively improves an evolving software project through repeated implementation, evaluation, and refinement.
% HoH instead keeps the harness--model configuration fixed within a run and
% organizes how it develops and verifies an evolving software project.

\paragraph{Agentic Systems for Software Development.}
Research has progressed from localized code generation and self-contained
programs~\citep{chen2021codex,huang2023agentcoder} to repository-level issue
resolution, agent--computer interfaces, general software-engineering agents,
and refactoring~\citep{jimenez2024swebench,yang2024sweagent,xia2025agentless,wang2025openhands,oueslati2026refagent}.
Beyond repository issue resolution, MetaGPT and ChatDev use predefined
role-based workflows for software generation~\citep{hong2024metagpt,qian2024chatdev};
AgileCoder and EvoDev organize incremental development around sprints or
dependent features~\citep{nguyen2025agilecoder,liu2026evodev}; and EvoMAC
adapts the multi-agent workflow using test feedback~\citep{hu2025evomac}.
Recent benchmarks broaden both the development settings and the capabilities
under evaluation~\citep{le2025sweevo,orlanski2026slopcodebench,fu2025catarena}. SWE-EVO and
SlopCodeBench study long-horizon evolution and degradation,
while Commit0, \mbox{ProjDevBench}, ProgramBench, and \mbox{GameCraft-Bench}
evaluate from-scratch construction of complete libraries or projects~\citep{zhao2025commit0,lu2026projdevbench,yang2026programbench,luo2026gamecraftbench}.
FrontierSWE further covers from-scratch implementation together with open-ended
performance and research objectives~\citep{proximal2026frontierswe}. Existing
coding harnesses typically organize development within a bounded episode,
providing limited support for preserving project decisions, verified
functionality, and evaluation evidence across subsequent revisions. HoH builds
on these harnesses and extends their use to iterative greenfield development by
maintaining continuity across planning, implementation, and evaluation cycles.

\section{Harness-of-Harness}

Harness-of-Harness (HoH) organizes a fixed coding-agent system into a
long-running cycle of planning, development, and independent testing. Each
cycle produces a bounded software increment, verifies the resulting candidate,
and carries both the candidate and its execution evidence into the next cycle.
The design follows iterative and incremental software development: the system
grows through small, testable changes while preserving behavior that has
already been validated.

\begin{figure}[t]
    \centering
    \includegraphics[width=\textwidth]{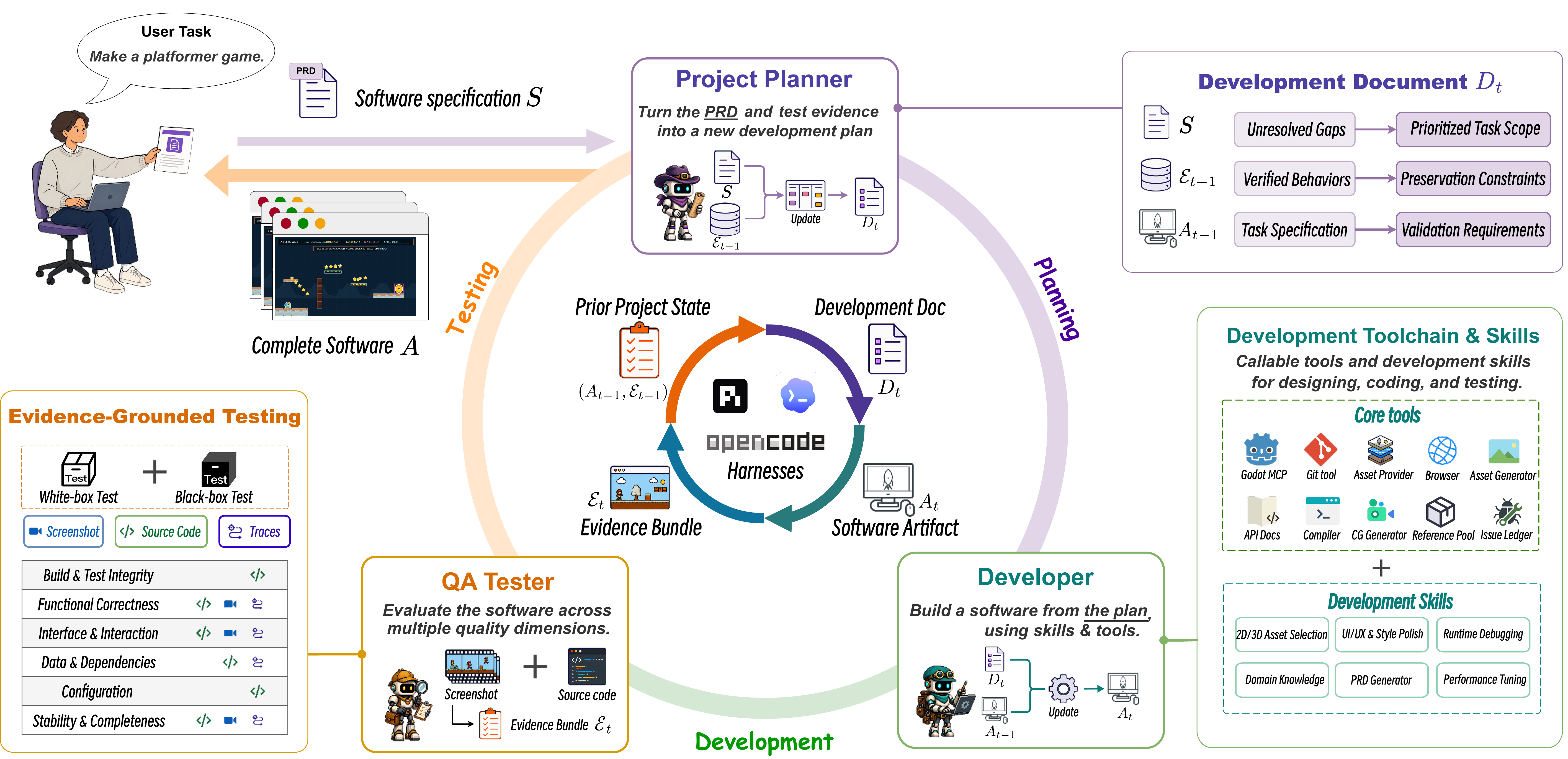}
    \caption{\textbf{Harness-of-Harness overview. }HoH repeatedly invokes a
    \emph{Project Planner}, \emph{Developer}, and \emph{QA Tester} around an
    evolving software artifact. The deterministic Runtime freezes each role's
    inputs, enforces its permissions, binds evidence to the tested candidate,
    and records the resulting project state. The model, base harness, role
    definitions, and runtime policy remain fixed within a run; the development
    document, software artifact, and execution evidence evolve across
    iterations.}
    \label{fig:framework-revised}
\end{figure}

\subsection{Problem Formulation and Challenges}

Given a software specification $\mathcal{S}$, the end-to-end software
development task is to construct a complete software artifact $A$ that
satisfies its functional and quality requirements. Let $M$ denote a language
model and $H$ the coding harness through which it interacts with a software
environment. HoH applies a fixed harness--model configuration to this task:
\begin{equation}
\operatorname{HoH}_{M,H}:\mathcal{S}\longmapsto A.
\end{equation}

This setting presents three challenges. (1) As the artifact evolves over a long
development trajectory, earlier requirements, design decisions, observed
failures, and previously validated behavior can be forgotten or become
disconnected from subsequent changes. (2) A high-level specification often
leaves the next useful change underdetermined. Component dependencies and
evolving implementation constraints mean that locally reasonable changes can
conflict with existing behavior, while repeated inspection and repair may
consume iterations without advancing the complete system. (3) Functional and
quality requirements manifest through heterogeneous, scenario-specific
behaviors. Missing or incorrect behavior may therefore remain undetected,
allowing an incomplete artifact to be accepted as complete. To address these
challenges, HoH organizes planning, implementation, and independent
verification into a three-agent loop that is repeated across iterations, with
the evolving artifact and accumulated development evidence carried between
loops.

\subsection{Harness-of-Harness Overview}

In end-to-end software development, the next useful change cannot be determined
from the specification alone; it requires jointly interpreting the
high-level specification, the current artifact, and the evidence accumulated
during development. The artifact exposes component dependencies,
implementation constraints, and missing capabilities. Execution evidence
reveals observed failures, changes the priority of unmet requirements, and
identifies validated behavior that subsequent work should preserve.

HoH organizes this changing decision process around a bounded development
loop. Each loop starts from the current project state and selects one coherent
objective that groups the interdependent work needed for an observable software
increment while excluding unrelated changes. It then implements the increment
and evaluates the resulting artifact before further development begins.
Evaluation results inform the next objective by revealing unmet requirements
and observed failures, while identifying validated behavior that subsequent
changes should preserve. Repeating this unit allows repair, extension, and
preservation demands to be reprioritized as the artifact evolves, keeping local
work aligned with the end-to-end objective.

Producing a validated increment requires three different decisions. The system
must first determine what to change next from the specification and retained
project state. The second decision concerns how to realize that change in the
current artifact, where the appropriate implementation depends on details
encountered during development. The final decision is whether the resulting
behavior satisfies observable requirements. These decisions require different
context and authority: objective selection requires a project-level view,
implementation requires write access and local technical autonomy, and
acceptance requires an assessment that is independent of the implementation
claim. HoH assigns these responsibilities to a \emph{Project Planner}, a
\emph{Developer}, and a \emph{QA Tester}, respectively. Each loop invokes the
same harness--model configuration once in each role, in
planning--development--testing order.

\subsection{Cross-Loop State Management}

Repeated loops support iterative and incremental development only when a later
loop inherits more than the latest implementation. A software artifact records
the code, resources, and configuration that currently exist, but it does not
fully record why earlier changes were selected, which observed failures remain
unresolved, or which behavior has already been validated. Since each harness
invocation has bounded context, information retained only in its interaction
history disappears when the invocation ends. A later loop that receives only
the code must reconstruct the development state from the implementation. This
reconstruction can overlook unmet requirements, repeat work whose outcome is
already known, forget unresolved failures, or regress validated behavior.

HoH therefore maintains two complementary states across loop boundaries. The
artifact state carries the current implementation from one loop to the next.
The evidence state carries the validated knowledge needed to decide how that
implementation should change. Together, they preserve both the object under
development and the information accumulated by developing and evaluating it.

Let $A_t$ denote the software artifact state after loop $t$, including its
source code, configuration, resources, and project metadata. It records what
the software currently is and provides the concrete starting point for the
next increment. Let $\mathcal{E}_t$ denote the execution evidence state
obtained by evaluating $A_t$ against the specification and the current
development objective. It records which behaviors have been verified, which
claims remain unsupported, and which observed failures require further work.
Neither state subsumes the other: $A_t$ supplies the implementation on which
development operates, whereas $\mathcal{E}_t$ supplies the validated project
knowledge used to direct that development.

Let \(A_0\) denote the empty project workspace before the first loop. With
$\mathcal{E}_0=\emptyset$, the transition across loop $t$ can be summarized as
\begin{equation}
\left(A_{t-1},\mathcal{E}_{t-1}\right)
\xrightarrow{\text{loop }t\text{ under }\mathcal{S}}
\left(A_t,\mathcal{E}_t\right).
\end{equation}

The two states enter a loop in different ways. The Project Planner combines
the fixed specification $\mathcal{S}$ with $\mathcal{E}_{t-1}$ to determine
the next bounded increment. It also reads $A_{t-1}$ as implementation context
so that the selected work is grounded in the current project. The Developer
then starts from $A_{t-1}$ and realizes the increment, producing $A_t$. The QA
Tester evaluates this updated artifact and produces $\mathcal{E}_t$ for the
next planning decision.

At the loop boundary, $(A_t,\mathcal{E}_t)$ becomes the starting state of loop
$t+1$. Carrying $A_t$ forward allows implementation work to accumulate instead
of being reconstructed in every loop. Interpreting $\mathcal{E}_t$ under
$\mathcal{S}$ allows new observations to revise development priorities,
unresolved gaps to remain visible, and validated behavior to become a
preservation requirement. The next objective can therefore build on prior
progress without reconstructing the project trajectory from the artifact
alone. Artifact continuity makes development incremental, and evidence-guided
objective selection makes it iterative.

\subsection{Implementation of a HoH Loop}

A HoH loop converts retained project state into a coherent software increment
whose behavior is independently assessed. This transformation begins with
objective selection. The global specification and prior evidence may identify
many interdependent demands, so the loop needs a project-level decision about
which bounded, locally complete subset should be addressed next. Establishing
this scope before artifact modification gives the increment observable
completion conditions and separates it from unrelated work.

Realizing the selected objective is a different function. The current artifact
exposes implementation-specific choices that cannot be fully determined during
planning, so artifact modification requires write authority and autonomy over
local technical decisions. Assessing the result introduces a third function.
The implementing agent has direct knowledge of its changes, but its completion
claim cannot establish that the intended behavior is present. Acceptance must
instead be determined from observations of a fixed candidate by a role that did
not produce that candidate.

These functions differ in the information they require, the authority they
exercise, and the deliverable they produce. HoH therefore assigns objective
selection to a Project Planner, artifact modification to a Developer, and
independent acceptance to a QA Tester. The separation makes the target of an
increment explicit, preserves implementation autonomy within that target, and
prevents implementation and acceptance from collapsing into the same decision.

HoH instantiates the three roles as separate invocations of the same fixed
harness--model configuration. For each invocation, a role-specific prompt
specifies the role's responsibility, while a deterministic Runtime contract
enforces its execution authority. The prompt combines fixed role instructions
with loop-specific context to state the role's objective and required structured
output, without prescribing its reasoning process or tool sequence.

The Runtime controls which inputs an invocation can access, which tools and
write operations it may use, and which output schema it must satisfy. HoH thus
constrains what each role may read, change, and deliver while leaving the agent
free to determine how to complete its assigned work within those boundaries.

\subsubsection{Project Planning}

The global specification may describe capabilities whose implementation spans
interdependent components, while the retained project state adds observed
failures, unmet requirements, and behaviors that must be preserved. These
demands describe what remains relevant to the project, but they do not by
themselves define a tractable unit of work for one loop. Selecting an isolated
task can omit dependencies needed for observable behavior, whereas combining
too many unrelated demands enlarges the change surface. When the resulting
candidate fails, the source of the failure becomes harder to localize, and the
affected behavior becomes harder to verify.

The Project Planner converts these competing demands into one bounded
objective. It reconciles $\mathcal{S}$ with $\mathcal{E}_{t-1}$ to determine
what should be addressed next and what previously validated behavior must be
preserved. It reads $A_{t-1}$ as implementation context so that the objective
reflects the current project structure, but it cannot modify the artifact. The
result is a development document $D_t$ that defines the scope and validation
conditions of the current increment.

The objective is bounded but locally complete. Boundedness limits the amount
of unrelated behavior changed in one loop, which keeps implementation and
diagnosis tractable. Local completeness ensures that the selected capability
includes the related changes required to make it functional and testable. The
scope of an increment is therefore determined by a coherent observable
behavior, not simply by the number of files or components it touches.

Accordingly, $D_t$ contains a small set of related tasks, the functionality
that must be preserved, and observable requirements for validating the
increment. Related changes may span several files or components when they are
jointly required by the objective. Unrelated refactoring and opportunistic
feature expansion remain outside the loop. The document specifies expected
behavior and validation conditions, while leaving the Developer to choose its
reasoning process, tools, and implementation algorithm.

\subsubsection{Artifact Development}

A development document defines the intended behavior of an increment, but it
cannot anticipate every implementation decision exposed by the evolving
artifact. The Developer must interpret $D_t$ in the context of the existing
project and adapt its implementation as it encounters code structure,
dependencies, and runtime behavior. HoH therefore constrains the Developer by
the required outcome and artifact boundary instead of prescribing its internal
procedure. Within these constraints, the Developer remains free to select the
concrete design, tools, and debugging strategy appropriate to the current
artifact.

Artifact development follows a single-writer boundary: only the Developer may
modify the evolving artifact. The Developer warm-starts from $A_{t-1}$ so that
each increment extends the current implementation and retains the surrounding
project structure. The Planner may inspect $A_{t-1}$ to ground the objective,
and the QA Tester may inspect and execute the resulting candidate, but neither
may alter the artifact. This boundary makes responsibility for the transition
from $A_{t-1}$ to $A_t$ explicit and keeps the candidate lineage unambiguous.
Once the authorized modifications are complete, the updated project becomes
$A_t$.

Testing is also integrated into artifact development so that failures are
exposed close to the changes that cause them. Before editing, the Developer
establishes a baseline for the target behavior. After each meaningful change,
it reruns the corresponding path and inspects the affected implementation,
execution results, and adjacent regression surface. This
baseline--change--retest cycle follows the software-engineering principle
commonly known as \emph{shift-left testing}. Shortening the distance between a
change and its test makes local diagnosis and correction more tractable.

Developer testing and independent acceptance answer different questions. The
Developer uses self-tests to determine whether the implementation is ready to
be presented as a candidate and to repair failures encountered during its own
work. These tests do not establish that the product requirements have been
satisfied. Developer observations and completion claims therefore remain
inputs to subsequent verification; acceptance is reserved for the independent
QA stage.

\subsubsection{Independent Quality Assurance}

Independent QA determines whether the candidate exhibits the behavior required
by the current objective while preserving relevant existing functionality.
End-to-end software quality is multidimensional and cannot generally be reduced
to one fixed performance metric. The relevant functional behavior,
interaction flows, configuration, resources, and regression risks depend on
both the global specification and the selected increment. HoH therefore
derives scenario-specific, checkable evaluation criteria from $\mathcal{S}$
and $D_t$ rather than applying the same generic test to every candidate.

The QA Tester receives $A_t$ as a frozen, read-only candidate together with
the results of deterministic build and execution checks. Freezing separates
artifact production from artifact assessment: the implementation cannot change
while its evidence is being collected. It also gives every observation a
single candidate identity, so that an assessment cannot combine behavior from
different artifact versions. Read-only access prevents the QA stage from
silently repairing the candidate it is meant to evaluate.

For each criterion, the QA Tester selects observations appropriate to the
software scenario. Black-box tests exercise the candidate through ordinary
inputs and rendered outputs to examine user-observable behavior, state
transitions, and end-to-end flows. These observations establish whether the
required behavior is visible at the product boundary. White-box tests inspect
the source, configuration, resource bindings, runtime state, and logs. They
help diagnose failures and corroborate observations whose internal conditions
cannot be determined from outputs alone.

The two forms of testing provide complementary views of the same frozen
candidate. A criterion is verified only when candidate-bound records support
the required behavior. Observed failures, unmet requirements, regressions, and
insufficient evidence are recorded as gaps instead of being inferred as
successful completion. The resulting assessments and supporting execution
records form the evidence state $\mathcal{E}_t$ passed to the next loop. This
separation ensures that acceptance follows observable evidence rather than the
Developer's knowledge of its implementation or its completion claim.

The complete HoH procedure is summarized in
Algorithm~\ref{alg:hoh-revised}, which combines the cross-loop state transition
with project planning, artifact development, independent quality assurance,
and Runtime validation.

\begin{algorithm}[tb]
\caption{Harness-of-Harness}
\label{alg:hoh-revised}
\textbf{Input}: specification $\mathcal{S}$, initial artifact $A_0$,
iteration budget $T$\\
\textbf{Fixed}: model $M$, harness $H$, and role contracts\\
\textbf{Output}: final artifact $A_T$
\begin{algorithmic}[1]
\STATE $\mathcal{E}_0 \gets \emptyset$
\FOR{$t=1,\ldots,T$}
    \STATE $D_t \gets \operatorname{ProjectPlanner}
        (\mathcal{S},\mathcal{E}_{t-1};
        \operatorname{read\_only}(A_{t-1}))$
    % \STATE \texttt{Runtime.validate\_and\_freeze}($D_t$)
    \STATE $A_t \gets \operatorname{Developer}
        (A_{t-1};\mathcal{S},D_t)$
    % \STATE \texttt{Runtime.validate\_and\_freeze}($A_t$)
    \STATE $\mathcal{E}_t \gets \operatorname{QATester}
        (\operatorname{read\_only}(A_t);\mathcal{S},D_t,
        \operatorname{Runtime.check}(A_t))$
    % \STATE \texttt{Runtime.validate\_binding}($\mathcal{E}_t,A_t$)
\ENDFOR
\STATE \textbf{return} $A_T$
\end{algorithmic}
\end{algorithm}

\FloatBarrier
\section{Experiments: Benchmark Evaluation}
\label{sec:experiments}

We evaluate HoH on three software-development benchmarks and three harness--model configurations, comparing final artifact quality with the corresponding Vanilla baselines.

% Required packages:
% \usepackage{booktabs}
% \usepackage{array}
% \usepackage[table]{xcolor}

% Sources: supplementary/data/{codex_gpt55_scores,
% opencode_deepseek_v4_pro_scores,minimax_m3_scores}.csv;
% supplementary/data/{frontierswe_task_records,
% frontierswe_main_aggregates}.csv; and
% supplementary/data/programbench/programbench_pass_rate_by_iteration.csv.
% FrontierSWE Dominance follows the 12-configuration macro-average procedure
% documented in supplementary/supplementary.tex. ProgramBench uses loop_1--3,
% not best_of_3.
\definecolor{codexbase}{RGB}{103,92,193}
\definecolor{opencodebase}{RGB}{51,123,119}
\definecolor{pibase}{RGB}{153,101,70}
\definecolor{improvegreen}{RGB}{0,105,0}

\newcommand{\MainResultsTable}{%
\begin{table}[ht]
    \centering
    \begingroup
    \footnotesize
    \setlength{\tabcolsep}{1.2pt}
    \renewcommand{\arraystretch}{1.05}

    \newcolumntype{R}[1]{>{\raggedleft\arraybackslash}p{##1}}
    \newcolumntype{C}[1]{>{\centering\arraybackslash}p{##1}}
    \newcommand{\harnessrow}[3]{%
        \addlinespace[0.22em]
        \multicolumn{12}{c}{%
            \textcolor{##1}{\rule{0.65em}{0.65em}}%
            \hspace{0.45em}\textbf{##2}%
            \hspace{0.35em}+\hspace{0.35em}##3%
        }\\[0.15em]
    }
    \newcommand{\groupdivider}{%
        \cmidrule{1-12}%
    }
    \newcommand{\metrichead}[1]{%
        \multicolumn{1}{c}{##1}%
    }
    \newcommand{\best}[1]{\textbf{##1}}
    \newcommand{\resultdelta}[2]{%
        \makebox[\linewidth][c]{%
            ##1\hspace{0.28em}%
            {\fontsize{5.5}{5.5}\selectfont\textcolor{improvegreen}{(##2)}}%
        }%
    }

    \begin{tabular*}{\textwidth}{
        @{\extracolsep{\fill}}
        >{\raggedright\arraybackslash}p{0.070\textwidth}
        R{0.058\textwidth} % GC Action
        R{0.058\textwidth} % GC Timing
        R{0.058\textwidth} % GC Strategy
        R{0.058\textwidth} % GC Simulation
        R{0.058\textwidth} % GC Adventure
        C{0.110\textwidth} % GC Overall
        @{\hspace{0.4em}}
        R{0.073\textwidth} % FS Implementation
        R{0.073\textwidth} % FS Performance
        R{0.073\textwidth} % FS Research
        C{0.110\textwidth} % FS Dominance
        @{\hspace{0.4em}}
        C{0.110\textwidth} % ProgramBench Avg. Test Pass Rate
        @{}
    }
        \toprule
        \multirow{2}{*}{\textbf{Setting}}
        & \multicolumn{6}{c}{\textbf{GameCraft-Bench}}
        & \multicolumn{4}{c}{\textbf{FrontierSWE}}
        & \multicolumn{1}{c}{\textbf{ProgramBench}} \\
        \cmidrule(lr){2-7}
        \cmidrule(lr){8-11}
        \cmidrule(lr){12-12}
        & \metrichead{Action}
        & \metrichead{Timing}
        & \metrichead{Strat.}
        & \metrichead{Sim.}
        & \metrichead{Adv.}
        & \metrichead{\textbf{\textit{Overall}}}
        & \metrichead{Impl.}
        & \metrichead{Perf.}
        & \metrichead{Research}
        & \metrichead{\textbf{\textit{Dominance}}}
        & \metrichead{\textbf{\textit{Pass Rate\textsuperscript{$\dagger$}}}} \\
        \midrule

        \harnessrow{codexbase}{Codex}{GPT-5.5 (high)}
        Vanilla
        & 48.74 & 48.80 & 44.06 & 53.63 & 52.68 & 49.58
        & 0.21 & 0.17 & 1.15 & 44\% & 60.41 \\
        HoH@1
        & 59.73 & 53.79 & 57.01 & 66.75 & 61.24 & 59.71
        & 0.24 & 0.44 & 1.30 & 58\% & 65.42 \\
        HoH@2
        & 64.34 & 62.03 & 59.97
        & 71.77 & 66.11 & 64.84
        & 0.28 & 0.45 & 1.18 & 60\% & 65.79 \\
        HoH@3
        & \best{71.02} & \best{70.26} & \best{66.13}
        & \best{78.42} & \best{71.76} & \resultdelta{\best{71.52}}{+21.93}
        & \best{0.30} & \best{0.45} & \best{1.45} & \resultdelta{\best{71\%}}{+27} & \resultdelta{\best{66.50}}{+6.09} \\

        \groupdivider
        \harnessrow{opencodebase}{OpenCode}{DeepSeek-V4-Pro}
        Vanilla
        & 26.21 & 24.05 & 21.27 & 37.12 & 25.84 & 26.90
        & 0.08 & 0.23 & 0.57 & 25\% & 45.27 \\
        HoH@1
        & 27.75 & 27.40 & 21.73 & 43.75 & 22.44 & 28.61
        & 0.09 & 0.23 & 0.78 & 28\% & 55.33 \\
        HoH@2
        & 43.22 & 36.56 & 33.51
        & 52.26 & 36.05 & 40.32
        & 0.10 & 0.27 & 0.78
        & 42\% & 55.66 \\
        HoH@3
        & \best{49.00} & \best{45.05} & \best{43.34}
        & \best{55.86} & \best{51.64} & \resultdelta{\best{48.98}}{+22.08}
        & \best{0.15} & \best{0.27} & \best{0.78}
        & \resultdelta{\best{44\%}}{+19} & \resultdelta{\best{57.56}}{+12.29} \\

        \groupdivider
        \harnessrow{pibase}{Pi}{MiniMax-M3}
        Vanilla
        & 45.64 & 38.20 & 34.53 & 48.19 & 44.26 & 42.16
        & 0.06 & 0.12 & 1.30 & 35\% & 35.83 \\
        HoH@1
        & 50.59 & 52.23 & 38.60 & 54.00 & 49.89 & 49.06
        & 0.10 & 0.42 & \best{1.89} & 62\% & 48.68 \\
        HoH@2
        & 54.70 & 56.52 & 42.33
        & 63.20 & 58.47 & 55.04
        & 0.11 & 0.43 & \best{1.89}
        & \best{66\%} & \best{53.57} \\
        HoH@3
        & \best{58.24} & \best{62.10} & \best{44.86}
        & \best{64.25} & \best{64.44} & \resultdelta{\best{58.78}}{+16.62}
        & \best{0.11} & \best{0.45} & 1.88
        & \resultdelta{64\%}{+29} & \resultdelta{52.68}{+16.85} \\
        \bottomrule
    \end{tabular*}
    \endgroup

    \caption{%
        \textbf{Main results on GameCraft-Bench, FrontierSWE, and ProgramBench.}
        Each harness is evaluated under Vanilla and HoH@1--3. Within each
        harness and metric, bold values mark the best setting; rankings use
        unrounded values, and exact ties share the same formatting. Small green
        values shown only for HoH@3 give absolute gains over Vanilla;
        Dominance gains are percentage points. Bold italic labels distinguish
        aggregate metrics from task categories.
        Strat., Sim., Adv., Impl., and Perf. denote Strategy, Simulation,
        Adventure, Implementation, and Performance, respectively.
        FrontierSWE reports category scores and Dominance. For ProgramBench,
        \emph{Pass Rate}\textsuperscript{$\dagger$} denotes the benchmark's
        \emph{Avg. Test Pass Rate}.%
    }
    \label{tab:main-results}
\end{table}
}

\subsection{Experimental Setup}

\paragraph{Benchmarks.}

We evaluate HoH on three benchmarks: \mbox{GameCraft-Bench}~\citep{luo2026gamecraftbench}, \mbox{FrontierSWE}~\citep{proximal2026frontierswe}, and \mbox{ProgramBench}~\citep{yang2026programbench}. GameCraft-Bench comprises 140 tasks across 15 game families, each requiring an agent to construct a complete, playable Godot project from a natural-language specification. We sample 45 tasks using stratified random sampling by game family, selecting three tasks from each of the 15 families with a fixed random seed. For coarse-grained analysis, we additionally organize the 15 families into five broader groups defined in this work: Action, Timing, Strategy, Simulation, and Adventure. The complete sampled task list and our family-to-group mapping are provided in the supplementary material.
Due to computational resource constraints, we select $15$ tasks from FrontierSWE's $17$ tasks for evaluation, comprising $4$ Implementation (Impl.), $9$ Performance (Perf.), and $2$ Research tasks. ProgramBench is a cleanroom program-reconstruction benchmark in which agents receive only a compiled executable and documentation and must rebuild a codebase whose behavior matches the reference program. More details are provided in the supplementary material.

\paragraph{Harnesses and Models.}

We evaluate HoH with three harness--model configurations: Codex CLI\footnote{\url{https://github.com/openai/codex}; version 0.142.5.} with GPT-5.5 at high reasoning effort, OpenCode\footnote{\url{https://github.com/anomalyco/opencode}; version 1.14.30.} with DeepSeek-V4-Pro, and Pi Coding Agent\footnote{\url{https://github.com/earendil-works/pi}; version 0.80.10.} with MiniMax-M3.

\paragraph{Baseline and Evaluation Protocol.}

We compare HoH against \emph{Vanilla}, the corresponding harness--model configuration without the HoH protocol. Vanilla performs one standard development pass, whereas \emph{HoH@$T$} performs $T$ planning--coding--testing iterations, with the software artifact and execution evidence carried across iterations; the main experiments use $T=3$. Intermediate and final HoH artifacts are evaluated only after the complete run, and evaluator outputs are not returned to the development loop. For each task, Vanilla and HoH use identical benchmark-provided initial states and the same underlying harness--model configuration, differing only in the application of the HoH protocol.

\paragraph{Metrics.}

For GameCraft-Bench, we report the benchmark's \emph{Overall} score. Under this metric, game artifacts that fail to compile or run receive a score of zero, while runnable artifacts are scored by combining Core Mechanics, Content Depth, Functional Visuals, and Art and Presentation using the benchmark-defined weights. We average task-level scores over the 45 tasks and use the benchmark's 0--100 scale. For FrontierSWE, task-specific verifiers assign official rewards, and we report the mean reward over the 15 evaluated tasks. We additionally report the official dominance score, defined as the average task-level win rate against a randomly selected competing configuration from the 12 evaluated harness--condition combinations. For ProgramBench, we report \emph{Avg. Test Pass Rate}, computed as the mean across tasks of the fraction of hidden behavioral tests passed for each task. We use this continuous signal for relative comparisons between Vanilla and HoH and abbreviate it as \emph{Pass Rate\textsuperscript{$\dagger$}} in Table~\ref{tab:main-results}. As a proxy for model-interaction volume, we report provider-reported cumulative input and output tokens from coding-harness model calls, excluding benchmark evaluation. Input totals may include cached context reads; because cache accounting differs across providers, we use these values for within-configuration comparisons rather than direct cross-provider cost comparisons.

\subsection{Main Results}

\MainResultsTable

\paragraph{HoH improves software artifact quality across three benchmarks spanning game development, repository-level software engineering, and program reconstruction.}
Table~\ref{tab:main-results} reports Vanilla and all three HoH iterations for each harness--model configuration. HoH@3 outperforms Vanilla across the three benchmarks under all three configurations. On GameCraft-Bench, mean Overall scores increase from 49.58 to 71.52 for Codex, from 26.90 to 48.98 for OpenCode, and from 42.16 to 58.78 for Pi. On FrontierSWE, rewards increase from 0.31 to 0.54, from 0.23 to 0.31, and from 0.26 to 0.55, respectively. On ProgramBench, \emph{Avg. Test Pass Rate} increases from 60.41 to 66.50 for Codex, from 45.27 to 57.56 for OpenCode, and from 35.83 to 52.68 for Pi. HoH@3 also outperforms Vanilla in every reported task category across the three benchmarks under all three configurations.

\paragraph{HoH yields consistent gains over Vanilla across all three harness--model pairs.}
The gains are not limited to configurations with a particular level of Vanilla performance. Codex with GPT-5.5 (high), the strongest Vanilla configuration, reaches the highest final GameCraft-Bench score of 71.52 after improving by 21.93 points. Pi with MiniMax-M3 records the largest gain on FrontierSWE, increasing by 0.29 from 0.26 to 0.55, and the largest ProgramBench gain, increasing the average test pass rate by 16.85 points. OpenCode with DeepSeek-V4-Pro starts from the lowest Vanilla score on GameCraft-Bench and FrontierSWE, yet HoH@3 raises its scores to 48.98 and 0.31, respectively, while increasing its ProgramBench average test pass rate from 45.27 to 57.56. OpenCode with HoH@3 further exceeds Codex Vanilla in Action and Simulation on GameCraft-Bench and in Performance on FrontierSWE. Thus, HoH improves configurations that begin at substantially different levels of Vanilla performance.

\paragraph{HoH continues to improve software artifact quality as development loops progress.}
Table~\ref{tab:main-results} traces the gains accumulated over the first three development loops. On GameCraft-Bench, \emph{Overall} scores increase monotonically from HoH@1 to HoH@3 under all three harness--model pairs. This trend is particularly pronounced for OpenCode, whose gain over Vanilla grows from 1.71 points at HoH@1 to 13.42 at HoH@2 and 22.08 at HoH@3. On FrontierSWE, the cross-configuration \emph{Dominance} of Codex increases from 44\% under Vanilla to 58\%, 60\%, and 71\% at HoH@1--3, respectively. ProgramBench exhibits a similar overall pattern: Codex and OpenCode attain their highest \emph{Pass Rates} at HoH@3, while Pi peaks at HoH@2.

\subsection{Analysis and Ablation Study}

\begin{figure}[ht]
    \centering
    \includegraphics[width=\textwidth]{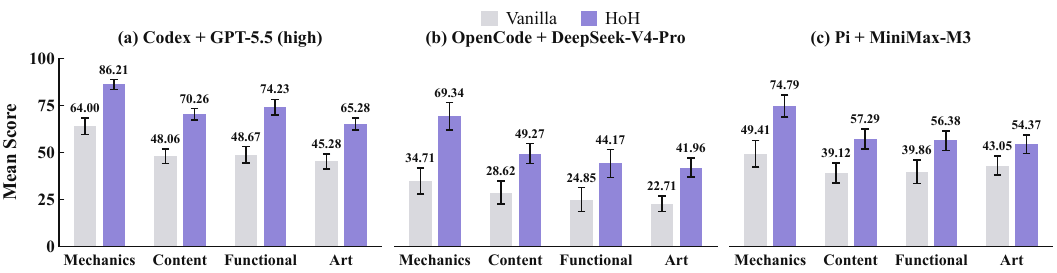}
    \caption{\textbf{Vanilla and HoH@3 scores across the four GameCraft-Bench rubric categories.} Panels (a)--(c) show results for Codex with GPT-5.5 (high), OpenCode with DeepSeek-V4-Pro, and Pi with MiniMax-M3, respectively. Bars report mean category scores over 45 tasks for Core Mechanics, Content Depth, Functional Visuals, and Art and Presentation; error bars indicate 95\% bootstrap confidence intervals.}
    \label{fig:gamecraft-component-scores}
\end{figure}

\paragraph{On GameCraft-Bench, HoH improves software quality across mechanics, content, visuals, and presentation.}
Figure~\ref{fig:gamecraft-component-scores} reports the four benchmark-defined GameCraft-Bench quality components separately. HoH@3 improves all four components under every harness--model configuration, with gains of 20.00--25.56 points for Codex, 19.25--34.63 points for OpenCode, and 11.32--25.38 points for Pi. For Codex, Functional Visuals shows the largest increase, from 48.67 to 74.23, while Art and Presentation rises from 45.28 to 65.28. The improvements therefore span gameplay mechanics, content richness, visual clarity, and presentation quality.

\begin{wrapfigure}{r}{0.40\textwidth}
    \vspace{-1.5\baselineskip}
    \centering
    \includegraphics[width=0.36\textwidth]{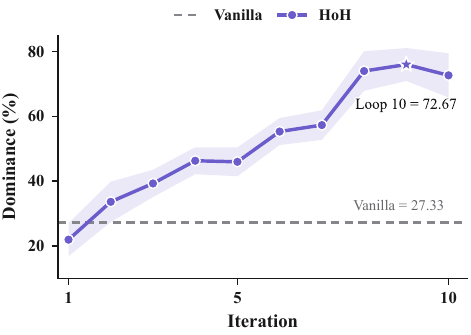}
    \captionsetup{hypcap=false,skip=2pt,font=footnotesize,justification=raggedright}
    \caption{\fontsize{9pt}{8.5pt}\selectfont\textbf{FrontierSWE Dominance over 10 loops.} Shading shows $\pm 1$ SE; the star marks the best checkpoint and the dashed line denotes Vanilla baseline.}
    \label{fig:frontierswe-force10-dominance}
    \vspace{-0.4\baselineskip}
\end{wrapfigure}

\paragraph{On \mbox{FrontierSWE}, HoH sustains quality gains over ten loops.}
{\looseness=-1
To examine whether these gains extend beyond three loops, we continue running Codex with GPT-5.5 (high) through HoH@10 on the same 15 \mbox{FrontierSWE} tasks and report \emph{Dominance} over a fixed 11-checkpoint comparison pool comprising Vanilla and HoH@1--10.
\par}

As shown in Figure~\ref{fig:frontierswe-force10-dominance}, \emph{Dominance} increases from 39.33\% at HoH@3 to 72.67\% at HoH@10 and reaches 76.00\% at HoH@9, whereas Vanilla obtains 27.33\%. HoH@10 therefore improves upon HoH@3 by a further 33.34 percentage points and exceeds Vanilla by 45.34 points.

\WFclear

\begin{figure*}[!t]
    \centering
    \includegraphics[width=\textwidth]{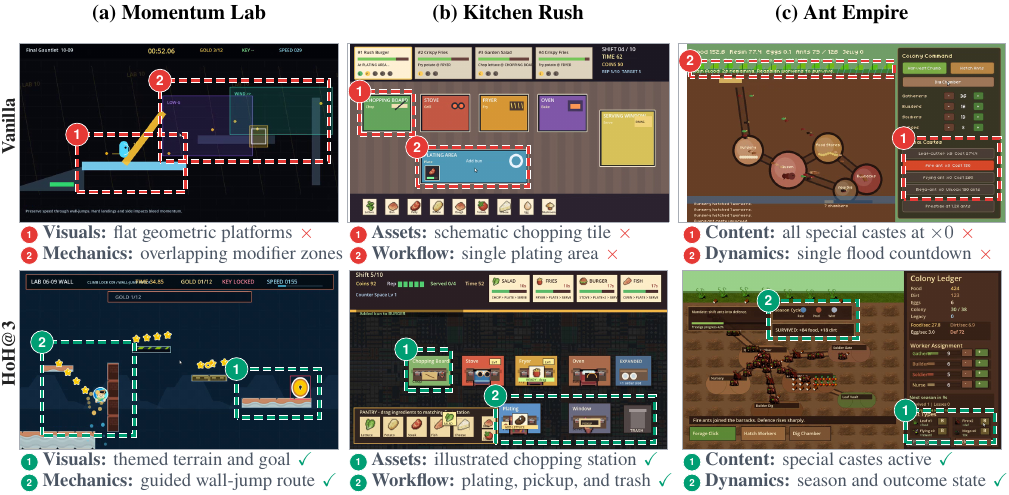}
    \captionsetup{width=\textwidth}
    \caption{\textbf{Qualitative comparison of final game artifacts produced by Vanilla and HoH@3 using Codex with GPT-5.5 (high).} Columns show three GameCraft-Bench tasks from distinct game families: Momentum Lab (momentum-based platformer), Kitchen Rush (restaurant-management simulation), and Ant Empire (idle colony-management game). Rows show gameplay frames from Vanilla (top) and HoH@3 (bottom). Numbered dashed boxes identify the regions discussed in the annotations; red crosses and green checks denote limitations and implemented functionality, respectively.}
    \label{fig:qualitative-examples}
\end{figure*}

\paragraph{For the same number of development passes, HoH consistently outperforms the Vanilla baseline.}
\label{subsec:budget-comparison}

To distinguish the contribution of HoH from the effect of running the coding agent for more passes, we compare it with Vanilla Continuation using Codex with GPT-5.5 (high). Vanilla uses the official harness configuration with the same model and inference settings as HoH. After each pass, Vanilla Continuation submits an additional iteration prompt to continue the same session for another development pass. Table~\ref{tab:budget-comparison} reports the resulting pass-controlled comparison.

\Needspace{12\baselineskip}
\begin{wraptable}{r}{0.48\textwidth}
    \vspace{-0.6\baselineskip}
    \centering
    \begingroup
    \footnotesize
    \setlength{\tabcolsep}{2.5pt}
    \begin{tabular}{@{}lccc@{}}
        \toprule
        \textbf{Method} & \textbf{Dev. Passes} & \textbf{Score} & \textbf{Tokens ($\mathrm{M}$)} \\
        \midrule
        Vanilla & 1 & 49.58 & 2.59 \\
        Vanilla Continuation & 2 & 54.99 & 4.56 \\
        Vanilla Continuation & 3 & 58.24 & 6.33 \\
        \midrule
        HoH & 1 & 59.71 & 2.88 \\
        HoH & 2 & 64.84 & 5.67 \\
        HoH & 3 & \textbf{71.52} & 8.41 \\
        \bottomrule
    \end{tabular}
    \endgroup
    \caption{\textbf{Comparison of HoH and repeated Vanilla development after 1, 2, and 3 development passes on GameCraft-Bench.} Results use Codex with GPT-5.5 (high). Score denotes the mean GameCraft-Bench Overall score over 45 tasks, and tokens are mean cumulative coding-harness tokens per task.}
    \label{tab:budget-comparison}
    \vspace{-0.6\baselineskip}
\end{wraptable}

At matched budgets of one, two, and three development passes, HoH achieves scores of 59.71, 64.84, and 71.52, compared with 49.58, 54.99, and 58.24 for Vanilla, corresponding to gains of 10.13, 9.85, and 13.28 points. The advantage is not explained by greater token use alone: HoH@2 achieves 64.84 with 5.67M tokens, exceeding the 58.24 obtained by three-pass Vanilla Continuation with 6.33M tokens. HoH therefore produces higher-quality artifacts than repeated Vanilla development under the same pass budget and a comparable inference budget. Further details of the pass-controlled experimental design and complete results are provided in the supplementary material.

\paragraph{Qualitative Analysis.}

On GameCraft-Bench, HoH produces more complete and refined game artifacts, with richer gameplay mechanics, clearer visual presentation, and deeper progression. Figure~\ref{fig:qualitative-examples} compares gameplay frames from Vanilla and HoH@3 for three tasks from distinct game families.

\FloatBarrier

In Momentum Lab, themed terrain and visual cues make the objective and wall-jump route explicit. In Kitchen Rush, distinct pickup, preparation, plating, and disposal stations form a complete and legible restaurant workflow. In Ant Empire, specialist caste counts, seasonal state, and outcome state expose longer-term colony progression. The corresponding \emph{Overall} scores increase from 34.05 to 70.61, from 42.63 to 73.38, and from 65.52 to 87.88, respectively.

\begin{wraptable}{r}{0.48\textwidth}
    \vspace{-0.6\baselineskip}
    \centering
    \begingroup
    \footnotesize
    \setlength{\tabcolsep}{3pt}
    \begin{tabular}{@{}lcc@{}}
        \toprule
        \textbf{Variant} &\textbf{Score} & \textbf{Tokens ($\mathrm{M}$)} \\
        \midrule
        \textit{w/o Plan Update} & 63.39 {\small($-8.13$)} & 7.56 \\
        \textit{w/o Evidence Feedback} & 65.23 {\small($-6.28$)} & 7.46 \\
        \textit{w/o Warm-Start} & 63.67 {\small($-7.85$)} & 11.12 \\
        \midrule
        Full HoH@3 & \textbf{71.52} & 8.41 \\
        \bottomrule
    \end{tabular}
    \endgroup
    \caption{\textbf{GameCraft-Bench ablation study with Codex and GPT-5.5 (high).} Parentheses show score differences from Full HoH@3; tokens are mean cumulative totals per task.}
    \label{tab:ablation}
    \vspace{-0.6\baselineskip}
\end{wraptable}

\paragraph{Ablation Study.}

To assess the role of HoH's cross-iteration mechanisms, we evaluate three variants on all 45 GameCraft-Bench tasks using Codex with GPT-5.5 (high), with $T=3$ for each variant. \textit{w/o Plan Update} freezes the first development document for subsequent iterations ($D_t=D_1$ for $t\geq2$), whereas \textit{w/o Evidence Feedback} replans without the preceding execution evidence. Both retain artifact warm-start. \textit{w/o Warm-Start} retains evidence-conditioned planning but rebuilds the artifact from the empty initial workspace $A_0$ in every iteration.

As shown in Table~\ref{tab:ablation}, all three variants underperform \textit{Full HoH@3} on every task. Removing plan updates, excluding execution evidence from replanning, and removing warm-start lowers the score by 8.13, 6.28, and 7.85 points, respectively. Without warm-start, token usage also increases from 8.41M to 11.12M per task because of repeated reconstruction. These results show that later HoH iterations benefit from both revising the development document with execution evidence and continuing from the preceding implementation. Detailed ablation protocols and complete per-task results are provided in the supplementary material.

\FloatBarrier
\section{Experiments: Multi-Day Autonomous FPS Game Development}
\label{sec:case-study-revised}

Benchmark evaluations measure artifact quality after a small number of
development loops. A multi-day case study examines a complementary property:
whether a fixed harness--model configuration can maintain coherent project
evolution as implementation constraints, validated behavior, and observed
failures accumulate over many loops. We study this property through
\textit{Fusepoint}, a single-player narrative first-person shooter developed
from an empty workspace containing only a user-provided product requirements
document (PRD). The case analyzes whether HoH can sustain incremental progress
over 70 loops as implementation constraints, validated behavior, and observed
failures accumulate.

\subsection{Case Design and Autonomy Boundary}

\paragraph{Task and Autonomy Boundary.}
\textit{Fusepoint} provides a demanding end-to-end development case. The
product contract specifies a five-minute, single-player bomb-defusal mission.
It requires the ordered capture of two control points, a three-stage defusal at
the final objective, a fixed roster of 18 enemies distributed as 3, 5, and 10
across the three encounter regions, and distinct success and detonation
branches. Satisfying these requirements depends on integrating a 3D environment
and external assets with mission logic, combat mechanics, narrative
progression, interface feedback, and runtime reliability. Progress therefore
requires both the construction of new capabilities and the continued operation
of behavior introduced in earlier loops.

Development began in an empty workspace containing the PRD. The PRD
specified the intended gameplay and player-observable acceptance criteria,
while leaving the engineering decomposition, implementation order, and
validation plan open. HoH was responsible for translating this product
contract into an executable Godot project and for selecting, implementing, and
evaluating the increments used to construct it.

We ran HoH with Codex CLI and GPT-5.6-Sol at high reasoning effort. At the
analysis cutoff, the system had completed 70 development loops. Human
involvement was limited to restoring network or API availability and did not
extend to planning, implementation, debugging, testing, or acceptance.

\paragraph{Domain-Specific Skills and Tools.}
Interactive game development requires more than source-code editing: the
agents must manipulate engine state, produce compatible media assets, maintain
a coherent interface, and test behavior through the running game. We therefore
equipped HoH with domain-specific tools and reusable skills. Godot
4.7\footnote{Godot Engine 4.7:
\url{https://godotengine.org/releases/4.7/}} served as the development and
runtime environment, while Godot MCP provided engine-level development,
execution, and debugging capabilities. An asset-generation skill specified the
target visual style, dimensions, and file formats for image, 3D, and video
assets retrieved or generated by the corresponding tools. A UI/UX presentation
skill supplied guidance on visual appearance and style consistency. A testing
skill recorded scenario-specific testing considerations and guided debugging
and validation through Godot MCP. All external assets incorporated during
development were obtained under licenses permitting reuse, including CC0 and
CC BY, with their source and required attribution preserved.

\paragraph{Verification Scope.}
Benchmark configurations evaluate through bounded task-provided checks,
including screenshots and smoke tests. For this case, the Tester additionally
examined live keyboard input and the resulting game responses, audio behavior,
and the integration of 3D assets.

\paragraph{Project State and Traceability.}
We used GitHub for version control and issue tracking, retaining the commit and
issue histories of the evolving project. At each loop, the working state was
materialized in a development document, the versioned software workspace, and
testing records comprising an issue table and evidence-packet files. These
records allowed the three roles to modify, execute, and inspect the same
project while retaining the artifact and observed test outcomes across loops.
The GitHub repository linked on the title page provides the released project
materials and selected development records for inspection of the process.

\subsection{Development Dynamics Across Loops}

Figure~\ref{fig:fusepoint-development-trajectory} summarizes how
\textit{Fusepoint} evolved through 70 HoH loops. To characterize the development
dynamics underlying this trajectory, we tracked newly recorded issues,
QA-verified closures, reopened issues, and the unresolved issue count using the
GitHub issue and commit histories together with the evidence packets produced
during testing. These records reflect whether capability growth was
accompanied by accumulating gaps and whether later loops returned to failures
exposed during earlier development.

% \begin{figure}[H]
%     \centering
%     \includegraphics[width=\textwidth]{Figures/fig_case_fusepoint_trajectory.pdf}
%     \caption{\textbf{Development and issue trajectory of \textit{Fusepoint}.}
%     The solid curve shows the smoothed trend in unresolved issues, and markers show the observed loop-end counts through Loop~70. Blue and plum tiles denote newly recorded and QA-verified closed issue events, respectively. Counts cover all product, evidence, and infrastructure issues in the canonical ledger.}
%     \label{fig:fusepoint-development-trajectory-revised}
% \end{figure}

The observed trajectory through 70 loops contains three broad phases,
distinguished by the relative prevalence of capability addition, issue
discovery, and issue resolution. During initial construction (Loops~1--27), HoH established
the executable project and its core interaction paths. Adding these initial
capabilities also exposed missing requirements and defects, so the active issue
backlog increased as the artifact became more testable.

Capability expansion (Loops~28--49) then combined new functionality with continued diagnosis
and repair. Later loops operated on an increasingly integrated artifact, where
a local change could affect mission state, combat, interface feedback, or
runtime behavior established previously. As the planned capabilities
approached completion, feature additions slowed and issue resolution became
more prevalent, producing a stabilization phase in which the active backlog
began to decline.

Issue resolution remained non-monotonic throughout this process. By Loop~70,
65 of the 81 recorded issues had been closed, leaving 16 unresolved. Seventeen
issues were reopened after an earlier closure when a subsequent change caused
previously verified behavior to fail again. A reopened record identifies both
the failed behavior and its earlier verification history, making regression
repair available as explicit project work to subsequent planning instead of
requiring that history to be reconstructed from the latest artifact.

% One interface regression illustrates this process. The Loop~25 evaluation
% recorded an opaque route-guidance card that violated the product's interface
% requirements. QA verified the issue as closed after Loop~26, but changes to the
% product shell and HUD in Loop~27 caused the same violation to reappear. The
% ledger retained the earlier verification and marked the issue as regressed;
% after a further repair, QA verified it again in Loop~29. Subsequent planning
% could therefore treat the recurrence as a regression of the same requirement,
% with its previous verification intact, instead of reconstructing its history
% from the modified interface.

% Loops~69--70 provide a complementary example of issue-level assessment. The
% Loop~69 evaluation found that rifle-equipped enemies still resolved
% pistol-authored animation clips. Loop~70 revised the enemy animation bindings,
% and the Tester verified the specific binding repair from source inspection and
% live actor state across all three encounter regions. The broader animation
% assessment remained unresolved, however, since the evaluation had not yet
% observed the complete aim, fire, reload, hurt, and death sequence. The evidence
% state thus separated a verified repair from the remaining coverage gap, so
% later development could preserve the corrected binding while continuing the
% incomplete validation.

The trajectory consequently reflects the two forms of continuity required by
iterative development. The versioned workspace allowed implementation work to
accumulate, while its GitHub commit history made individual changes traceable.
The issue history and evidence packets kept unfinished work, verified behavior,
and regressions available for later planning. Development could therefore
alternate among capability growth, repair, and preservation as the state of
the project changed.

\FloatBarrier

\FloatBarrier
\section{Conclusion and Future Work}
We introduced Harness-of-Harness (HoH), which extends existing coding-agent
harnesses to support software development from scratch without modifying their
implementations. HoH organizes a fixed harness--model configuration into a
continuous planning--coding--testing cycle, carrying evolving artifacts and
execution evidence across iterations. On the benchmark tasks, HoH improves
final artifact quality for all three evaluated configurations and continues to
benefit from additional iterations. In the multi-day \textit{Fusepoint} case,
HoH developed a game project over 70 loops, with the versioned workspace, issue
history, and evidence packets recording the development trajectory. Just
as a coding harness structures model operation, HoH structures harness
participation in long-horizon development. More broadly, HoH offers a practical
path toward end-to-end software development through persistent,
evidence-grounded orchestration of coding-agent harnesses. Future work will
extend HoH to a broader range of real-world development
scenarios, including different types of games and other software
systems~\citep{zhang2024advancing,zhang2024training,10989764,zhang2024mutualtheorymindhumanai,wang2023quantifying,zhang2025dpt}, toward a general framework
for autonomous software development.

\bibliographystyle{style/plainnat}
\bibliography{references}

\clearpage
\appendix
\begin{center}
  {\LARGE\sffamily\bfseries Supplementary Material}
\end{center}
\vspace{0.4cm}
\section{Method and Implementation Details}
\label{sec:supp-method}

\subsection{HoH Execution Protocol}

This section expands the method specification in the main paper into its
executable interfaces. Within one experimental condition, the \emph{Project
Planner}, \emph{Developer}, and \emph{Quality Assurance (QA) Tester} are three
independent invocations of the same harness--model configuration $H$. Their
model and native harness capabilities remain fixed, while role-specific
instructions determine what each invocation may read, modify, and return.
The three roles coordinate around the same evolving software artifact. The
Developer writes to the active project workspace, whereas the Planner consumes
materialized documents and the QA Tester inspects an isolated copy of the
current artifact. The latter two return structured records rather than
modifying the active artifact through an interactive conversation.

Table~\ref{tab:role-io} details the implementation-level inputs and outputs
corresponding to the notation in the main paper. The Planner receives the
public specification $\mathcal{S}$ and preceding evidence
$\mathcal{E}_{t-1}$ and produces the current development document $D_t$. The
Developer receives $\mathcal{S}$ and $D_t$ in the workspace containing
$A_{t-1}$ and writes the updated artifact $A_t$. The QA Tester receives
$\mathcal{S}$, $D_t$, and $A_t$, executes and inspects the artifact, and
produces $\mathcal{E}_t$.

Accordingly, one HoH iteration is implemented by three harness invocations:
\begin{equation}
\begin{aligned}
D_t
&= \mathrm{Plan}_{H}\left(\mathcal{S},\mathcal{E}_{t-1}\right),\\
A_t
&= \mathrm{Dev}_{H}\left(A_{t-1};\mathcal{S},D_t\right),\\
\mathcal{E}_t
&= \mathrm{Test}_{H}\left(A_t;\mathcal{S},D_t\right).
\end{aligned}
\label{eq:supp-hoh-iteration}
\end{equation}
The three invocations share the fixed configuration $H$, but each receives the
role-specific inputs shown in Table~\ref{tab:role-io}. The artifact and evidence
bundle cross the iteration boundary; within iteration $t$, $D_t$ provides the
common specification for coding and testing.

\begin{table}[H]
    \centering
    \caption{Inputs, invocation contracts, and materialized outputs of the
    three HoH roles. All roles use the harness--model configuration associated
    with the corresponding experimental condition.}
    \label{tab:role-io}
    \small
    \renewcommand{\arraystretch}{1.16}
    \setlength{\tabcolsep}{4pt}
    \begin{tabular}{@{}>{\centering\arraybackslash}m{0.14\textwidth}L{0.24\textwidth}L{0.34\textwidth}L{0.20\textwidth}@{}}
        \toprule
        \textbf{Role} & \textbf{Inputs} & \textbf{Invocation contract} &
        \textbf{Materialized output} \\
        \midrule
        Project Planner &
        $\mathcal{S}$ and $\mathcal{E}_{t-1}$ &
        Select bounded priorities from public requirements and preceding
        evidence; identify verified functionality to preserve; specify
        observable acceptance requirements; do not modify production code &
        Development document $D_t$ \\
        \cmidrule(lr){1-4}
        Developer &
        $\mathcal{S}$, $D_t$, and the workspace containing $A_{t-1}$ &
        Address prioritized targets with native coding tools; preserve
        verified functionality; keep the project buildable and runnable; write
        changes into the existing workspace &
        Updated artifact $A_t$ and execution records \\
        \cmidrule(lr){1-4}
        QA Tester &
        $\mathcal{S}$, $D_t$, $A_t$, and public execution records &
        Derive checkable claims; execute and inspect the artifact; associate
        findings with observable records; distinguish supported functionality
        from unresolved or insufficiently evidenced requirements &
        Evidence bundle $\mathcal{E}_t$ \\
        \bottomrule
    \end{tabular}
\end{table}

\subsection{Role-Specific Prompt Construction}

Each role prompt is rendered from reusable Markdown modules and runtime values.
The fixed modules specify role boundaries, public-information policy, and the
output contract; runtime slots insert the public task, current iteration state,
materialized documents, and execution records. The public task specification
is inserted without modification. At $t=1$, the Planner's evidence slot is
empty; for $t>1$, it contains the structured evidence bundle from the preceding
artifact. The templates below are schematic, interface-preserving renderings
of the runtime prompts: they retain the role contracts and data dependencies
used by the method while omitting repeated benchmark-specific examples and
checklists. In the templates,
\promptslot{runtime\_slot} denotes substituted content,
\promptpath{/workspace/path} denotes a materialized file or directory, and
\promptmodule{conditional module} denotes a block included only when its
runtime condition is satisfied.

\begin{plannerprompt}{Project Planner Prompt}
\small
\textbf{Template assembly.}\quad
\promptmodule{role instruction} $\oplus$
\promptmodule{public specification} $\oplus$
\promptmodule{preceding evidence} $\oplus$
\promptmodule{document scaffold} $\oplus$
\promptmodule{output contract}

\medskip
\textbf{Method correspondence.}\quad
$\left(\mathcal{S},\mathcal{E}_{t-1}\right)
\xrightarrow{\mathrm{Plan}_{H}} D_t$

\medskip
\promptpath{/role/project-planner}

You are the Project Planner for iteration \promptslot{loop\_index} of an
iterative software-development run. This is a planning-only harness
invocation. Do not implement, edit, test, or inspect production code. Return
only a prioritization overlay for the deterministic development document.

\medskip
\promptpath{/source-of-truth}

The public \promptslot{task\_source\_name} specification below is the complete
product source of truth. Do not use benchmark scores, hidden tests, private
rubrics, evaluator feedback, or other non-public information.

\promptslot{public\_task\_instruction}

\medskip
\promptpath{/previous-iteration-evidence}

\promptslot{evidence\_packet}

For iteration 1, there is no previous-iteration evidence. For later
iterations, identify verified functionality to preserve, visible bugs and
unmet requirements to repair, and evidence that remains insufficient. Do not
request or reconstruct the previous development document.

\medskip
\promptpath{/planning-policy}

Prioritize blockers and regressions before product extensions. Select at most
three achievable priorities already supported by the document scaffold.
Convert each priority into a concrete implementation target and an observable
validation requirement; avoid broad rewrites or unrelated architecture
changes.

\medskip
\promptpath{/document-scaffold}

\promptslot{scaffold\_document}

\medskip
\promptpath{/output-contract}

Return only the following Markdown structure:

\texttt{\#\# Project Planner Priorities}\par
\texttt{\#\#\# Priority Order}\par
\texttt{1. **Priority name** -- action and observable outcome}\par
\texttt{\#\#\# Preservation Gate}\par
\texttt{- Working functionality and evidence that must not regress}\par
\texttt{\#\#\# Acceptance Gate}\par
\texttt{- Smallest end-to-end validation for the selected priorities}
\end{plannerprompt}

\clearpage
\begin{developerprompt}{Developer Prompt}
\small
\textbf{Template assembly.}\quad
\promptmodule{benchmark guidance} $\oplus$
\promptmodule{iteration wrapper} $\oplus$
\promptmodule{warm-start block} $\oplus$
\promptmodule{development document}

\medskip
\textbf{Method correspondence.}\quad
$\left(A_{t-1};\mathcal{S},D_t\right)
\xrightarrow{\mathrm{Dev}_{H}} A_t$

\medskip
\promptpath{/role/developer}

You are the Developer for iteration \promptslot{loop\_index}. Build or improve
the complete project at \promptpath{/workspace/game}. Treat the public task
instruction as the PRD and the current development document as the
implementation and validation brief for this iteration.

\medskip
\promptpath{/iteration-context}

\texttt{\# Outer-loop attempt }\promptslot{attempt}

\promptmodule{if warm-started: \promptslot{warm\_start\_section}}

Continue from the artifact already present in \promptpath{/workspace/game}.
Preserve verified functionality and repair the next observable gap rather than
replacing a working project with a smaller reset.

\medskip
\promptpath{/development-document}

\texttt{Current document: }\promptslot{development\_doc\_filename}\par
\texttt{Planning inputs: }\promptslot{planning\_inputs}\par
\texttt{Build status: }\promptslot{build\_ok}\par
\texttt{Observed demos: }\promptslot{num\_demos}

\promptslot{public\_task\_instruction}

\promptslot{focus\_items}

\promptslot{preserve\_visible\_section}

\promptslot{development\_brief}

\promptslot{evidence\_history\_section}

\medskip
\promptpath{/development-policy}

Repair build and runtime blockers first, then address the ordered targets in
the document. Use the harness's native file, repository, shell, build,
execution, and local-testing tools. Keep the project launchable and make every
claimed mechanic observable through a valid replay trace. Public assets may be
read from \promptpath{/workspace/assets/library} and
\promptpath{/workspace/assets/library-oga}; copied assets count only when they
are visibly used by the artifact.

\medskip
\promptpath{/output-contract}

Leave the updated artifact in \promptpath{/workspace/game}, including
\promptpath{/workspace/game/project.godot}, a launchable main scene, and valid
\promptpath{/workspace/game/demo\_outputs/*.json} traces. Preserve the public
runtime records required for QA testing.
\end{developerprompt}

\begin{testprompt}{QA Tester Prompt}
\small
\textbf{Template assembly.}\quad
\promptmodule{tester role} $\oplus$
\promptmodule{phase instruction} $\oplus$
\promptmodule{development document} $\oplus$
\promptmodule{media manifest} $\oplus$
\promptmodule{phase-specific output contract}

\medskip
\textbf{Method correspondence.}\quad
$\left(A_t;\mathcal{S},D_t\right)
\xrightarrow{\mathrm{Test}_{H}} \mathcal{E}_t$

\medskip
\promptpath{/role/qa-tester}

You are the QA Tester for iteration \promptslot{loop\_index}. Review the
updated artifact as a player-facing product using only the public task text,
the current development document, visible project files, screenshots, replay
traces, and videos. Do not modify production code.

\promptslot{tester\_phase\_instruction}

\medskip
\promptpath{/inputs}

\texttt{Task directory: }\promptslot{task\_dir}\par
\texttt{Trial: }\promptslot{trial\_name}\par
\texttt{Trial directory: }\promptslot{trial\_dir}\par
\texttt{Development document: }\promptslot{development\_doc\_filename}

\medskip
\promptpath{/development-document}

\promptslot{development\_document}

\medskip
\promptpath{/public-visual-evidence}

\promptslot{media\_lines}

\medskip
\promptpath{/assessment-policy}

Derive checkable claims from the public requirements and validation targets.
Assign a claim--evidence record to the verified subset only when the cited
execution records provide sufficient observable support. Record visible
failures, regressions, unmet requirements, and insufficient evidence as gaps.
Inspect replay event types and reject traces that cannot be reproduced through
the public mouse, keyboard, and wait-event interface.

\medskip
\promptpath{/restrictions}

Do not read, list, infer from, or summarize \promptpath{/tests}, benchmark
scores, private evaluator files, hidden task metadata, formulas, or host-only
materials outside the public task and generated artifact.

\medskip
\promptpath{/output-contract}

Write \texttt{visual\_playtest\_report.md} and
\texttt{visual\_playtest\_report.json}. Each report records status, cited
evidence, player impact, issue ownership, and a concrete recommendation.

\promptmodule{next-loop phase: remaining bugs, preservation records, and
next-loop goals}

\promptslot{tester\_phase\_output\_contract}
\end{testprompt}

The QA prompt requests evidence-bearing findings in a benchmark-appropriate
JSON report. The report need not reproduce the mathematical tuple notation
verbatim. After the QA invocation, the benchmark adapter normalizes its claims,
cited execution records, and statuses into the evidence bundle
$\mathcal{E}_t$ used in the main paper. Thus, $\mathrm{Test}_{H}$ denotes the
complete testing interface, including both evidence collection by the QA Tester
and deterministic normalization of its report.

\subsection{Development Tools and Workspace Operations}

HoH does not replace the tools exposed by the underlying coding harness.
Instead, the Developer uses those tools under the current development document
and writes all changes to the same project workspace. The workspace contains
source code, configuration, project resources, and any public runtime artifacts
produced during development. Warm-starting therefore preserves not only source
files but also the project structure and resources required to continue
development from $A_{t-1}$.

Table~\ref{tab:development-tools} summarizes the principal capabilities used
by the implementation. The exact commands depend on the selected harness and
benchmark, but the role of each capability is fixed across iterations.

\begin{table}[H]
    \centering
    \caption{Development and inspection capabilities used by HoH. Private
    benchmark evaluators and their outputs are excluded from these interfaces.}
    \label{tab:development-tools}
    \small
    \renewcommand{\arraystretch}{1.16}
    \setlength{\tabcolsep}{5pt}
    \begin{tabular}{@{}L{0.22\textwidth}L{0.42\textwidth}L{0.27\textwidth}@{}}
        \toprule
        \textbf{Capability} & \textbf{Representative operations} &
        \textbf{Retained records} \\
        \midrule
        Project operations &
        Inspect and edit source files, configuration, assets, and repository
        state &
        Updated files and change state \\
        Build and execution &
        Invoke shell commands, build the project, launch the artifact, and run
        public local checks &
        Exit status, standard output, standard error, and runtime logs \\
        Game interaction &
        Launch Godot scenes, exercise controls, and execute deterministic
        interaction traces &
        Replay traces and observable state transitions \\
        Visual inspection &
        Capture screenshots or videos and inspect visible asset use, interface
        state, feedback, and result screens &
        Media manifest and referenced frames \\
        Task-specific tools &
        Use public benchmark containers, dependencies, and task-provided
        verification utilities where available &
        Public test and execution results \\
        \bottomrule
    \end{tabular}
\end{table}

For GameCraft-Bench, $A_t$ is a Godot project containing source scripts,
scenes, configuration, assets, and replay outputs. The benchmark adapter
materializes the current development document as additional context for the
Developer and records the harness command, process outcome, and resulting
trial. For FrontierSWE, the same interfaces operate on the task repository and
its public execution environment. Benchmark-specific adapters change how an
artifact is launched and observed; they do not change the planning, coding, or
testing roles.

\subsection{Evidence Collection and Representation}

The QA stage and benchmark adapter together convert observable behavior into
the claim--evidence records defined in the main paper. The correspondence is
\begin{equation}
\begin{aligned}
\mathcal{C}_t
&= \mathrm{Claims}(\mathcal{S},D_t),\\
r_i
&= \mathrm{Observe}(A_t,c_i),\\
s_i
&= \mathrm{Assess}(c_i,r_i),\\
\mathcal{E}_t
&= \left\{(c_i,r_i,s_i)\right\}_{c_i\in\mathcal{C}_t}.
\end{aligned}
\label{eq:appendix-evidence-process}
\end{equation}
Here, $\mathcal{C}_t$ contains the checkable claims, $r_i$ denotes the public
execution records collected for claim $c_i$, and $s_i$ is the normalized QA
status. Claims are instantiated from the public requirements, current
development targets, preservation constraints, and validation requirements.

Evidence collection first executes or inspects the artifact using the
capabilities above. Build and test outcomes establish whether the artifact can
run; runtime logs and traces expose state transitions; screenshots, videos, and
replays provide player-visible observations; and asset inspection determines
whether project resources are used in the executed artifact. Source-code
presence alone is not treated as behavioral verification.

The QA Tester then assesses every claim against its cited records. A record is
placed in $\mathcal{E}_t^{\mathrm{ver}}$ only when the evidence visibly supports
the corresponding claim. Observed failures, unmet requirements, regression
risks, and claims without sufficient evidence are placed in
$\mathcal{E}_t^{\mathrm{gap}}$. Formally, the two subsets are
\begin{equation}
\begin{aligned}
\mathcal{E}_t^{\mathrm{ver}}
&=
\left\{(c_i,r_i,s_i)\in\mathcal{E}_t
\mid s_i=\mathrm{verified}\right\},\\
\mathcal{E}_t^{\mathrm{gap}}
&=
\left\{(c_i,r_i,s_i)\in\mathcal{E}_t
\mid s_i=\mathrm{gap}\right\}.
\end{aligned}
\label{eq:appendix-evidence-subsets}
\end{equation}
They form a disjoint partition of the evidence bundle:
\begin{equation}
\mathcal{E}_t
=
\mathcal{E}_t^{\mathrm{ver}}
\cup
\mathcal{E}_t^{\mathrm{gap}},
\qquad
\mathcal{E}_t^{\mathrm{ver}}
\cap
\mathcal{E}_t^{\mathrm{gap}}
=\emptyset .
\label{eq:appendix-evidence-partition}
\end{equation}
The implementation retains both a human-readable tester report and structured
records for subsequent planning. Listing~\ref{lst:tester-output} shows a
normalized excerpt organized according to the verified- and gap-record
subsets above. Each record preserves the claim, the public execution records
used to assess it, and the resulting status. The final block shows how these
records are converted into planning inputs for the next iteration.

\begin{listing}[H]
\begin{lstlisting}
{
  "iteration": 2,
  "qa_status": "partial",
  "verified_records": [
    {
      "claim_id": "player_control",
      "claim": "Player input changes avatar motion.",
      "execution_records": [
        {
          "type": "replay",
          "path": "replays/core_loop.json",
          "observation": "Left and right inputs move the avatar."
        },
        {
          "type": "runtime_trace",
          "path": "traces/core_loop.json",
          "observation": "Position changes after each input event."
        }
      ],
      "status": "verified"
    }
  ],
  "gap_records": [
    {
      "claim_id": "result_state",
      "claim": "Completing the objective produces a visible result.",
      "execution_records": [
        {
          "type": "screenshot",
          "path": "screenshots/frame_018.png",
          "observation": "The objective ends without a result screen."
        }
      ],
      "status": "gap",
      "player_impact": "Completion is not visible to the player.",
      "recommended_update": "Add and replay a result state."
    }
  ],
  "planner_handoff": {
    "preservation_constraints": [
      "Preserve verified player movement."
    ],
    "update_targets": [
      "Implement a visible completion state."
    ],
    "validation_requirements": [
      "Replay objective completion through the result screen."
    ]
  }
}
\end{lstlisting}
\caption{Normalized structured evidence report. Each claim is linked to its
public execution records and QA status. Verified records yield preservation
constraints, whereas gap records yield update targets and follow-up validation
requirements for the next iteration.}
\label{lst:tester-output}
\end{listing}

For GameCraft-Bench, the evidence bundle is materialized through a screenshot
manifest, playtest report, structured status record, replay traces, and tester
logs. FrontierSWE uses the same claim--evidence abstraction with the public
execution and task-specific test records available in its repository
environment.

\subsection{Cross-Iteration State Transfer and Evaluation Isolation}

The implementation preserves the two state channels defined in the main paper.
The artifact channel carries the updated project $A_t$ into the next Developer
invocation, while the evidence channel carries $\mathcal{E}_t$ into the next
Planner invocation. Within iteration $t$, $D_t$ is the shared specification for
coding and testing; the next Planner constructs a new document from
$\mathcal{S}$ and $\mathcal{E}_t$ rather than treating $D_t$ as a third
persistent state channel.

The implementation retains the development document, harness command, process
logs, public media manifest, structured QA report, and resulting project
workspace for each iteration. These records make the transition from
$A_{t-1}$ to $A_t$ and the construction of $\mathcal{E}_t$ auditable without
exposing evaluator-only information.

Benchmark evaluation is separated from development. Hidden tests, benchmark
scores, private rubrics, evaluation formulas, and evaluator rationales are not
included in the role prompts or evidence bundle and are never returned to a
subsequent iteration. HoH therefore adapts to observable execution and QA
findings while the public task specification remains the authoritative
requirement source.

\clearpage
\section{Experimental Protocol}
\label{sec:supp-protocol}

\subsection{Harness and Model Configurations}

Table~\ref{tab:harness-model-configurations} lists the three configurations
used throughout the main experiments. Harness versions, models, and exposed
reasoning settings are held fixed between Vanilla and HoH within each
configuration. The main HoH results use $T=3$.

\begin{table}[H]
    \centering
    \caption{Harness--model configurations used in the experiments.}
    \label{tab:harness-model-configurations}
    \small
    \begin{tabular}{@{}llll@{}}
        \toprule
        \textbf{Harness} & \textbf{Version} & \textbf{Model} &
        \textbf{Reasoning setting} \\
        \midrule
        Codex CLI\footnotemark[1] & 0.142.5 & GPT-5.5 & High \\
        OpenCode\footnotemark[2] & 1.14.30 & DeepSeek-V4-Pro & --- \\
        Pi Coding Agent\footnotemark[3] & 0.80.10 & MiniMax-M3 & Client-side high \\
        \bottomrule
    \end{tabular}
\end{table}
\footnotetext[1]{\url{https://github.com/openai/codex}}
\footnotetext[2]{\url{https://github.com/anomalyco/opencode}}
\footnotetext[3]{\url{https://github.com/earendil-works/pi}}

\subsection{Run Configuration and Repetition}

The main experiments use three HoH iterations. Vanilla performs one standard
development pass, and the budget-controlled experiment additionally evaluates
two and three sequential Vanilla development passes. HoH reports the artifact
produced after the prescribed iteration budget; intermediate artifacts are
evaluated only for analysis and are not selected using benchmark scores.
Evaluator outputs are not returned to the planning, coding, or testing stages.

Table~\ref{tab:run-configurations} summarizes the run structure used for each
experiment. Every reported task--condition score is obtained from one valid
run. When an attempt fails because of an infrastructure or model-provider
transport error, the failed attempt is replaced rather than included as an
additional replicate. Aggregate scores therefore average over tasks, not over
multiple generations of the same task--condition pair.

\begin{table}[H]
    \centering
    \caption{Run configurations used in the reported experiments.}
    \label{tab:run-configurations}
    \small
    \setlength{\tabcolsep}{4pt}
    \begin{tabular}{@{}L{0.27\textwidth}L{0.25\textwidth}L{0.25\textwidth}L{0.15\textwidth}@{}}
        \toprule
        \textbf{Experiment} & \textbf{Benchmark} &
        \textbf{Development structure} & \textbf{Runs per task--condition} \\
        \midrule
        Main comparison & GameCraft-Bench and FrontierSWE &
        Vanilla: one coding pass; HoH: $T=3$ iterations & 1 \\
        \cmidrule(lr){1-4}
        Budget comparison & GameCraft-Bench &
        Vanilla Continuation: one, two, or three coding passes; HoH: $T=3$ & 1 \\
        \cmidrule(lr){1-4}
        Ablation study & GameCraft-Bench &
        Full HoH and each ablation: $T=3$ & 1 \\
        \bottomrule
    \end{tabular}
\end{table}

Task--condition runs start from separate copies of the benchmark-provided
workspace. Within a harness--model configuration, Vanilla and HoH use the same
model, native harness settings, public task materials, and benchmark tools.
The selected clients do not expose a common reproducible generation seed, and
we do not override temperature or top-$p$; the corresponding client and
provider defaults are used throughout. The fixed seeds reported below control
task sampling and statistical resampling rather than model generation.

\subsection{Computing Environments}

The two benchmarks use different execution environments because they exercise
different software artifacts. Table~\ref{tab:computing-environments} records
the shared runtime components. Model inference is provided through the remote
services associated with the configurations in
Table~\ref{tab:harness-model-configurations}; the listed machines execute the
harnesses, generated artifacts, and benchmark verifiers.

\begin{table}[H]
    \centering
    \caption{Computing environments used for artifact development and
    evaluation. FrontierSWE task images retain their benchmark-defined
    dependencies and resource declarations.}
    \label{tab:computing-environments}
    \small
    \setlength{\tabcolsep}{4pt}
    \begin{tabular}{@{}L{0.18\textwidth}L{0.37\textwidth}L{0.37\textwidth}@{}}
        \toprule
        \textbf{Component} & \textbf{GameCraft-Bench} & \textbf{FrontierSWE} \\
        \midrule
        Isolation &
        Run-local workspace in a local-subprocess environment &
        Official task container launched through a run-local Docker daemon \\
        \cmidrule(lr){1-3}
        Host &
        Ubuntu 24.04.3; Intel Core i7-14700; 64 GB RAM &
        Linux compute workers; NVIDIA H200 for tasks requiring a GPU \\
        \cmidrule(lr){1-3}
        Runtime &
        Python 3.12.3; Godot 4.6.2; Xvfb-backed display capture &
        Docker with the official task-specific software image \\
        \cmidrule(lr){1-3}
        Resource control &
        No task-specific GPU allocation &
        CPU, memory, storage, and GPU limits specified by each official task \\
        \bottomrule
    \end{tabular}
\end{table}

FrontierSWE uses a Docker-in-Docker execution design. An outer execution
container starts a run-local Docker daemon, which launches the official
task-specific image. The inner container receives the CPU, memory, storage,
and accelerator limits declared by that task. GPU passthrough is enabled only
for tasks that request an accelerator. This preserves the task software stack
and prevents dependencies from one task from affecting another.

\subsection{Benchmark Sampling and Evaluated Tasks}

Table~\ref{tab:benchmark-inventory} summarizes the benchmark subsets used in
the main experiments. GameCraft-Bench~\citep{luo2026gamecraftbench} is sampled
by its 15 public game families and is additionally organized into five coarse
reporting groups for analysis. FrontierSWE~\citep{proximal2026frontierswe}
uses the benchmark's three official categories.

\begin{table}[H]
    \centering
    \caption{Composition of the evaluated benchmark subsets. Counts refer to
    the tasks used for every harness--model configuration in the main
    experiments.}
    \label{tab:benchmark-inventory}
    \small
    \begin{tabular}{@{}lrrrl@{}}
        \toprule
        \textbf{Benchmark} & \textbf{Tasks} &
        \textbf{\shortstack{Fine\\categories}} &
        \textbf{\shortstack{Reporting\\groups}} &
        \textbf{\shortstack{Tasks per\\category}} \\
        \midrule
        GameCraft-Bench & 45 & 15 & 5 & 3 per family \\
        FrontierSWE & 15 & 3 & 3 & 4 / 9 / 2 \\
        \bottomrule
    \end{tabular}
\end{table}

\subsubsection{GameCraft-Bench}

We use a fixed 45-task subset with three tasks from each of the benchmark's 15
public game families. The subset was constructed incrementally from a fixed
earlier subset and completed to three tasks per family by seeded stratified
sampling (seed 20260707), without reference to model scores.

For compact reporting, we group the 15 families into five coarse categories,
each containing three families and nine tasks: Action contains Platformer,
Shooter, and Roguelike; Timing contains Racing, Rhythm, and Sports; Strategy
contains Strategy, Card Game, and Puzzle; Simulation contains Tycoon, Idle,
and Simulation; and Adventure contains Horror, Open World, and Visual Novel.
These five groups are introduced only for aggregate analysis; all task scores
continue to use the benchmark's original family definitions.

% Generated by supplementary/scripts/generate_result_tables.py.
\begingroup
\footnotesize
\renewcommand{\arraystretch}{1.12}
\begin{longtable}{@{}L{0.16\textwidth}L{0.24\textwidth}L{0.52\textwidth}@{}}
\caption{GameCraft-Bench reporting groups, benchmark families, and sampled tasks. Each family contributes three tasks.}
\label{tab:gamecraft-task-catalog}\\
\toprule
\rowcolor{headerrow}
\textbf{Family} & \textbf{Task} & \textbf{Benchmark identifier} \\
\midrule
\endfirsthead
\multicolumn{3}{l}{\small\itshape Table~\thetable\ (continued)}\\
\toprule
\rowcolor{headerrow}
\textbf{Family} & \textbf{Task} & \textbf{Benchmark identifier} \\
\midrule
\endhead
\rowcolor{actionrow}\multicolumn{3}{c}{\textbf{Action}} \\*
\multirow{3}{0.16\textwidth}{\textbf{Platformer}} & Momentum Lab & \path{platformer-momentum-lab} \\*
 & Ivory Beats & \path{platformer-ivory-beats} \\*
 & Thunder Valkyrie & \path{platformer-thunder-valkyrie} \\
\addlinespace[2pt]
\multirow{3}{0.16\textwidth}{\textbf{Shooter}} & Void Patrol & \path{shooter-void-patrol} \\*
 & Wave Commander & \path{shooter-wave-commander} \\*
 & Hotline Heist & \path{shooter-hotline-heist} \\
\addlinespace[2pt]
\multirow{3}{0.16\textwidth}{\textbf{Roguelike}} & Dungeon Shop & \path{roguelike-dungeon-shop} \\*
 & Breach Tactics & \path{roguelike-breach-tactics} \\*
 & Void Harvest & \path{roguelike-action-void-harvest} \\
\addlinespace[2pt]
\rowcolor{timingrow}\multicolumn{3}{c}{\textbf{Timing}} \\*
\multirow{3}{0.16\textwidth}{\textbf{Racing}} & Drift Circuit & \path{racing-drift-circuit} \\*
 & Rocket Trials & \path{racing-rocket-trials} \\*
 & Trick Runner & \path{racing-trick-runner} \\
\addlinespace[2pt]
\multirow{3}{0.16\textwidth}{\textbf{Rhythm}} & Note Highway & \path{rhythm-note-highway} \\*
 & Beat Dungeon & \path{rhythm-beat-dungeon} \\*
 & Garden & \path{rhythm-garden} \\
\addlinespace[2pt]
\multirow{3}{0.16\textwidth}{\textbf{Sports}} & Skateboard Park & \path{sports-skateboard-park} \\*
 & Boxing Gym & \path{sports-boxing-gym} \\*
 & Archery Quest & \path{sports-archery-quest} \\
\addlinespace[2pt]
\rowcolor{strategyrow}\multicolumn{3}{c}{\textbf{Strategy}} \\*
\multirow{3}{0.16\textwidth}{\textbf{Strategy}} & Tower Defense & \path{strategy-towerdefense} \\*
 & Chess Variant & \path{strategy-chess-variant} \\*
 & Spell Tactics & \path{strategy-spell-tactics} \\
\addlinespace[2pt]
\multirow{3}{0.16\textwidth}{\textbf{Card Game}} & Spire Descent & \path{cardgame-spire-descent} \\*
 & Poker Roguelike & \path{cardgame-poker-roguelike} \\*
 & Autobattler & \path{cardgame-autobattler} \\
\addlinespace[2pt]
\multirow{3}{0.16\textwidth}{\textbf{Puzzle}} & Sokoban Dungeon & \path{puzzle-sokoban-dungeon} \\*
 & Circuit Wizard & \path{puzzle-circuit-wizard} \\*
 & Pipe Crisis & \path{puzzle-pipe-crisis} \\
\addlinespace[2pt]
\rowcolor{simulationrow}\multicolumn{3}{c}{\textbf{Simulation}} \\*
\multirow{3}{0.16\textwidth}{\textbf{Tycoon}} & Space Colony & \path{tycoon-space-colony} \\*
 & Pirate Port & \path{tycoon-pirate-port} \\*
 & Wildhaven & \path{tycoon-wildhaven} \\
\addlinespace[2pt]
\multirow{3}{0.16\textwidth}{\textbf{Idle}} & Ant Empire & \path{idle-ant-empire} \\*
 & Factory Planet & \path{idle-factory-planet} \\*
 & Dungeon Guild & \path{idle-dungeon-guild} \\
\addlinespace[2pt]
\multirow{3}{0.16\textwidth}{\textbf{Simulation}} & Kitchen Rush & \path{simulation-kitchen-rush} \\*
 & Air Control & \path{simulation-air-control} \\*
 & Border Check & \path{simulation-border-check} \\
\addlinespace[2pt]
\rowcolor{adventurerow}\multicolumn{3}{c}{\textbf{Adventure}} \\*
\multirow{3}{0.16\textwidth}{\textbf{Horror}} & Floor 13 & \path{horror-floor-13} \\*
 & Dollhouse & \path{horror-dollhouse} \\*
 & Lighthouse & \path{horror-lighthouse} \\
\addlinespace[2pt]
\multirow{3}{0.16\textwidth}{\textbf{Open World}} & Sky Islands & \path{openworld-sky-islands} \\*
 & Airship Trader & \path{openworld-airship-trader} \\*
 & Bounty & \path{openworld-bounty} \\
\addlinespace[2pt]
\multirow{3}{0.16\textwidth}{\textbf{Visual Novel}} & Detective Noir & \path{visualnovel-detective-noir} \\*
 & Arcane Academy & \path{visualnovel-arcaneacademy} \\*
 & Time Paradox & \path{visualnovel-time-paradox} \\
\addlinespace[2pt]
\bottomrule
\end{longtable}
\endgroup

\subsubsection{FrontierSWE}

We evaluate 15 FrontierSWE tasks under the benchmark's official taxonomy:
four Implementation tasks, nine Performance tasks, and two Research tasks.
We additionally distinguish tasks that construct an independent deliverable
from a scaffold or task specification from those that optimize an existing
system. Under this criterion, 10 tasks are labeled end-to-end and five are
labeled optimization.

\begingroup
    \small
    \renewcommand{\arraystretch}{1.16}
    \setlength{\tabcolsep}{3pt}
    \begin{longtable}{@{}L{0.25\textwidth}>{\centering\arraybackslash}p{0.12\textwidth}L{0.55\textwidth}@{}}
        \caption{FrontierSWE tasks grouped by official category and construction scope.}
        \label{tab:frontierswe-task-categories}\\
        \toprule
        \rowcolor{headerrow}
        \textbf{Task} & \textbf{Scope} & \textbf{Brief task description} \\
        \midrule
        \endfirsthead
        \multicolumn{3}{l}{\small\itshape Table~\thetable\ (continued)}\\
        \toprule
        \rowcolor{headerrow}
        \textbf{Task} & \textbf{Scope} & \textbf{Brief task description} \\
        \midrule
        \endhead
        \rowcolor{implementationrow}\multicolumn{3}{c}{\textbf{Implementation}} \\*
        Dart Style Haskell & End-to-end &
        Reimplement the Dart formatter in Haskell as a Cabal-built executable
        compatible with the relevant CLI behavior and golden formatting cases. \\
        Git to Zig & End-to-end &
        Reimplement Git~2.47 as a Zig binary compatible with Git's CLI,
        output, and exit-code behavior, without reusing the existing Git
        implementation or network access. \\
        Lua Native Compiler & End-to-end &
        Compile Lua~5.4 bytecode to a standalone native x86-64 executable with
        reference-equivalent output, rather than an interpreter or API wrapper. \\
        PostgreSQL--SQLite Wire Adapter & End-to-end &
        Build a Zig server backed by SQLite that emulates the required
        PostgreSQL server, wire-protocol, lifecycle, and CLI behavior. \\
        \rowcolor{performancerow}\multicolumn{3}{c}{\textbf{Performance}} \\*
        Cranelift Codegen Optimization & Optimization &
        Optimize compiled WebAssembly runtime performance in Wasmtime's
        Cranelift backend, subject to correctness gates and weighted speedup
        scoring. \\
        Dependent Type Checker & End-to-end &
        Implement a correct, high-throughput Martin-L{\"o}f type checker in
        Rust; correctness thresholds must be met before throughput is scored. \\
        FFmpeg Swscale Rewrite & End-to-end &
        Rewrite \texttt{libswscale} in Zig or Rust behind its required C ABI,
        with image-quality gates before geometric-mean speedup scoring. \\
        Granite Mamba2 Inference Optimization & Optimization &
        Optimize a standalone Granite Mamba2 layer while preserving CUDA
        bfloat16 outputs and cache behavior across the evaluated workloads. \\
        Inference System Optimization & Optimization &
        Accelerate a Qwen-based SGLang serving system while preserving
        token-level output equivalence under latency and throughput workloads. \\
        Libexpat to x86 Assembly & End-to-end &
        Reimplement the required \texttt{libexpat} API as an independent
        x86-64 assembly shared library without delegating to the existing
        implementation. \\
        Notebook Compression & End-to-end &
        Build a lossless domain-specific notebook compressor with \texttt{fit},
        \texttt{compress}, and \texttt{decompress} interfaces; exact recovery
        is required before compression ratio is scored. \\
        Pyright Type-Checking Optimization & Optimization &
        Optimize Pyright's type-evaluation hot paths while preserving build
        success, all required tests, and reference-equivalent diagnostics. \\
        Revideo Performance Optimization & Optimization &
        Optimize Revideo's programmatic rendering pipeline without frame
        skipping, quality reduction, resolution changes, or visible-output
        deviations. \\
        \rowcolor{researchrow}\multicolumn{3}{c}{\textbf{Research}} \\*
        Optimizer Design & End-to-end &
        Implement one \texttt{torch.optim.Optimizer} and a shared
        hyperparameter configuration that generalizes across heterogeneous ML
        workloads. \\
        PCQM4Mv2 Autoresearch & End-to-end &
        Train a 2D molecular-graph regressor under data, model, and parameter
        constraints to minimize the evaluated molecular-property error. \\
        \bottomrule
\end{longtable}
\endgroup

Two official FrontierSWE tasks are not included in the evaluated subset. Their
omission is determined by execution requirements rather than model outcomes.

\begin{table}[H]
    \centering
    \caption{FrontierSWE tasks excluded from the evaluated subset.}
    \label{tab:frontierswe-excluded-tasks}
    \small
    \setlength{\tabcolsep}{5pt}
    \begin{tabular}{@{}L{0.27\textwidth}L{0.65\textwidth}@{}}
        \toprule
        \textbf{Task} & \textbf{Reason} \\
        \midrule
        \texttt{frogsgame-rl} &
        Requires authenticated access to the external Tinker API, which was
        unavailable in the evaluation environment. \\
        \cmidrule(lr){1-2}
        \texttt{modular-stack-wan21} &
        Requires a Modular MAX software stack needing an NVIDIA driver of at
        least 580 (CUDA 13), whereas the available H200 worker used driver
        570.133.20. \\
        \bottomrule
    \end{tabular}
\end{table}

\subsection{Baseline and Budget-Controlled Protocols}
\label{sec:supp-budget-protocol}

\paragraph{Vanilla.}
Vanilla uses the corresponding harness--model configuration without the HoH
protocol and performs one standard development pass from the
benchmark-provided initial artifact.

\paragraph{Vanilla Continuation.}
The budget-controlled comparison extends the selected Vanilla artifact through
two additional invocations of the same harness--model configuration. Let
$A_1^{\mathrm{VC}}$ denote the artifact produced by the standard Vanilla pass.
For $k\in\{2,3\}$, Vanilla Continuation applies
\begin{equation}
A_k^{\mathrm{VC}}
=
\mathrm{Dev}_{H}
\left(A_{k-1}^{\mathrm{VC}};\mathcal{S},p_{\mathrm{cont}}\right),
\label{eq:vanilla-continuation}
\end{equation}
where $p_{\mathrm{cont}}$ is the fixed instruction shown below. Thus, each
additional pass starts from the latest artifact, but receives neither a
development document nor evidence from a separate QA Tester invocation.

\begin{controlprompt}{Vanilla Continuation Prompt}
\small
Continue developing and testing the current game.
\end{controlprompt}

The comparison therefore holds the initial task set and harness--model
configuration fixed while separating repeated coding passes from the
planning--coding--testing structure of HoH.

\subsection{Ablation Protocols}
\label{sec:supp-ablation-protocol}

The ablations retain the three-iteration budget and the same harness--model
configuration as full HoH. Relative to the full iteration in
Eq.~\ref{eq:supp-hoh-iteration}, each variant changes one cross-iteration
input while leaving the remaining interfaces unchanged:
\begin{equation}
\begin{aligned}
\text{\emph{w/o Plan Update}:}\quad
    &D_t = D_1,\qquad t>1,\\
\text{\emph{w/o Evidence Feedback}:}\quad
    &D_t = \mathrm{Plan}_{H}\left(\mathcal{S},\emptyset\right),\\
\text{\emph{w/o Warm-Start}:}\quad
    &A_t = \mathrm{Dev}_{H}\left(A_0;\mathcal{S},D_t\right).
\end{aligned}
\label{eq:ablation-interventions}
\end{equation}
In \emph{w/o Plan Update}, the first development document is reused in all
later iterations, although coding and testing continue on the evolving
artifact. In \emph{w/o Evidence Feedback}, the QA Tester still evaluates each
updated artifact, but its evidence is withheld from the next Planner
invocation. In \emph{w/o Warm-Start}, evidence-conditioned planning and QA
testing remain active, while every Developer invocation begins from the
benchmark-provided initial artifact $A_0$.

\begin{table}[H]
    \centering
    \caption{Information channels retained by the ablation variants.}
    \label{tab:ablation-protocols}
    \small
    \setlength{\tabcolsep}{5pt}
    \begin{tabular}{@{}L{0.25\textwidth}L{0.23\textwidth}L{0.20\textwidth}L{0.22\textwidth}@{}}
        \toprule
        \textbf{Variant} & \textbf{Development document} &
        \textbf{Coding start} & \textbf{Evidence for next plan} \\
        \midrule
        Full HoH &
        Updated from preceding evidence &
        $A_{t-1}$ &
        $\mathcal{E}_{t-1}$ \\
        w/o Plan Update &
        Fixed to $D_1$ &
        $A_{t-1}$ &
        Not consumed \\
        w/o Evidence Feedback &
        Updated from $\mathcal{S}$ only &
        $A_{t-1}$ &
        Withheld \\
        w/o Warm-Start &
        Updated from preceding evidence &
        $A_0$ &
        $\mathcal{E}_{t-1}$ \\
        \bottomrule
    \end{tabular}
\end{table}

\subsection{Metrics and Resource Accounting}

\paragraph{GameCraft-Bench dimensions.}
GameCraft-Bench evaluates runnable game artifacts along four dimensions.
\emph{Core Mechanics} measures implementation of the required gameplay
mechanics and interaction loop. \emph{Content Depth} measures the breadth and
variety of stages, challenges, objectives, and progression. \emph{Functional
Visuals} measures the visibility, readability, and feedback of gameplay
states. \emph{Art and Presentation} measures visual coherence, asset quality,
interface styling, and polish. Let $M$, $D$, $V$, and $A$ denote the mean
rubric-item scores for these four dimensions, respectively.

\paragraph{GameCraft-Bench Overall.}
The benchmark combines the four dimensions as
\begin{equation}
\operatorname{Overall}
=100B\left(0.15M+0.35D+0.15V+0.35A\right),
\label{eq:gamecraft-overall}
\end{equation}
where $B=1$ if the game artifact compiles and runs and $B=0$ otherwise.

\paragraph{FrontierSWE reward.}
We report the task-specific official reward and its mean over the 15 evaluated
tasks.

\paragraph{Task-level aggregation.}
For a benchmark task set $\mathcal{B}$ and evaluated condition $c$, the
reported aggregate is the unweighted mean of its task-level scores:
\begin{equation}
\overline{s}_{\mathcal{B}}(c)
=
\frac{1}{|\mathcal{B}|}
\sum_{i\in\mathcal{B}} s_i(c).
\label{eq:supp-task-mean}
\end{equation}
For GameCraft-Bench, $s_i$ is the 0--100 Overall score and
$|\mathcal{B}|=45$; for FrontierSWE, $s_i$ is the official reward and
$|\mathcal{B}|=15$.

\paragraph{Bootstrap uncertainty.}
The 95\% confidence intervals for the GameCraft-Bench component analysis in
the main paper are percentile intervals from 20,000 task-bootstrap resamples.
Tasks are sampled with replacement, all four component scores for a sampled
task are retained together, and means are recomputed for every resample. The
bootstrap uses seed 20260729.

\paragraph{Model-interaction volume.}
Token counts include the provider-reported input and output tokens from
coding-harness model calls and exclude benchmark evaluation. Input totals may
include cached context reads, whose accounting differs across providers.
Let $\mathcal{I}_i(c)$ denote the model calls made for task $i$ under condition
$c$. Cumulative token use, in millions of tokens, is
\begin{equation}
C_i(c)
=
10^{-6}\sum_{j\in\mathcal{I}_i(c)}
\left(n^{\mathrm{in}}_j+n^{\mathrm{out}}_j\right),
\qquad
\overline{C}(c)
=
\frac{1}{|\mathcal{B}|}\sum_{i\in\mathcal{B}} C_i(c).
\label{eq:supp-token-accounting}
\end{equation}
We compare token usage within each harness--model configuration because cache
accounting differs across providers. In the budget-controlled comparison, we
also report the quality gained per additional million tokens relative to
Vanilla:
\begin{equation}
\eta(c)
=
\frac{\overline{s}_{\mathrm{GC}}(c)
-\overline{s}_{\mathrm{GC}}(\mathrm{Vanilla})}
{\overline{C}(c)-\overline{C}(\mathrm{Vanilla})}.
\label{eq:supp-budget-efficiency}
\end{equation}

\newpage
\paragraph{Official dominance.}
We apply the official dominance procedure to the final 15-task subset.  Within
each domain, the comparison pool contains all $3\times4=12$
system--condition configurations: the three systems under Vanilla, HoH@1,
HoH@2, and HoH@3.  For a configuration $a$, let $d$ denote a domain, let $t$
denote a task in that domain, and let $r_{a,t}$ denote $a$'s official reward
on $t$.

All pairwise comparisons occur on the same task.  For any opponent $j$ among
the other 11 configurations, the comparison score is
\begin{equation}
s(x,y)=
\begin{cases}
1, & x>y,\\
0.5, & x=y,\\
0, & x<y.
\end{cases}
\label{eq:appendix-dominance-score}
\end{equation}
The task-level dominance of $a$ is therefore
\begin{equation}
\operatorname{Dominance}_{d,t}(a)
=
\frac{1}{11}\sum_{j\ne a}s(r_{a,t},r_{j,t}).
\label{eq:appendix-task-dominance}
\end{equation}
Equivalently, this quantity is the expected comparison score when the opponent
is selected uniformly from the other 11 configurations. The implementation
computes this expectation exactly by averaging over all 11 opponents.
The denominator is 11 because $a$ is compared with every other member of the
12-configuration pool, but not with itself.  Domain-level dominance averages
these values equally over the $N_d$ tasks in domain $d$:
\begin{equation}
\operatorname{Dominance}_{d}(a)
=
\frac{1}{N_d}\sum_{t=1}^{N_d}
\left[\frac{1}{11}\sum_{j\ne a}s(r_{a,t},r_{j,t})\right].
\label{eq:appendix-domain-dominance}
\end{equation}
Here, $N_d$ is 4, 9, and 2 for Implementation, Performance, and Research,
respectively.  The reported FrontierSWE dominance is the macro average over
the three domains:
\begin{equation}
\operatorname{Dominance}(a)
=
\frac{1}{3}
\sum_d
\operatorname{Dominance}_{d}(a).
\label{eq:appendix-overall-dominance}
\end{equation}
In this expression, $d$ ranges over Implementation, Performance, and Research,
and each domain receives equal weight.

\clearpage
\subsection{Player-Experience Evaluation and PXI Aggregation}
\label{sec:supp-pxi-scoring}

The source-blinded Fusepoint playtest uses the full Player Experience
Inventory (PXI)~\citep{vandenabeele2020pxi}. The validated core comprises ten
constructs, each measured by three items on the official seven-point scale
from $-3$ to $+3$. For evaluator $p$, construct $k$, and its three item
responses $x_{p,k,j}$, we compute
\begin{equation}
s_{p,k}=\frac{1}{3}\sum_{j=1}^{3}x_{p,k,j}.
\label{eq:supp-pxi-construct-score}
\end{equation}
The official questionnaire's separate three-item Enjoyment outcome is scored
in the same way but is not treated as an eleventh core PXI construct. For each
reported outcome, the main-paper table gives the mean and sample standard
deviation of $s_{p,k}$ across evaluators; individual ratings and comments are
retained for auditability.

We do not compute a global PXI total. A review conducted during an independent
validation found that some prior applications averaged the ten, or sometimes
eleven, outcomes into a single general player-experience score. However, the
preregistered validation with 1,518 players found better fit for the ten-factor
model---or the eleven-factor model when Enjoyment is included---than for models
with a general player-experience factor or higher-order consequence factors
\citep{perrig2024pxi}. We therefore interpret the constructs separately. The
main-paper table additionally reports unweighted descriptive averages over the
five Functional and five Psychosocial construct scores for compact summary;
these averages are not treated as validated higher-order PXI scales. No score
combining all ten constructs, sum-score, percentage conversion, or cutoff is
reported as a PXI total.

\subsection{Reproducibility Artifacts}

The anonymous code package accompanying the submission contains the core HoH
implementation, role prompt templates, the GameCraft-Bench adapter, and the
necessary wrappers for the evaluated harness--model configurations. Benchmark
repositories, task data, raw run artifacts, analysis records, environment
files, private credentials, provider secrets, and benchmark-hidden evaluator
contents are not included.

\clearpage
\section{Complete Experimental Results}
\label{sec:supp-results}

\subsection{GameCraft-Bench Per-Task Scores}

\begin{figure}[H]
    \centering
    \includegraphics[width=\textwidth]{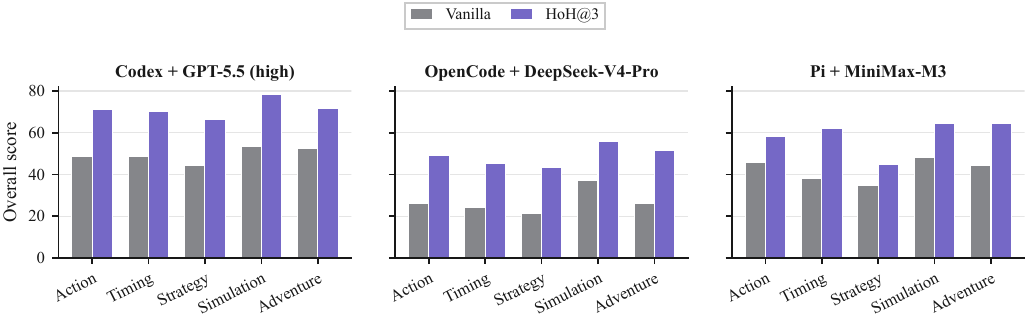}
    \caption{GameCraft-Bench Overall scores by reporting group and
    harness--model configuration. Each group contains nine tasks.}
    \label{fig:supp-gamecraft-categories}
\end{figure}

Figure~\ref{fig:supp-gamecraft-categories} summarizes Vanilla and HoH@3 over
the five reporting groups before the complete task-level results. Each bar is
the unweighted mean of the nine tasks in that group.

Tables~\ref{tab:gamecraft-task-scores-1}--\ref{tab:gamecraft-task-scores-3}
report the four observed conditions for every sampled GameCraft-Bench task.
Scores are converted to the benchmark's 0--100 presentation scale and grouped
using the five coarse categories defined in
Table~\ref{tab:gamecraft-task-catalog}. The final column reports the
task-specific change from Vanilla to HoH@3.

% Generated from the frozen GameCraft score CSV files.
\begingroup
\footnotesize
\begin{longtable}{@{}L{0.15\textwidth}L{0.24\textwidth}rrrrr@{}}
\caption{Complete GameCraft-Bench task scores for Codex + GPT-5.5 (high). Scores use the benchmark's 0--100 scale; $\Delta$ denotes HoH@3 minus Vanilla.}
\label{tab:gamecraft-task-scores-1}\\
\toprule
\rowcolor{headerrow}
\textbf{Family} & \textbf{Task} & \textbf{Vanilla} & \textbf{HoH@1} & \textbf{HoH@2} & \cellcolor{finalcol}\textbf{HoH@3} & \textbf{$\Delta$} \\
\midrule
\endfirsthead
\multicolumn{7}{l}{\small\itshape Table~\thetable\ (continued)}\\
\toprule
\rowcolor{headerrow}
\textbf{Family} & \textbf{Task} & \textbf{Vanilla} & \textbf{HoH@1} & \textbf{HoH@2} & \cellcolor{finalcol}\textbf{HoH@3} & \textbf{$\Delta$} \\
\midrule
\endhead
\rowcolor{actionrow}\multicolumn{7}{c}{\textbf{Action}} \\*
Platformer & Momentum Lab & 34.05 & 64.44 & 56.15 & \cellcolor{finalcol}70.61 & +36.56 \\
 & Ivory Beats & 46.81 & 73.28 & 78.55 & \cellcolor{finalcol}85.42 & +38.61 \\
 & Thunder Valkyrie & 53.51 & 68.24 & 71.82 & \cellcolor{finalcol}73.67 & +20.15 \\
\addlinespace[1pt]
Shooter & Void Patrol & 58.59 & 73.90 & 76.02 & \cellcolor{finalcol}87.83 & +29.23 \\
 & Wave Commander & 66.97 & 69.35 & 74.64 & \cellcolor{finalcol}75.25 & +8.28 \\
 & Hotline Heist & 43.37 & 35.71 & 59.17 & \cellcolor{finalcol}62.81 & +19.44 \\
\addlinespace[1pt]
Roguelike & Dungeon Shop & 40.89 & 50.55 & 50.15 & \cellcolor{finalcol}65.09 & +24.20 \\
 & Breach Tactics & 53.44 & 55.67 & 57.11 & \cellcolor{finalcol}61.09 & +7.65 \\
 & Void Harvest & 41.05 & 46.39 & 55.43 & \cellcolor{finalcol}57.43 & +16.38 \\
\addlinespace[1pt]
\rowcolor{timingrow}\multicolumn{7}{c}{\textbf{Timing}} \\*
Racing & Drift Circuit & 43.58 & 58.29 & 60.29 & \cellcolor{finalcol}70.08 & +26.50 \\
 & Rocket Trials & 45.19 & 61.32 & 68.00 & \cellcolor{finalcol}70.31 & +25.12 \\
 & Trick Runner & 51.50 & 36.09 & 54.44 & \cellcolor{finalcol}64.78 & +13.28 \\
\addlinespace[1pt]
Rhythm & Note Highway & 39.44 & 31.35 & 56.15 & \cellcolor{finalcol}64.68 & +25.24 \\
 & Beat Dungeon & 47.94 & 56.95 & 60.55 & \cellcolor{finalcol}60.81 & +12.88 \\
 & Garden & 52.75 & 65.33 & 69.01 & \cellcolor{finalcol}70.69 & +17.94 \\
\addlinespace[1pt]
Sports & Skateboard Park & 57.64 & 69.51 & 72.62 & \cellcolor{finalcol}81.14 & +23.49 \\
 & Boxing Gym & 43.08 & 40.34 & 45.61 & \cellcolor{finalcol}73.14 & +30.06 \\
 & Archery Quest & 58.09 & 64.96 & 71.56 & \cellcolor{finalcol}76.69 & +18.60 \\
\addlinespace[1pt]
\rowcolor{strategyrow}\multicolumn{7}{c}{\textbf{Strategy}} \\*
Strategy & Tower Defense & 58.85 & 63.98 & 64.12 & \cellcolor{finalcol}76.92 & +18.07 \\
 & Chess Variant & 30.04 & 56.93 & 52.83 & \cellcolor{finalcol}59.72 & +29.68 \\
 & Spell Tactics & 53.10 & 47.39 & 60.70 & \cellcolor{finalcol}63.32 & +10.22 \\
\addlinespace[1pt]
Card Game & Spire Descent & 25.85 & 45.47 & 54.81 & \cellcolor{finalcol}61.92 & +36.07 \\
 & Poker Roguelike & 44.66 & 63.94 & 67.99 & \cellcolor{finalcol}69.27 & +24.61 \\
 & Autobattler & 55.06 & 72.39 & 71.28 & \cellcolor{finalcol}72.22 & +17.16 \\
\addlinespace[1pt]
Puzzle & Sokoban Dungeon & 55.75 & 58.98 & 58.56 & \cellcolor{finalcol}70.13 & +14.38 \\
 & Circuit Wizard & 31.35 & 38.97 & 40.82 & \cellcolor{finalcol}49.52 & +18.17 \\
 & Pipe Crisis & 41.88 & 65.07 & 68.65 & \cellcolor{finalcol}72.17 & +30.28 \\
\addlinespace[1pt]
\rowcolor{simulationrow}\multicolumn{7}{c}{\textbf{Simulation}} \\*
Tycoon & Space Colony & 36.88 & 64.60 & 69.15 & \cellcolor{finalcol}76.55 & +39.67 \\
 & Pirate Port & 51.32 & 63.06 & 70.75 & \cellcolor{finalcol}77.81 & +26.48 \\
 & Wildhaven & 49.43 & 74.09 & 77.78 & \cellcolor{finalcol}80.39 & +30.96 \\
\addlinespace[1pt]
Idle & Ant Empire & 65.52 & 71.42 & 69.64 & \cellcolor{finalcol}87.88 & +22.36 \\
 & Factory Planet & 75.16 & 78.94 & 82.27 & \cellcolor{finalcol}87.78 & +12.63 \\
 & Dungeon Guild & 63.92 & 68.94 & 76.95 & \cellcolor{finalcol}79.50 & +15.57 \\
\addlinespace[1pt]
Simulation & Kitchen Rush & 42.62 & 48.07 & 65.64 & \cellcolor{finalcol}73.38 & +30.75 \\
 & Air Control & 54.24 & 63.96 & 66.72 & \cellcolor{finalcol}69.73 & +15.50 \\
 & Border Check & 43.58 & 67.70 & 67.00 & \cellcolor{finalcol}72.74 & +29.17 \\
\addlinespace[1pt]
\rowcolor{adventurerow}\multicolumn{7}{c}{\textbf{Adventure}} \\*
Horror & Floor 13 & 48.68 & 70.73 & 68.56 & \cellcolor{finalcol}73.41 & +24.73 \\
 & Dollhouse & 56.33 & 64.28 & 66.93 & \cellcolor{finalcol}72.59 & +16.27 \\
 & Lighthouse & 51.95 & 63.89 & 71.47 & \cellcolor{finalcol}80.90 & +28.95 \\
\addlinespace[1pt]
Open World & Sky Islands & 53.25 & 47.19 & 60.91 & \cellcolor{finalcol}68.25 & +15.00 \\
 & Airship Trader & 59.58 & 73.60 & 74.40 & \cellcolor{finalcol}74.41 & +14.83 \\
 & Bounty & 47.00 & 67.47 & 63.31 & \cellcolor{finalcol}68.27 & +21.27 \\
\addlinespace[1pt]
Visual Novel & Detective Noir & 47.46 & 63.15 & 55.54 & \cellcolor{finalcol}65.31 & +17.85 \\
 & Arcane Academy & 59.30 & 37.60 & 62.71 & \cellcolor{finalcol}70.69 & +11.39 \\
 & Time Paradox & 50.63 & 63.28 & 71.15 & \cellcolor{finalcol}71.97 & +21.34 \\
\addlinespace[1pt]
\midrule
\multicolumn{2}{l}{\textbf{Mean}} & 49.58 & 59.71 & 64.84 & \cellcolor{finalcol}\textbf{71.52} & \textbf{+21.93} \\
\bottomrule
\end{longtable}
\endgroup

\begingroup
\footnotesize
\begin{longtable}{@{}L{0.15\textwidth}L{0.24\textwidth}rrrrr@{}}
\caption{Complete GameCraft-Bench task scores for OpenCode + DeepSeek-V4-Pro. Scores use the benchmark's 0--100 scale; $\Delta$ denotes HoH@3 minus Vanilla.}
\label{tab:gamecraft-task-scores-2}\\
\toprule
\rowcolor{headerrow}
\textbf{Family} & \textbf{Task} & \textbf{Vanilla} & \textbf{HoH@1} & \textbf{HoH@2} & \cellcolor{finalcol}\textbf{HoH@3} & \textbf{$\Delta$} \\
\midrule
\endfirsthead
\multicolumn{7}{l}{\small\itshape Table~\thetable\ (continued)}\\
\toprule
\rowcolor{headerrow}
\textbf{Family} & \textbf{Task} & \textbf{Vanilla} & \textbf{HoH@1} & \textbf{HoH@2} & \cellcolor{finalcol}\textbf{HoH@3} & \textbf{$\Delta$} \\
\midrule
\endhead
\rowcolor{actionrow}\multicolumn{7}{c}{\textbf{Action}} \\*
Platformer & Momentum Lab & 14.33 & 23.13 & 32.92 & \cellcolor{finalcol}30.75 & +16.42 \\
 & Ivory Beats & 31.73 & 45.78 & 42.09 & \cellcolor{finalcol}61.08 & +29.35 \\
 & Thunder Valkyrie & 19.37 & 28.65 & 52.20 & \cellcolor{finalcol}60.54 & +41.17 \\
\addlinespace[1pt]
Shooter & Void Patrol & 37.12 & 57.44 & 53.49 & \cellcolor{finalcol}57.77 & +20.66 \\
 & Wave Commander & 14.74 & 2.31 & 35.61 & \cellcolor{finalcol}46.46 & +31.71 \\
 & Hotline Heist & 33.85 & 12.78 & 33.42 & \cellcolor{finalcol}38.05 & +4.20 \\
\addlinespace[1pt]
Roguelike & Dungeon Shop & 39.98 & 41.52 & 53.22 & \cellcolor{finalcol}46.50 & +6.52 \\
 & Breach Tactics & 35.28 & 23.77 & 36.76 & \cellcolor{finalcol}46.99 & +11.71 \\
 & Void Harvest & 9.50 & 14.34 & 49.29 & \cellcolor{finalcol}52.83 & +43.33 \\
\addlinespace[1pt]
\rowcolor{timingrow}\multicolumn{7}{c}{\textbf{Timing}} \\*
Racing & Drift Circuit & 37.87 & 31.48 & 39.17 & \cellcolor{finalcol}38.69 & +0.82 \\
 & Rocket Trials & 2.41 & 3.02 & 19.90 & \cellcolor{finalcol}22.68 & +20.27 \\
 & Trick Runner & 14.37 & 19.62 & 25.26 & \cellcolor{finalcol}39.24 & +24.86 \\
\addlinespace[1pt]
Rhythm & Note Highway & 25.25 & 20.88 & 29.61 & \cellcolor{finalcol}40.34 & +15.09 \\
 & Beat Dungeon & 3.06 & 18.60 & 17.05 & \cellcolor{finalcol}23.81 & +20.74 \\
 & Garden & 49.29 & 58.11 & 55.21 & \cellcolor{finalcol}63.74 & +14.44 \\
\addlinespace[1pt]
Sports & Skateboard Park & 19.28 & 24.69 & 46.28 & \cellcolor{finalcol}57.60 & +38.32 \\
 & Boxing Gym & 31.31 & 40.69 & 42.25 & \cellcolor{finalcol}55.75 & +24.44 \\
 & Archery Quest & 33.57 & 29.50 & 54.31 & \cellcolor{finalcol}63.62 & +30.06 \\
\addlinespace[1pt]
\rowcolor{strategyrow}\multicolumn{7}{c}{\textbf{Strategy}} \\*
Strategy & Tower Defense & 29.47 & 25.89 & 39.18 & \cellcolor{finalcol}49.54 & +20.07 \\
 & Chess Variant & 18.76 & 23.69 & 40.45 & \cellcolor{finalcol}38.51 & +19.75 \\
 & Spell Tactics & 15.03 & 35.17 & 21.57 & \cellcolor{finalcol}42.05 & +27.02 \\
\addlinespace[1pt]
Card Game & Spire Descent & 13.33 & 4.38 & 19.47 & \cellcolor{finalcol}23.00 & +9.67 \\
 & Poker Roguelike & 20.50 & 24.89 & 38.77 & \cellcolor{finalcol}57.90 & +37.40 \\
 & Autobattler & 30.76 & 18.75 & 23.12 & \cellcolor{finalcol}18.65 & -12.12 \\
\addlinespace[1pt]
Puzzle & Sokoban Dungeon & 34.27 & 33.67 & 51.95 & \cellcolor{finalcol}51.67 & +17.40 \\
 & Circuit Wizard & 3.90 & 14.56 & 3.90 & \cellcolor{finalcol}35.84 & +31.95 \\
 & Pipe Crisis & 25.39 & 14.55 & 63.17 & \cellcolor{finalcol}72.90 & +47.50 \\
\addlinespace[1pt]
\rowcolor{simulationrow}\multicolumn{7}{c}{\textbf{Simulation}} \\*
Tycoon & Space Colony & 34.24 & 47.98 & 57.87 & \cellcolor{finalcol}60.27 & +26.03 \\
 & Pirate Port & 25.89 & 61.33 & 52.00 & \cellcolor{finalcol}52.16 & +26.26 \\
 & Wildhaven & 47.66 & 46.52 & 55.13 & \cellcolor{finalcol}66.07 & +18.41 \\
\addlinespace[1pt]
Idle & Ant Empire & 40.73 & 42.64 & 55.49 & \cellcolor{finalcol}49.08 & +8.36 \\
 & Factory Planet & 38.17 & 35.20 & 41.02 & \cellcolor{finalcol}67.84 & +29.66 \\
 & Dungeon Guild & 15.66 & 32.89 & 69.71 & \cellcolor{finalcol}61.34 & +45.67 \\
\addlinespace[1pt]
Simulation & Kitchen Rush & 33.31 & 33.41 & 36.44 & \cellcolor{finalcol}39.96 & +6.64 \\
 & Air Control & 28.23 & 35.40 & 33.49 & \cellcolor{finalcol}35.30 & +7.07 \\
 & Border Check & 70.22 & 58.34 & 69.22 & \cellcolor{finalcol}70.74 & +0.52 \\
\addlinespace[1pt]
\rowcolor{adventurerow}\multicolumn{7}{c}{\textbf{Adventure}} \\*
Horror & Floor 13 & 29.22 & 14.56 & 41.53 & \cellcolor{finalcol}42.12 & +12.90 \\
 & Dollhouse & 8.75 & 10.95 & 13.00 & \cellcolor{finalcol}54.39 & +45.64 \\
 & Lighthouse & 52.11 & 35.78 & 46.30 & \cellcolor{finalcol}83.33 & +31.23 \\
\addlinespace[1pt]
Open World & Sky Islands & 24.37 & 11.87 & 28.69 & \cellcolor{finalcol}46.51 & +22.15 \\
 & Airship Trader & 38.94 & 35.41 & 54.64 & \cellcolor{finalcol}62.37 & +23.42 \\
 & Bounty & 7.09 & 13.66 & 0.73 & \cellcolor{finalcol}25.35 & +18.26 \\
\addlinespace[1pt]
Visual Novel & Detective Noir & 22.22 & 18.06 & 56.59 & \cellcolor{finalcol}61.57 & +39.35 \\
 & Arcane Academy & 28.76 & 26.72 & 51.36 & \cellcolor{finalcol}49.05 & +20.29 \\
 & Time Paradox & 21.13 & 34.96 & 31.62 & \cellcolor{finalcol}40.04 & +18.91 \\
\addlinespace[1pt]
\midrule
\multicolumn{2}{l}{\textbf{Mean}} & 26.90 & 28.61 & 40.32 & \cellcolor{finalcol}\textbf{48.98} & \textbf{+22.08} \\
\bottomrule
\end{longtable}
\endgroup

\begingroup
\footnotesize
\begin{longtable}{@{}L{0.15\textwidth}L{0.24\textwidth}rrrrr@{}}
\caption{Complete GameCraft-Bench task scores for Pi + MiniMax-M3. Scores use the benchmark's 0--100 scale; $\Delta$ denotes HoH@3 minus Vanilla.}
\label{tab:gamecraft-task-scores-3}\\
\toprule
\rowcolor{headerrow}
\textbf{Family} & \textbf{Task} & \textbf{Vanilla} & \textbf{HoH@1} & \textbf{HoH@2} & \cellcolor{finalcol}\textbf{HoH@3} & \textbf{$\Delta$} \\
\midrule
\endfirsthead
\multicolumn{7}{l}{\small\itshape Table~\thetable\ (continued)}\\
\toprule
\rowcolor{headerrow}
\textbf{Family} & \textbf{Task} & \textbf{Vanilla} & \textbf{HoH@1} & \textbf{HoH@2} & \cellcolor{finalcol}\textbf{HoH@3} & \textbf{$\Delta$} \\
\midrule
\endhead
\rowcolor{actionrow}\multicolumn{7}{c}{\textbf{Action}} \\*
Platformer & Momentum Lab & 38.28 & 34.49 & 45.03 & \cellcolor{finalcol}48.80 & +10.52 \\
 & Ivory Beats & 36.95 & 39.64 & 39.89 & \cellcolor{finalcol}42.23 & +5.29 \\
 & Thunder Valkyrie & 60.52 & 50.14 & 61.74 & \cellcolor{finalcol}64.71 & +4.20 \\
\addlinespace[1pt]
Shooter & Void Patrol & 56.14 & 67.37 & 66.21 & \cellcolor{finalcol}76.66 & +20.52 \\
 & Wave Commander & 69.66 & 64.54 & 72.85 & \cellcolor{finalcol}75.52 & +5.86 \\
 & Hotline Heist & 41.46 & 47.81 & 44.65 & \cellcolor{finalcol}43.31 & +1.84 \\
\addlinespace[1pt]
Roguelike & Dungeon Shop & 25.08 & 40.12 & 47.16 & \cellcolor{finalcol}46.11 & +21.02 \\
 & Breach Tactics & 33.70 & 43.79 & 53.61 & \cellcolor{finalcol}59.56 & +25.86 \\
 & Void Harvest & 48.96 & 67.37 & 61.15 & \cellcolor{finalcol}67.28 & +18.32 \\
\addlinespace[1pt]
\rowcolor{timingrow}\multicolumn{7}{c}{\textbf{Timing}} \\*
Racing & Drift Circuit & 36.91 & 59.79 & 55.68 & \cellcolor{finalcol}55.24 & +18.33 \\
 & Rocket Trials & 35.50 & 38.21 & 37.33 & \cellcolor{finalcol}42.36 & +6.86 \\
 & Trick Runner & 41.79 & 51.66 & 59.92 & \cellcolor{finalcol}59.50 & +17.71 \\
\addlinespace[1pt]
Rhythm & Note Highway & 10.35 & 39.65 & 54.91 & \cellcolor{finalcol}62.58 & +52.23 \\
 & Beat Dungeon & 41.75 & 55.41 & 53.18 & \cellcolor{finalcol}64.83 & +23.08 \\
 & Garden & 41.70 & 63.78 & 56.48 & \cellcolor{finalcol}67.03 & +25.33 \\
\addlinespace[1pt]
Sports & Skateboard Park & 71.06 & 55.25 & 72.20 & \cellcolor{finalcol}84.99 & +13.92 \\
 & Boxing Gym & 40.44 & 55.00 & 57.71 & \cellcolor{finalcol}60.53 & +20.09 \\
 & Archery Quest & 24.25 & 51.31 & 61.23 & \cellcolor{finalcol}61.85 & +37.60 \\
\addlinespace[1pt]
\rowcolor{strategyrow}\multicolumn{7}{c}{\textbf{Strategy}} \\*
Strategy & Tower Defense & 43.51 & 57.22 & 46.99 & \cellcolor{finalcol}55.98 & +12.47 \\
 & Chess Variant & 24.75 & 31.46 & 27.15 & \cellcolor{finalcol}30.88 & +6.13 \\
 & Spell Tactics & 46.51 & 37.10 & 47.10 & \cellcolor{finalcol}53.93 & +7.42 \\
\addlinespace[1pt]
Card Game & Spire Descent & 26.61 & 18.09 & 10.96 & \cellcolor{finalcol}10.96 & -15.65 \\
 & Poker Roguelike & 3.19 & 30.31 & 50.81 & \cellcolor{finalcol}49.78 & +46.59 \\
 & Autobattler & 55.44 & 46.45 & 50.42 & \cellcolor{finalcol}53.37 & -2.07 \\
\addlinespace[1pt]
Puzzle & Sokoban Dungeon & 41.98 & 59.80 & 61.21 & \cellcolor{finalcol}57.66 & +15.68 \\
 & Circuit Wizard & 15.50 & 31.61 & 46.89 & \cellcolor{finalcol}53.51 & +38.01 \\
 & Pipe Crisis & 53.26 & 35.39 & 39.39 & \cellcolor{finalcol}37.65 & -15.61 \\
\addlinespace[1pt]
\rowcolor{simulationrow}\multicolumn{7}{c}{\textbf{Simulation}} \\*
Tycoon & Space Colony & 62.72 & 51.54 & 55.76 & \cellcolor{finalcol}58.33 & -4.39 \\
 & Pirate Port & 35.07 & 73.08 & 71.44 & \cellcolor{finalcol}65.70 & +30.63 \\
 & Wildhaven & 58.08 & 57.75 & 61.88 & \cellcolor{finalcol}65.75 & +7.66 \\
\addlinespace[1pt]
Idle & Ant Empire & 61.62 & 54.00 & 82.34 & \cellcolor{finalcol}82.56 & +20.94 \\
 & Factory Planet & 44.02 & 54.35 & 77.25 & \cellcolor{finalcol}80.35 & +36.33 \\
 & Dungeon Guild & 73.24 & 63.13 & 70.12 & \cellcolor{finalcol}73.84 & +0.60 \\
\addlinespace[1pt]
Simulation & Kitchen Rush & 16.80 & 42.60 & 49.42 & \cellcolor{finalcol}62.56 & +45.76 \\
 & Air Control & 28.61 & 55.82 & 63.76 & \cellcolor{finalcol}45.69 & +17.08 \\
 & Border Check & 53.54 & 33.74 & 36.83 & \cellcolor{finalcol}43.44 & -10.10 \\
\addlinespace[1pt]
\rowcolor{adventurerow}\multicolumn{7}{c}{\textbf{Adventure}} \\*
Horror & Floor 13 & 57.29 & 35.94 & 59.74 & \cellcolor{finalcol}66.60 & +9.31 \\
 & Dollhouse & 35.53 & 55.12 & 67.94 & \cellcolor{finalcol}71.90 & +36.37 \\
 & Lighthouse & 60.00 & 76.82 & 73.41 & \cellcolor{finalcol}74.72 & +14.72 \\
\addlinespace[1pt]
Open World & Sky Islands & 19.19 & 52.99 & 49.33 & \cellcolor{finalcol}59.73 & +40.54 \\
 & Airship Trader & 44.42 & 52.55 & 56.95 & \cellcolor{finalcol}60.52 & +16.10 \\
 & Bounty & 43.78 & 40.19 & 45.38 & \cellcolor{finalcol}58.44 & +14.66 \\
\addlinespace[1pt]
Visual Novel & Detective Noir & 40.87 & 46.41 & 66.07 & \cellcolor{finalcol}58.16 & +17.29 \\
 & Arcane Academy & 51.85 & 42.98 & 45.88 & \cellcolor{finalcol}65.71 & +13.86 \\
 & Time Paradox & 45.44 & 45.99 & 61.56 & \cellcolor{finalcol}64.22 & +18.78 \\
\addlinespace[1pt]
\midrule
\multicolumn{2}{l}{\textbf{Mean}} & 42.16 & 49.06 & 55.04 & \cellcolor{finalcol}\textbf{58.78} & \textbf{+16.62} \\
\bottomrule
\end{longtable}
\endgroup

\subsection{FrontierSWE Per-Task Rewards}

\begin{figure}[H]
    \centering
    \includegraphics[width=\textwidth]{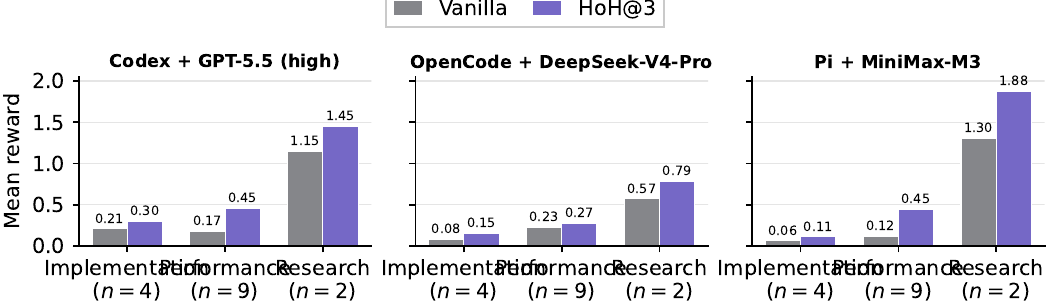}
    \caption{FrontierSWE mean rewards by official category and harness--model
    configuration.}
    \label{fig:supp-frontierswe-categories}
\end{figure}

Figure~\ref{fig:supp-frontierswe-categories} reports category means under the
official FrontierSWE taxonomy. The unequal task counts are shown explicitly
on the horizontal axis.

Tables~\ref{tab:frontierswe-task-scores-1}--\ref{tab:frontierswe-task-scores-3}
report every FrontierSWE task and condition.

% Generated from supplementary/data/frontierswe_task_records.csv.
\begingroup
\footnotesize
\begin{longtable}{@{}L{0.47\textwidth}rrrr@{}}
\caption{Complete FrontierSWE task rewards for Codex + GPT-5.5 (high).}
\label{tab:frontierswe-task-scores-1}\\
\toprule
\rowcolor{headerrow}
\textbf{Task} & \textbf{Vanilla} & \textbf{HoH@1} & \textbf{HoH@2} & \cellcolor{finalcol}\textbf{HoH@3} \\
\midrule
\endfirsthead
\multicolumn{5}{l}{\small\itshape Table~\thetable\ (continued)}\\
\toprule
\rowcolor{headerrow}
\textbf{Task} & \textbf{Vanilla} & \textbf{HoH@1} & \textbf{HoH@2} & \cellcolor{finalcol}\textbf{HoH@3} \\
\midrule
\endhead
\rowcolor{implementationrow}\multicolumn{5}{c}{\textbf{Implementation}} \\*
Dart Style Haskell & 0.00 & 0.06 & 0.07 & \cellcolor{finalcol}0.16 \\
Git to Zig & 0.18 & 0.18 & 0.17 & \cellcolor{finalcol}0.18 \\
Lua Native Compiler & 0.52 & 0.60 & 0.72 & \cellcolor{finalcol}0.70 \\
PostgreSQL--SQLite Wire Adapter & 0.14 & 0.15 & 0.15 & \cellcolor{finalcol}0.15 \\
\rowcolor{performancerow}\multicolumn{5}{c}{\textbf{Performance}} \\*
Cranelift Codegen Optimization & 0.00 & 0.00 & 0.00 & \cellcolor{finalcol}0.00 \\
Dependent Type Checker & 0.00 & 0.00 & 0.00 & \cellcolor{finalcol}0.00 \\
FFmpeg Swscale Rewrite & 0.00 & 0.00 & 0.00 & \cellcolor{finalcol}0.00 \\
Granite Mamba2 Inference Optimization & 0.20 & 1.01 & 1.07 & \cellcolor{finalcol}1.02 \\
Inference System Optimization & 0.00 & 0.00 & 0.00 & \cellcolor{finalcol}0.00 \\
Libexpat to x86 Assembly & 0.20 & 0.20 & 0.21 & \cellcolor{finalcol}0.21 \\
Notebook Compression & 0.00 & 0.69 & 0.69 & \cellcolor{finalcol}0.69 \\
Pyright Type-Checking Optimization & 1.16 & 1.11 & 1.17 & \cellcolor{finalcol}1.16 \\
Revideo Performance Optimization & 0.00 & 0.91 & 0.91 & \cellcolor{finalcol}0.99 \\
\rowcolor{researchrow}\multicolumn{5}{c}{\textbf{Research}} \\*
Optimizer Design & 1.40 & 1.71 & 1.46 & \cellcolor{finalcol}2.00 \\
PCQM4Mv2 Autoresearch & 0.90 & 0.89 & 0.89 & \cellcolor{finalcol}0.89 \\
\midrule
\textbf{Mean} & 0.31 & 0.50 & 0.50 & \cellcolor{finalcol}\textbf{0.54} \\
\bottomrule
\end{longtable}
\endgroup

\begingroup
\footnotesize
\begin{longtable}{@{}L{0.47\textwidth}rrrr@{}}
\caption{Complete FrontierSWE task rewards for OpenCode + DeepSeek-V4-Pro.}
\label{tab:frontierswe-task-scores-2}\\
\toprule
\rowcolor{headerrow}
\textbf{Task} & \textbf{Vanilla} & \textbf{HoH@1} & \textbf{HoH@2} & \cellcolor{finalcol}\textbf{HoH@3} \\
\midrule
\endfirsthead
\multicolumn{5}{l}{\small\itshape Table~\thetable\ (continued)}\\
\toprule
\rowcolor{headerrow}
\textbf{Task} & \textbf{Vanilla} & \textbf{HoH@1} & \textbf{HoH@2} & \cellcolor{finalcol}\textbf{HoH@3} \\
\midrule
\endhead
\rowcolor{implementationrow}\multicolumn{5}{c}{\textbf{Implementation}} \\*
Dart Style Haskell & 0.02 & 0.04 & 0.05 & \cellcolor{finalcol}0.04 \\
Git to Zig & 0.13 & 0.17 & 0.17 & \cellcolor{finalcol}0.17 \\
Lua Native Compiler & 0.02 & 0.02 & 0.03 & \cellcolor{finalcol}0.25 \\
PostgreSQL--SQLite Wire Adapter & 0.15 & 0.14 & 0.15 & \cellcolor{finalcol}0.15 \\
\rowcolor{performancerow}\multicolumn{5}{c}{\textbf{Performance}} \\*
Cranelift Codegen Optimization & 0.00 & 0.00 & 0.00 & \cellcolor{finalcol}0.00 \\
Dependent Type Checker & 0.00 & 0.00 & 0.00 & \cellcolor{finalcol}0.00 \\
FFmpeg Swscale Rewrite & 0.00 & 0.00 & 0.00 & \cellcolor{finalcol}0.00 \\
Granite Mamba2 Inference Optimization & 0.20 & 0.19 & 0.21 & \cellcolor{finalcol}0.20 \\
Inference System Optimization & 0.00 & 0.00 & 0.00 & \cellcolor{finalcol}0.00 \\
Libexpat to x86 Assembly & 0.00 & 0.00 & 0.00 & \cellcolor{finalcol}0.00 \\
Notebook Compression & 0.00 & 0.00 & 0.00 & \cellcolor{finalcol}0.00 \\
Pyright Type-Checking Optimization & 1.04 & 1.17 & 1.30 & \cellcolor{finalcol}1.29 \\
Revideo Performance Optimization & 0.80 & 0.75 & 0.92 & \cellcolor{finalcol}0.95 \\
\rowcolor{researchrow}\multicolumn{5}{c}{\textbf{Research}} \\*
Optimizer Design & 1.14 & 1.55 & 1.55 & \cellcolor{finalcol}1.57 \\
PCQM4Mv2 Autoresearch & 0.00 & 0.00 & 0.00 & \cellcolor{finalcol}0.00 \\
\midrule
\textbf{Mean} & 0.23 & 0.27 & 0.29 & \cellcolor{finalcol}\textbf{0.31} \\
\bottomrule
\end{longtable}
\endgroup

\begingroup
\footnotesize
\begin{longtable}{@{}L{0.47\textwidth}rrrr@{}}
\caption{Complete FrontierSWE task rewards for Pi + MiniMax-M3.}
\label{tab:frontierswe-task-scores-3}\\
\toprule
\rowcolor{headerrow}
\textbf{Task} & \textbf{Vanilla} & \textbf{HoH@1} & \textbf{HoH@2} & \cellcolor{finalcol}\textbf{HoH@3} \\
\midrule
\endfirsthead
\multicolumn{5}{l}{\small\itshape Table~\thetable\ (continued)}\\
\toprule
\rowcolor{headerrow}
\textbf{Task} & \textbf{Vanilla} & \textbf{HoH@1} & \textbf{HoH@2} & \cellcolor{finalcol}\textbf{HoH@3} \\
\midrule
\endhead
\rowcolor{implementationrow}\multicolumn{5}{c}{\textbf{Implementation}} \\*
Dart Style Haskell & 0.12 & 0.04 & 0.06 & \cellcolor{finalcol}0.07 \\
Git to Zig & 0.00 & 0.19 & 0.19 & \cellcolor{finalcol}0.19 \\
Lua Native Compiler & 0.00 & 0.03 & 0.03 & \cellcolor{finalcol}0.03 \\
PostgreSQL--SQLite Wire Adapter & 0.13 & 0.15 & 0.15 & \cellcolor{finalcol}0.15 \\
\rowcolor{performancerow}\multicolumn{5}{c}{\textbf{Performance}} \\*
Cranelift Codegen Optimization & 0.00 & 0.00 & 0.00 & \cellcolor{finalcol}0.00 \\
Dependent Type Checker & 0.00 & 0.00 & 0.00 & \cellcolor{finalcol}0.00 \\
FFmpeg Swscale Rewrite & 0.00 & 0.00 & 0.00 & \cellcolor{finalcol}0.00 \\
Granite Mamba2 Inference Optimization & 1.06 & 1.97 & 1.98 & \cellcolor{finalcol}2.18 \\
Inference System Optimization & 0.00 & 0.00 & 0.00 & \cellcolor{finalcol}0.00 \\
Libexpat to x86 Assembly & 0.00 & 0.00 & 0.00 & \cellcolor{finalcol}0.00 \\
Notebook Compression & 0.00 & 0.67 & 0.67 & \cellcolor{finalcol}0.67 \\
Pyright Type-Checking Optimization & 0.00 & 1.18 & 1.18 & \cellcolor{finalcol}1.17 \\
Revideo Performance Optimization & 0.00 & 0.00 & 0.00 & \cellcolor{finalcol}0.00 \\
\rowcolor{researchrow}\multicolumn{5}{c}{\textbf{Research}} \\*
Optimizer Design & 2.60 & 2.90 & 2.90 & \cellcolor{finalcol}2.87 \\
PCQM4Mv2 Autoresearch & 0.00 & 0.88 & 0.88 & \cellcolor{finalcol}0.88 \\
\midrule
\textbf{Mean} & 0.26 & 0.53 & 0.54 & \cellcolor{finalcol}\textbf{0.55} \\
\bottomrule
\end{longtable}
\endgroup

\subsection{Budget-Controlled Comparison}

\begin{figure}[H]
    \centering
    \includegraphics[width=0.80\textwidth]{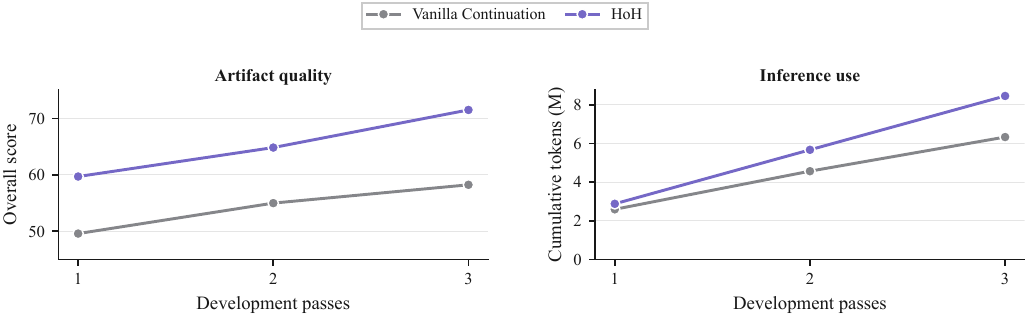}
    \caption{Score and cumulative token trajectories in the
    budget-controlled GameCraft-Bench comparison using Codex with GPT-5.5
    (high). HoH includes planning, coding, and testing at each pass.}
    \label{fig:supp-budget-tradeoff}
\end{figure}

The pass-controlled experiment uses Codex with GPT-5.5 (high) on the same 45
GameCraft-Bench tasks under the protocol in
Section~\ref{sec:supp-budget-protocol}. HoH uses three complete
planning--coding--testing iterations.
Figure~\ref{fig:supp-budget-tradeoff} shows how artifact quality and cumulative
token use change over the three development passes.

Tables~\ref{tab:gamecraft-budget-score-results}
and~\ref{tab:gamecraft-budget-token-results} report the task-level scores and
cumulative coding-harness tokens, respectively. Averaged over the 45 tasks,
Vanilla, three-pass Vanilla Continuation, and HoH obtain scores of 49.58,
58.24, and 71.52 using 2.59M, 6.33M, and 8.41M tokens per task, respectively.
By Eq.~\ref{eq:supp-budget-efficiency}, the three-pass conditions gain 2.32
and 3.77 score points per additional million tokens for Vanilla Continuation
and HoH, respectively.

% Generated from codex_gpt55_budget_comparison.csv.
\begingroup
\footnotesize
\begin{longtable}{@{}L{0.42\textwidth}rrrr@{}}
\caption{Task-level scores for the budget-controlled comparison on GameCraft-Bench using Codex + GPT-5.5 (high).}
\label{tab:gamecraft-budget-score-results}\\
\toprule
\rowcolor{headerrow}
\textbf{Task} & \textbf{Vanilla} & \textbf{Vanilla Cont.@2} & \textbf{Vanilla Cont.@3} & \cellcolor{finalcol}\textbf{HoH@3} \\
\midrule
\endfirsthead
\multicolumn{5}{l}{\small\itshape Table~\thetable\ (continued)}\\
\toprule
\rowcolor{headerrow}
\textbf{Task} & \textbf{Vanilla} & \textbf{Vanilla Cont.@2} & \textbf{Vanilla Cont.@3} & \cellcolor{finalcol}\textbf{HoH@3} \\
\midrule
\endhead
\rowcolor{actionrow}\multicolumn{5}{c}{\textbf{Action}} \\*
Ivory Beats & 46.81 & 44.58 & 44.84 & \cellcolor{finalcol}85.42 \\
Momentum Lab & 34.05 & 44.51 & 41.15 & \cellcolor{finalcol}70.61 \\
Thunder Valkyrie & 53.51 & 53.44 & 66.41 & \cellcolor{finalcol}73.67 \\
Hotline Heist & 43.37 & 41.70 & 44.21 & \cellcolor{finalcol}62.81 \\
Void Patrol & 58.59 & 74.05 & 72.66 & \cellcolor{finalcol}87.83 \\
Wave Commander & 66.97 & 69.79 & 63.52 & \cellcolor{finalcol}75.25 \\
Void Harvest & 41.05 & 41.06 & 40.60 & \cellcolor{finalcol}57.43 \\
Breach Tactics & 53.44 & 54.04 & 55.93 & \cellcolor{finalcol}61.09 \\
Dungeon Shop & 40.89 & 41.59 & 41.09 & \cellcolor{finalcol}65.09 \\
\rowcolor{timingrow}\multicolumn{5}{c}{\textbf{Timing}} \\*
Drift Circuit & 43.58 & 41.72 & 63.77 & \cellcolor{finalcol}70.08 \\
Rocket Trials & 45.19 & 43.81 & 49.82 & \cellcolor{finalcol}70.31 \\
Trick Runner & 51.50 & 48.38 & 53.64 & \cellcolor{finalcol}64.78 \\
Beat Dungeon & 47.94 & 46.70 & 50.41 & \cellcolor{finalcol}60.81 \\
Garden & 52.75 & 56.44 & 58.62 & \cellcolor{finalcol}70.69 \\
Note Highway & 39.44 & 40.84 & 44.28 & \cellcolor{finalcol}64.68 \\
Archery Quest & 58.09 & 66.64 & 72.67 & \cellcolor{finalcol}76.69 \\
Boxing Gym & 43.07 & 52.75 & 56.44 & \cellcolor{finalcol}73.14 \\
Skateboard Park & 57.64 & 68.37 & 67.47 & \cellcolor{finalcol}81.14 \\
\rowcolor{strategyrow}\multicolumn{5}{c}{\textbf{Strategy}} \\*
Chess Variant & 30.04 & 32.89 & 32.46 & \cellcolor{finalcol}59.72 \\
Spell Tactics & 53.10 & 52.98 & 52.75 & \cellcolor{finalcol}63.32 \\
Tower Defense & 58.85 & 69.76 & 71.02 & \cellcolor{finalcol}76.92 \\
Autobattler & 55.06 & 63.80 & 59.77 & \cellcolor{finalcol}72.22 \\
Poker Roguelike & 44.66 & 48.69 & 51.48 & \cellcolor{finalcol}69.27 \\
Spire Descent & 25.85 & 41.60 & 55.81 & \cellcolor{finalcol}61.92 \\
Circuit Wizard & 31.35 & 34.47 & 36.78 & \cellcolor{finalcol}49.52 \\
Pipe Crisis & 41.88 & 39.64 & 45.39 & \cellcolor{finalcol}72.17 \\
Sokoban Dungeon & 55.75 & 65.11 & 67.14 & \cellcolor{finalcol}70.13 \\
\rowcolor{simulationrow}\multicolumn{5}{c}{\textbf{Simulation}} \\*
Pirate Port & 51.32 & 48.05 & 72.20 & \cellcolor{finalcol}77.81 \\
Space Colony & 36.87 & 43.97 & 45.09 & \cellcolor{finalcol}76.55 \\
Wildhaven & 49.42 & 60.30 & 64.68 & \cellcolor{finalcol}80.39 \\
Ant Empire & 65.52 & 81.58 & 85.84 & \cellcolor{finalcol}87.88 \\
Dungeon Guild & 63.92 & 72.76 & 75.41 & \cellcolor{finalcol}79.50 \\
Factory Planet & 75.16 & 79.27 & 80.69 & \cellcolor{finalcol}87.78 \\
Air Control & 54.24 & 44.85 & 55.18 & \cellcolor{finalcol}69.73 \\
Border Check & 43.57 & 69.05 & 69.45 & \cellcolor{finalcol}72.74 \\
Kitchen Rush & 42.62 & 58.57 & 54.23 & \cellcolor{finalcol}73.38 \\
\rowcolor{adventurerow}\multicolumn{5}{c}{\textbf{Adventure}} \\*
Dollhouse & 56.33 & 64.55 & 70.36 & \cellcolor{finalcol}72.59 \\
Floor 13 & 48.68 & 65.44 & 66.90 & \cellcolor{finalcol}73.41 \\
Lighthouse & 51.95 & 69.16 & 72.70 & \cellcolor{finalcol}80.90 \\
Airship Trader & 59.58 & 67.71 & 65.15 & \cellcolor{finalcol}74.41 \\
Bounty & 47.00 & 51.46 & 54.09 & \cellcolor{finalcol}68.27 \\
Sky Islands & 53.25 & 61.56 & 63.38 & \cellcolor{finalcol}68.25 \\
Arcane Academy & 59.30 & 54.02 & 52.10 & \cellcolor{finalcol}70.69 \\
Detective Noir & 47.46 & 45.40 & 51.15 & \cellcolor{finalcol}65.31 \\
Time Paradox & 50.63 & 57.48 & 61.92 & \cellcolor{finalcol}71.97 \\
\midrule
\textbf{Mean} & 49.58 & 54.99 & 58.24 & \cellcolor{finalcol}\textbf{71.52} \\
\bottomrule
\end{longtable}
\endgroup

\begingroup
\footnotesize
\begin{longtable}{@{}L{0.42\textwidth}rrrr@{}}
\caption{Task-level cumulative token usage (M) for the budget-controlled comparison on GameCraft-Bench using Codex + GPT-5.5 (high).}
\label{tab:gamecraft-budget-token-results}\\
\toprule
\rowcolor{headerrow}
\textbf{Task} & \textbf{Vanilla} & \textbf{Vanilla Cont.@2} & \textbf{Vanilla Cont.@3} & \cellcolor{finalcol}\textbf{HoH@3} \\
\midrule
\endfirsthead
\multicolumn{5}{l}{\small\itshape Table~\thetable\ (continued)}\\
\toprule
\rowcolor{headerrow}
\textbf{Task} & \textbf{Vanilla} & \textbf{Vanilla Cont.@2} & \textbf{Vanilla Cont.@3} & \cellcolor{finalcol}\textbf{HoH@3} \\
\midrule
\endhead
\rowcolor{actionrow}\multicolumn{5}{c}{\textbf{Action}} \\*
Ivory Beats & 1.49 & 2.32 & 3.17 & \cellcolor{finalcol}5.64 \\
Momentum Lab & 2.27 & 5.07 & 8.13 & \cellcolor{finalcol}9.00 \\
Thunder Valkyrie & 1.86 & 4.96 & 6.61 & \cellcolor{finalcol}8.27 \\
Hotline Heist & 1.41 & 2.28 & 3.70 & \cellcolor{finalcol}7.68 \\
Void Patrol & 2.42 & 4.21 & 6.39 & \cellcolor{finalcol}6.55 \\
Wave Commander & 2.61 & 4.66 & 5.82 & \cellcolor{finalcol}9.37 \\
Void Harvest & 2.75 & 4.62 & 6.37 & \cellcolor{finalcol}8.50 \\
Breach Tactics & 2.43 & 4.50 & 8.46 & \cellcolor{finalcol}7.54 \\
Dungeon Shop & 2.53 & 4.99 & 6.33 & \cellcolor{finalcol}6.31 \\
\rowcolor{timingrow}\multicolumn{5}{c}{\textbf{Timing}} \\*
Drift Circuit & 2.20 & 6.18 & 8.22 & \cellcolor{finalcol}9.16 \\
Rocket Trials & 1.92 & 4.55 & 5.90 & \cellcolor{finalcol}10.01 \\
Trick Runner & 4.13 & 5.83 & 7.68 & \cellcolor{finalcol}7.65 \\
Beat Dungeon & 1.85 & 3.20 & 4.31 & \cellcolor{finalcol}6.84 \\
Garden & 4.57 & 6.08 & 7.59 & \cellcolor{finalcol}9.65 \\
Note Highway & 2.73 & 5.26 & 9.05 & \cellcolor{finalcol}6.19 \\
Archery Quest & 1.51 & 5.62 & 7.24 & \cellcolor{finalcol}11.16 \\
Boxing Gym & 3.25 & 5.19 & 6.55 & \cellcolor{finalcol}7.82 \\
Skateboard Park & 2.41 & 3.31 & 4.44 & \cellcolor{finalcol}7.63 \\
\rowcolor{strategyrow}\multicolumn{5}{c}{\textbf{Strategy}} \\*
Chess Variant & 1.77 & 2.48 & 5.00 & \cellcolor{finalcol}9.13 \\
Spell Tactics & 2.22 & 3.83 & 5.71 & \cellcolor{finalcol}7.37 \\
Tower Defense & 2.12 & 3.77 & 6.12 & \cellcolor{finalcol}11.85 \\
Autobattler & 2.74 & 4.74 & 6.16 & \cellcolor{finalcol}10.83 \\
Poker Roguelike & 3.41 & 7.13 & 9.29 & \cellcolor{finalcol}7.35 \\
Spire Descent & 2.31 & 4.16 & 6.11 & \cellcolor{finalcol}11.51 \\
Circuit Wizard & 2.06 & 4.13 & 6.70 & \cellcolor{finalcol}7.01 \\
Pipe Crisis & 1.73 & 3.33 & 4.27 & \cellcolor{finalcol}5.97 \\
Sokoban Dungeon & 1.30 & 2.67 & 3.88 & \cellcolor{finalcol}6.27 \\
\rowcolor{simulationrow}\multicolumn{5}{c}{\textbf{Simulation}} \\*
Pirate Port & 2.43 & 3.32 & 5.81 & \cellcolor{finalcol}6.69 \\
Space Colony & 4.79 & 6.72 & 9.04 & \cellcolor{finalcol}9.27 \\
Wildhaven & 2.58 & 4.01 & 5.22 & \cellcolor{finalcol}9.93 \\
Ant Empire & 3.11 & 4.87 & 6.38 & \cellcolor{finalcol}11.02 \\
Dungeon Guild & 2.41 & 6.81 & 9.33 & \cellcolor{finalcol}9.28 \\
Factory Planet & 3.75 & 6.30 & 7.23 & \cellcolor{finalcol}9.15 \\
Air Control & 3.25 & 5.32 & 7.38 & \cellcolor{finalcol}7.97 \\
Border Check & 1.52 & 3.34 & 4.92 & \cellcolor{finalcol}7.30 \\
Kitchen Rush & 2.36 & 4.19 & 5.18 & \cellcolor{finalcol}8.98 \\
\rowcolor{adventurerow}\multicolumn{5}{c}{\textbf{Adventure}} \\*
Dollhouse & 1.20 & 2.22 & 3.61 & \cellcolor{finalcol}8.10 \\
Floor 13 & 1.68 & 3.25 & 4.90 & \cellcolor{finalcol}5.81 \\
Lighthouse & 2.91 & 4.25 & 6.07 & \cellcolor{finalcol}8.24 \\
Airship Trader & 1.92 & 3.58 & 4.78 & \cellcolor{finalcol}7.34 \\
Bounty & 4.18 & 7.97 & 9.94 & \cellcolor{finalcol}8.32 \\
Sky Islands & 3.60 & 4.78 & 5.72 & \cellcolor{finalcol}15.71 \\
Arcane Academy & 3.59 & 5.39 & 7.70 & \cellcolor{finalcol}6.56 \\
Detective Noir & 3.69 & 4.09 & 4.86 & \cellcolor{finalcol}8.04 \\
Time Paradox & 3.68 & 5.91 & 7.53 & \cellcolor{finalcol}8.23 \\
\midrule
\textbf{Mean} & 2.59 & 4.56 & 6.33 & \cellcolor{finalcol}\textbf{8.41} \\
\bottomrule
\end{longtable}
\endgroup

\subsection{Ablation Study}

\begin{figure}[H]
    \centering
    \includegraphics[width=0.92\textwidth]{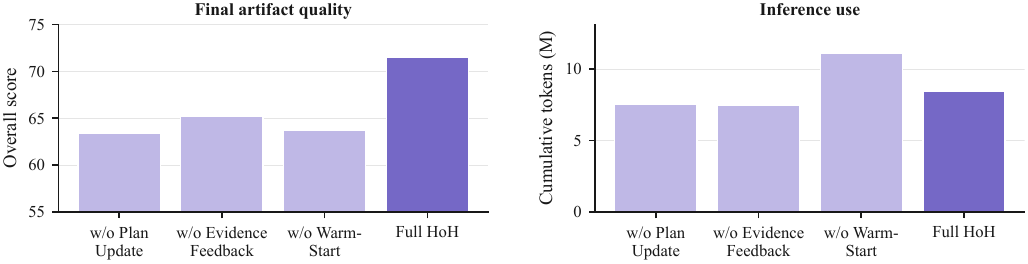}
    \caption{Final score and cumulative token use for full HoH and the three
    cross-iteration ablations on GameCraft-Bench.}
    \label{fig:supp-ablation}
\end{figure}

We evaluate three variants of HoH with $T=3$ on all 45 GameCraft-Bench tasks
using Codex with GPT-5.5 (high), following the interventions in
Section~\ref{sec:supp-ablation-protocol}. The complete task-level scores are
reported in
Table~\ref{tab:gamecraft-ablation-task-results}. Aggregate token usage is
reported separately in Table~\ref{tab:gamecraft-ablation-token-results}. The
mean score decreases from 71.52 for full HoH to 63.39 without plan update,
65.23 without evidence feedback, and 63.67 without artifact warm-start.
Figure~\ref{fig:supp-ablation} places the score changes beside their cumulative
token use.

% Generated from codex_gpt55_ablation.csv.
\begingroup
\footnotesize
\begin{longtable}{@{}L{0.31\textwidth}rrrr@{}}
\caption{Task-level GameCraft-Bench ablation scores using Codex + GPT-5.5 (high), with $T=3$. All values use the benchmark's 0--100 scale.}
\label{tab:gamecraft-ablation-task-results}\\
\toprule
\rowcolor{headerrow}
\textbf{Task} & \cellcolor{finalcol}\textbf{Full HoH} & \textbf{\shortstack{w/o\\Plan Update}} & \textbf{\shortstack{w/o\\Evidence Feedback}} & \textbf{\shortstack{w/o\\Warm-Start}} \\
\midrule
\endfirsthead
\multicolumn{5}{l}{\small\itshape Table~\thetable\ (continued)}\\
\toprule
\rowcolor{headerrow}
\textbf{Task} & \cellcolor{finalcol}\textbf{Full HoH} & \textbf{\shortstack{w/o\\Plan Update}} & \textbf{\shortstack{w/o\\Evidence Feedback}} & \textbf{\shortstack{w/o\\Warm-Start}} \\
\midrule
\endhead
\rowcolor{actionrow}\multicolumn{5}{c}{\textbf{Action}} \\*
Momentum Lab & \cellcolor{finalcol}70.61 & 67.81 & 56.67 & 58.77 \\
Ivory Beats & \cellcolor{finalcol}85.42 & 69.42 & 80.33 & 78.85 \\
Thunder Valkyrie & \cellcolor{finalcol}73.67 & 69.99 & 72.94 & 72.51 \\
Void Patrol & \cellcolor{finalcol}87.83 & 73.22 & 77.79 & 72.05 \\
Wave Commander & \cellcolor{finalcol}75.25 & 68.74 & 73.28 & 68.31 \\
Hotline Heist & \cellcolor{finalcol}62.81 & 52.40 & 58.24 & 35.21 \\
Dungeon Shop & \cellcolor{finalcol}65.09 & 57.73 & 59.33 & 64.00 \\
Breach Tactics & \cellcolor{finalcol}61.09 & 58.73 & 53.49 & 52.89 \\
Void Harvest & \cellcolor{finalcol}57.43 & 53.96 & 52.46 & 53.04 \\
\rowcolor{timingrow}\multicolumn{5}{c}{\textbf{Timing}} \\*
Drift Circuit & \cellcolor{finalcol}70.08 & 63.19 & 66.85 & 52.81 \\
Rocket Trials & \cellcolor{finalcol}70.31 & 63.56 & 68.15 & 64.09 \\
Trick Runner & \cellcolor{finalcol}64.78 & 62.55 & 62.45 & 57.59 \\
Note Highway & \cellcolor{finalcol}64.68 & 63.84 & 61.32 & 64.51 \\
Beat Dungeon & \cellcolor{finalcol}60.81 & 58.38 & 57.50 & 50.84 \\
Garden & \cellcolor{finalcol}70.69 & 65.61 & 64.19 & 67.89 \\
Skateboard Park & \cellcolor{finalcol}81.14 & 75.77 & 75.24 & 77.44 \\
Boxing Gym & \cellcolor{finalcol}73.14 & 47.90 & 43.73 & 71.16 \\
Archery Quest & \cellcolor{finalcol}76.69 & 66.11 & 72.52 & 65.61 \\
\rowcolor{strategyrow}\multicolumn{5}{c}{\textbf{Strategy}} \\*
Tower Defense & \cellcolor{finalcol}76.92 & 68.56 & 63.55 & 51.71 \\
Chess Variant & \cellcolor{finalcol}59.72 & 57.91 & 47.87 & 51.58 \\
Spell Tactics & \cellcolor{finalcol}63.32 & 50.70 & 58.73 & 58.81 \\
Spire Descent & \cellcolor{finalcol}61.92 & 46.15 & 60.31 & 60.21 \\
Poker Roguelike & \cellcolor{finalcol}69.27 & 64.70 & 64.11 & 58.89 \\
Autobattler & \cellcolor{finalcol}72.22 & 68.39 & 71.42 & 56.35 \\
Sokoban Dungeon & \cellcolor{finalcol}70.13 & 57.02 & 68.60 & 64.19 \\
Circuit Wizard & \cellcolor{finalcol}49.52 & 39.38 & 35.48 & 40.11 \\
Pipe Crisis & \cellcolor{finalcol}72.17 & 62.00 & 61.94 & 46.91 \\
\rowcolor{simulationrow}\multicolumn{5}{c}{\textbf{Simulation}} \\*
Space Colony & \cellcolor{finalcol}76.55 & 57.86 & 75.17 & 75.14 \\
Pirate Port & \cellcolor{finalcol}77.81 & 75.55 & 70.51 & 76.69 \\
Wildhaven & \cellcolor{finalcol}80.39 & 79.00 & 79.12 & 75.22 \\
Ant Empire & \cellcolor{finalcol}87.88 & 78.05 & 82.11 & 80.69 \\
Factory Planet & \cellcolor{finalcol}87.78 & 70.69 & 83.21 & 84.36 \\
Dungeon Guild & \cellcolor{finalcol}79.50 & 67.98 & 73.23 & 76.10 \\
Kitchen Rush & \cellcolor{finalcol}73.38 & 57.69 & 60.10 & 56.94 \\
Air Control & \cellcolor{finalcol}69.73 & 63.84 & 66.51 & 64.92 \\
Border Check & \cellcolor{finalcol}72.74 & 65.41 & 62.44 & 61.16 \\
\rowcolor{adventurerow}\multicolumn{5}{c}{\textbf{Adventure}} \\*
Floor 13 & \cellcolor{finalcol}73.41 & 68.94 & 69.89 & 63.37 \\
Dollhouse & \cellcolor{finalcol}72.59 & 60.84 & 60.38 & 67.71 \\
Lighthouse & \cellcolor{finalcol}80.90 & 72.72 & 74.72 & 81.50 \\
Sky Islands & \cellcolor{finalcol}68.25 & 62.50 & 63.07 & 59.85 \\
Airship Trader & \cellcolor{finalcol}74.41 & 61.49 & 69.94 & 71.21 \\
Bounty & \cellcolor{finalcol}68.27 & 67.41 & 68.20 & 64.57 \\
Detective Noir & \cellcolor{finalcol}65.31 & 57.41 & 51.77 & 60.54 \\
Arcane Academy & \cellcolor{finalcol}70.69 & 64.35 & 67.72 & 57.91 \\
Time Paradox & \cellcolor{finalcol}71.97 & 67.02 & 68.91 & 70.96 \\
\midrule
\textbf{Mean} & \cellcolor{finalcol}\textbf{71.52} & 63.39 & 65.23 & 63.67 \\
\bottomrule
\end{longtable}
\endgroup

\begin{table}[H]
\centering
\caption{Mean cumulative coding-harness tokens per task for the GameCraft-Bench ablation variants.}
\label{tab:gamecraft-ablation-token-results}
\small
\setlength{\tabcolsep}{8pt}
\begin{tabular}{@{}lr@{}}
\toprule
\textbf{Variant} & \textbf{Tokens (M)} \\
\midrule
w/o Plan Update & 7.56 \\
w/o Evidence Feedback & 7.46 \\
w/o Warm-Start & 11.12 \\
\midrule
Full HoH & \textbf{8.41} \\
\bottomrule
\end{tabular}
\end{table}

\subsection{Resource Usage}

\begin{figure}[H]
    \centering
    \includegraphics[width=\textwidth]{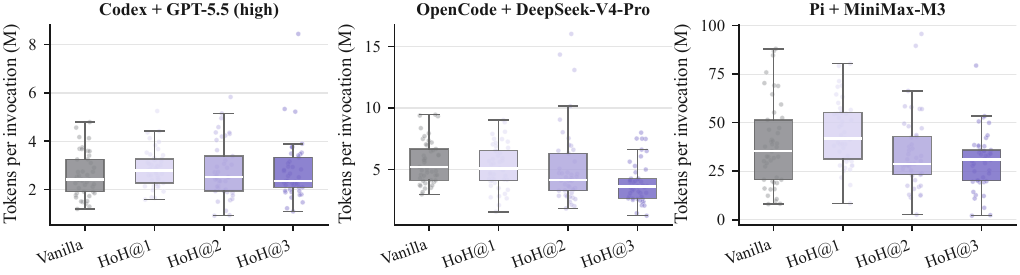}
    \caption{Per-invocation coding-harness token distributions on
    GameCraft-Bench. Points denote tasks; boxes show the median and
    interquartile range.}
    \label{fig:supp-iteration-tokens}
\end{figure}

Figure~\ref{fig:supp-iteration-tokens} shows the distribution of tokens used by
each invocation on the 45 GameCraft-Bench tasks. The panels retain the native
provider accounting for each harness--model configuration and should therefore
be compared within, rather than across, panels.

Table~\ref{tab:frontierswe-resources} aggregates the recorded resource use for
the FrontierSWE runs.

% Generated from supplementary/data/frontierswe_task_records.csv.
\begin{table}[H]
\centering
\caption{Aggregate FrontierSWE resource usage.}
\label{tab:frontierswe-resources}
\scriptsize
\setlength{\tabcolsep}{3.2pt}
\begin{tabular*}{\textwidth}{@{\extracolsep{\fill}}lrrrrrrrr@{}}
\toprule
\textbf{Harness--model} & \multicolumn{2}{c}{\textbf{Vanilla}} & \multicolumn{2}{c}{\textbf{HoH@1}} & \multicolumn{2}{c}{\textbf{HoH@2}} & \multicolumn{2}{c}{\textbf{HoH@3}} \\
\cmidrule(lr){2-3}\cmidrule(lr){4-5}\cmidrule(lr){6-7}\cmidrule(l){8-9}
 & \textbf{Tokens} & \textbf{Time} & \textbf{Tokens} & \textbf{Time} & \textbf{Tokens} & \textbf{Time} & \textbf{Tokens} & \textbf{Time} \\
 & \textbf{(M)} & \textbf{(h)} & \textbf{(M)} & \textbf{(h)} & \textbf{(M)} & \textbf{(h)} & \textbf{(M)} & \textbf{(h)} \\
\midrule
Codex + GPT-5.5 (high) & 103.43 & 18.65 & 109.26 & 14.32 & 83.19 & 13.21 & 71.71 & 10.28 \\
OpenCode + DeepSeek-V4-Pro & 384.84 & 32.00 & 332.61 & 23.58 & 345.66 & 23.32 & 229.77 & 16.16 \\
Pi + MiniMax-M3 & 541.73 & 40.19 & 1117.11 & 49.05 & 988.00 & 21.53 & 717.69 & 37.15 \\
\bottomrule
\end{tabular*}
\end{table}

\clearpage
\section{Qualitative Analysis}
\label{sec:supp-qualitative}

\begin{figure}[H]
    \centering
    \includegraphics[width=\textwidth]{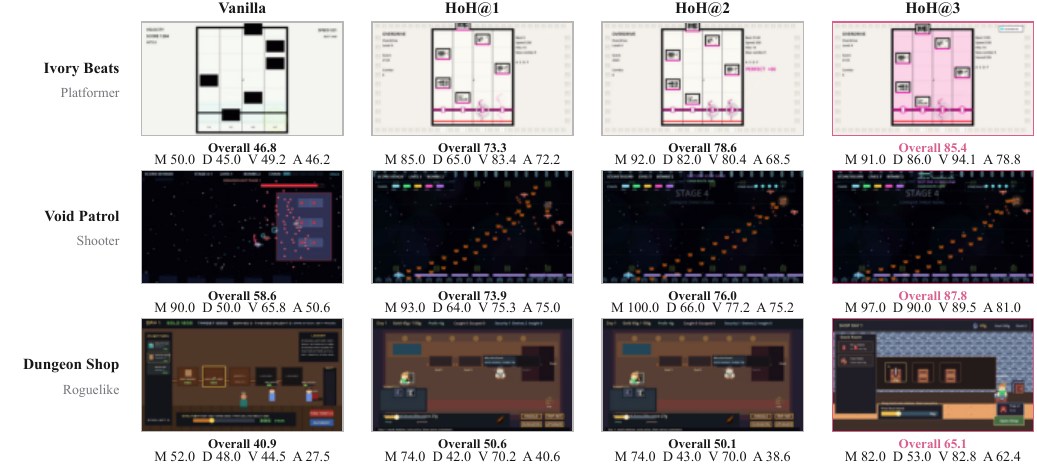}
    \caption{Action: Platformer, Shooter, and Roguelike.}
    \label{fig:supp-qualitative-action}
\end{figure}

\begin{figure}[H]
    \centering
    \includegraphics[width=\textwidth]{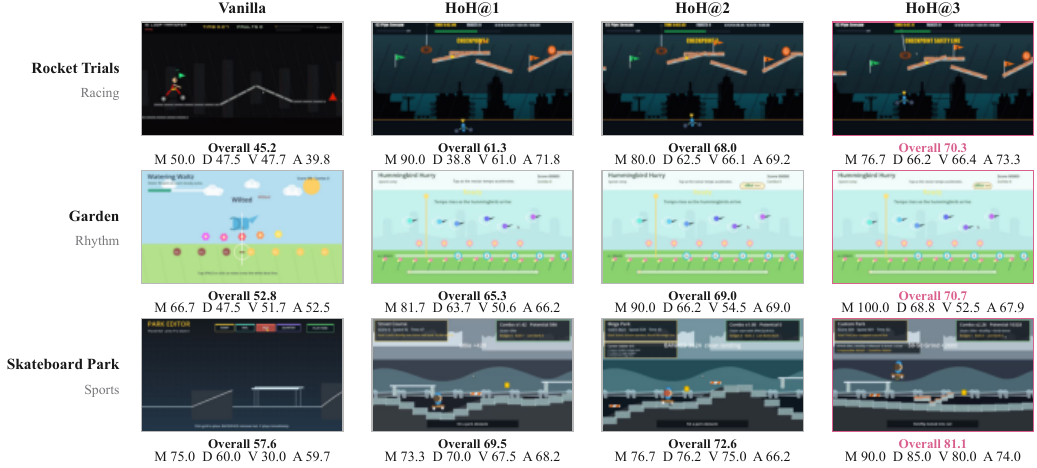}
    \caption{Timing: Racing, Rhythm, and Sports.}
    \label{fig:supp-qualitative-timing}
\end{figure}

\begin{figure}[H]
    \centering
    \includegraphics[width=\textwidth]{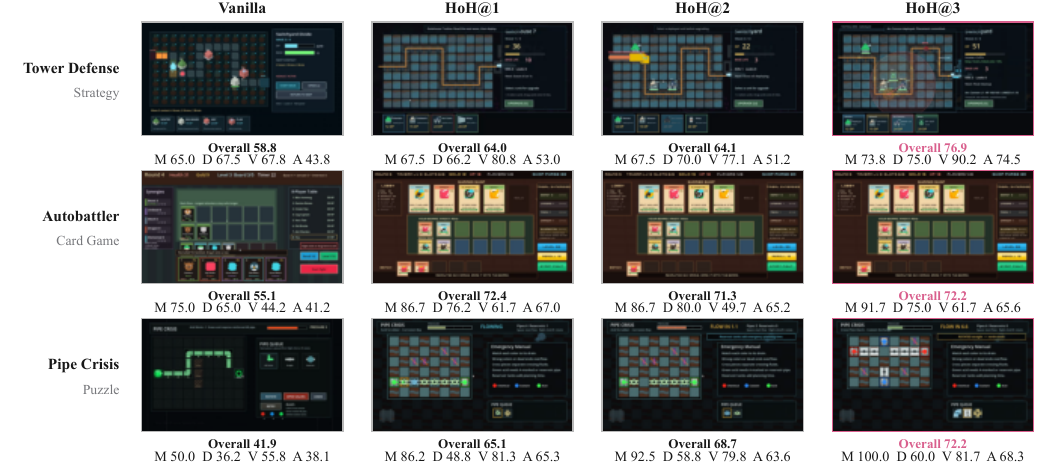}
    \caption{Strategy: Strategy, Card Game, and Puzzle.}
    \label{fig:supp-qualitative-strategy}
\end{figure}

\begin{figure}[H]
    \centering
    \includegraphics[width=\textwidth]{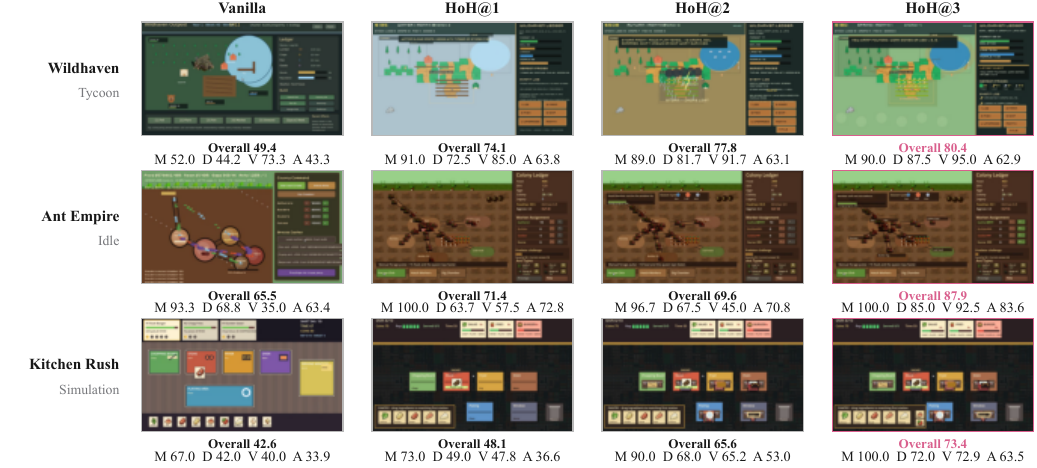}
    \caption{Simulation: Tycoon, Idle, and Simulation.}
    \label{fig:supp-qualitative-simulation}
\end{figure}

\begin{figure}[H]
    \centering
    \includegraphics[width=\textwidth]{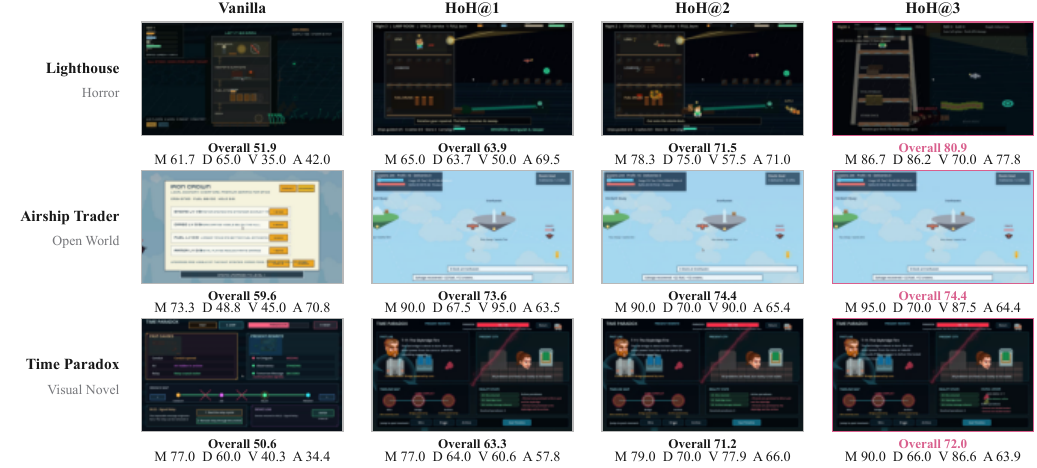}
    \caption{Adventure: Horror, Open World, and Visual Novel.}
    \label{fig:supp-qualitative-adventure}
\end{figure}

Figures~\ref{fig:supp-qualitative-action}--\ref{fig:supp-qualitative-adventure}
show one representative game from each of the 15 GameCraft-Bench families.
For each family, we select the game with the highest HoH@3 Overall score under
Codex with GPT-5.5 (high). Each row compares Vanilla and HoH@1--3; the values
beneath each artifact report Overall, Core Mechanics (M), Content Depth (D),
Functional Visuals (V), and Art and Presentation (A).

Table~\ref{tab:qualitative-dimension-scores} reports the exact Overall and
dimension scores underlying all 60 artifacts shown above.

% Generated by supplementary/scripts/generate_qualitative_analysis.py.
\begingroup
\footnotesize
\renewcommand{\arraystretch}{1.08}
\setlength{\tabcolsep}{4pt}
\begin{longtable}{@{}L{0.13\textwidth}L{0.19\textwidth}L{0.10\textwidth}rrrrr@{}}
\caption{Scores for the 15 GameCraft-Bench games shown in the qualitative comparison. One game is selected per benchmark family by the highest HoH@3 Overall score under Codex with GPT-5.5 (high).}
\label{tab:qualitative-dimension-scores}\\
\toprule
\rowcolor{headerrow}
\textbf{Family} & \textbf{Game} & \textbf{Condition} & \textbf{Overall} & \textbf{Mechanics} & \textbf{Depth} & \textbf{Visuals} & \textbf{Art} \\
\midrule
\endfirsthead
\multicolumn{8}{l}{\small\itshape Table~\thetable\ (continued)}\\
\toprule
\rowcolor{headerrow}
\textbf{Family} & \textbf{Game} & \textbf{Condition} & \textbf{Overall} & \textbf{Mechanics} & \textbf{Depth} & \textbf{Visuals} & \textbf{Art} \\
\midrule
\endhead
\rowcolor{actionrow}\multicolumn{8}{c}{\textbf{Action}} \\*
\multirow{4}{0.13\textwidth}{Platformer} & \multirow{4}{0.19\textwidth}{Ivory Beats} & Vanilla & 46.81 & 50.00 & 45.00 & 49.17 & 46.25 \\
 &  & HoH@1 & 73.28 & 85.00 & 65.00 & 83.44 & 72.19 \\
 &  & HoH@2 & 78.55 & 92.00 & 82.00 & 80.42 & 68.54 \\
 &  & \cellcolor{finalcol}\textbf{HoH@3} & \cellcolor{finalcol}\textbf{85.42} & \cellcolor{finalcol}\textbf{91.00} & \cellcolor{finalcol}\textbf{86.00} & \cellcolor{finalcol}\textbf{94.06} & \cellcolor{finalcol}\textbf{78.75} \\
\addlinespace[2pt]
\multirow{4}{0.13\textwidth}{Shooter} & \multirow{4}{0.19\textwidth}{Void Patrol} & Vanilla & 58.59 & 90.00 & 50.00 & 65.83 & 50.62 \\
 &  & HoH@1 & 73.90 & 93.00 & 64.00 & 75.31 & 75.00 \\
 &  & HoH@2 & 76.02 & 100.00 & 66.00 & 77.25 & 75.25 \\
 &  & \cellcolor{finalcol}\textbf{HoH@3} & \cellcolor{finalcol}\textbf{87.83} & \cellcolor{finalcol}\textbf{97.00} & \cellcolor{finalcol}\textbf{90.00} & \cellcolor{finalcol}\textbf{89.50} & \cellcolor{finalcol}\textbf{81.00} \\
\addlinespace[2pt]
\multirow{4}{0.13\textwidth}{Roguelike} & \multirow{4}{0.19\textwidth}{Dungeon Shop} & Vanilla & 40.89 & 52.00 & 48.00 & 44.46 & 27.50 \\
 &  & HoH@1 & 50.55 & 74.00 & 42.00 & 70.21 & 40.62 \\
 &  & HoH@2 & 50.15 & 74.00 & 43.00 & 70.00 & 38.57 \\
 &  & \cellcolor{finalcol}\textbf{HoH@3} & \cellcolor{finalcol}\textbf{65.09} & \cellcolor{finalcol}\textbf{82.00} & \cellcolor{finalcol}\textbf{53.00} & \cellcolor{finalcol}\textbf{82.75} & \cellcolor{finalcol}\textbf{62.38} \\
\addlinespace[2pt]
\rowcolor{timingrow}\multicolumn{8}{c}{\textbf{Timing}} \\*
\multirow{4}{0.13\textwidth}{Racing} & \multirow{4}{0.19\textwidth}{Rocket Trials} & Vanilla & 45.19 & 50.00 & 47.50 & 47.67 & 39.75 \\
 &  & HoH@1 & 61.33 & 90.00 & 38.75 & 61.00 & 71.75 \\
 &  & HoH@2 & 68.00 & 80.00 & 62.50 & 66.11 & 69.17 \\
 &  & \cellcolor{finalcol}\textbf{HoH@3} & \cellcolor{finalcol}\textbf{70.31} & \cellcolor{finalcol}\textbf{76.67} & \cellcolor{finalcol}\textbf{66.25} & \cellcolor{finalcol}\textbf{66.39} & \cellcolor{finalcol}\textbf{73.33} \\
\addlinespace[2pt]
\multirow{4}{0.13\textwidth}{Rhythm} & \multirow{4}{0.19\textwidth}{Garden} & Vanilla & 52.75 & 66.67 & 47.50 & 51.67 & 52.50 \\
 &  & HoH@1 & 65.33 & 81.67 & 63.75 & 50.56 & 66.25 \\
 &  & HoH@2 & 69.01 & 90.00 & 66.25 & 54.50 & 69.00 \\
 &  & \cellcolor{finalcol}\textbf{HoH@3} & \cellcolor{finalcol}\textbf{70.69} & \cellcolor{finalcol}\textbf{100.00} & \cellcolor{finalcol}\textbf{68.75} & \cellcolor{finalcol}\textbf{52.50} & \cellcolor{finalcol}\textbf{67.88} \\
\addlinespace[2pt]
\multirow{4}{0.13\textwidth}{Sports} & \multirow{4}{0.19\textwidth}{Skateboard Park} & Vanilla & 57.64 & 75.00 & 60.00 & 30.00 & 59.69 \\
 &  & HoH@1 & 69.51 & 73.33 & 70.00 & 67.50 & 68.25 \\
 &  & HoH@2 & 72.62 & 76.67 & 76.25 & 75.00 & 66.25 \\
 &  & \cellcolor{finalcol}\textbf{HoH@3} & \cellcolor{finalcol}\textbf{81.14} & \cellcolor{finalcol}\textbf{90.00} & \cellcolor{finalcol}\textbf{85.00} & \cellcolor{finalcol}\textbf{80.00} & \cellcolor{finalcol}\textbf{73.96} \\
\addlinespace[2pt]
\rowcolor{strategyrow}\multicolumn{8}{c}{\textbf{Strategy}} \\*
\multirow{4}{0.13\textwidth}{Strategy} & \multirow{4}{0.19\textwidth}{Tower Defense} & Vanilla & 58.85 & 65.00 & 67.50 & 67.75 & 43.75 \\
 &  & HoH@1 & 63.97 & 67.50 & 66.25 & 80.75 & 53.00 \\
 &  & HoH@2 & 64.12 & 67.50 & 70.00 & 77.08 & 51.25 \\
 &  & \cellcolor{finalcol}\textbf{HoH@3} & \cellcolor{finalcol}\textbf{76.92} & \cellcolor{finalcol}\textbf{73.75} & \cellcolor{finalcol}\textbf{75.00} & \cellcolor{finalcol}\textbf{90.25} & \cellcolor{finalcol}\textbf{74.50} \\
\addlinespace[2pt]
\multirow{4}{0.13\textwidth}{Card Game} & \multirow{4}{0.19\textwidth}{Autobattler} & Vanilla & 55.06 & 75.00 & 65.00 & 44.17 & 41.25 \\
 &  & HoH@1 & 72.39 & 86.67 & 76.25 & 61.67 & 67.00 \\
 &  & HoH@2 & 71.28 & 86.67 & 80.00 & 49.72 & 65.21 \\
 &  & \cellcolor{finalcol}\textbf{HoH@3} & \cellcolor{finalcol}\textbf{72.22} & \cellcolor{finalcol}\textbf{91.67} & \cellcolor{finalcol}\textbf{75.00} & \cellcolor{finalcol}\textbf{61.67} & \cellcolor{finalcol}\textbf{65.62} \\
\addlinespace[2pt]
\multirow{4}{0.13\textwidth}{Puzzle} & \multirow{4}{0.19\textwidth}{Pipe Crisis} & Vanilla & 41.88 & 50.00 & 36.25 & 55.83 & 38.06 \\
 &  & HoH@1 & 65.07 & 86.25 & 48.75 & 81.33 & 65.33 \\
 &  & HoH@2 & 68.65 & 92.50 & 58.75 & 79.76 & 63.57 \\
 &  & \cellcolor{finalcol}\textbf{HoH@3} & \cellcolor{finalcol}\textbf{72.17} & \cellcolor{finalcol}\textbf{100.00} & \cellcolor{finalcol}\textbf{60.00} & \cellcolor{finalcol}\textbf{81.67} & \cellcolor{finalcol}\textbf{68.33} \\
\addlinespace[2pt]
\rowcolor{simulationrow}\multicolumn{8}{c}{\textbf{Simulation}} \\*
\multirow{4}{0.13\textwidth}{Tycoon} & \multirow{4}{0.19\textwidth}{Wildhaven} & Vanilla & 49.42 & 52.00 & 44.17 & 73.33 & 43.33 \\
 &  & HoH@1 & 74.09 & 91.00 & 72.50 & 85.00 & 63.75 \\
 &  & HoH@2 & 77.78 & 89.00 & 81.67 & 91.67 & 63.12 \\
 &  & \cellcolor{finalcol}\textbf{HoH@3} & \cellcolor{finalcol}\textbf{80.39} & \cellcolor{finalcol}\textbf{90.00} & \cellcolor{finalcol}\textbf{87.50} & \cellcolor{finalcol}\textbf{95.00} & \cellcolor{finalcol}\textbf{62.89} \\
\addlinespace[2pt]
\multirow{4}{0.13\textwidth}{Idle} & \multirow{4}{0.19\textwidth}{Ant Empire} & Vanilla & 65.52 & 93.33 & 68.75 & 35.00 & 63.44 \\
 &  & HoH@1 & 71.42 & 100.00 & 63.75 & 57.50 & 72.81 \\
 &  & HoH@2 & 69.64 & 96.67 & 67.50 & 45.00 & 70.75 \\
 &  & \cellcolor{finalcol}\textbf{HoH@3} & \cellcolor{finalcol}\textbf{87.88} & \cellcolor{finalcol}\textbf{100.00} & \cellcolor{finalcol}\textbf{85.00} & \cellcolor{finalcol}\textbf{92.50} & \cellcolor{finalcol}\textbf{83.57} \\
\addlinespace[2pt]
\multirow{4}{0.13\textwidth}{Simulation} & \multirow{4}{0.19\textwidth}{Kitchen Rush} & Vanilla & 42.62 & 67.00 & 42.00 & 40.00 & 33.93 \\
 &  & HoH@1 & 48.07 & 73.00 & 49.00 & 47.81 & 36.56 \\
 &  & HoH@2 & 65.64 & 90.00 & 68.00 & 65.25 & 53.00 \\
 &  & \cellcolor{finalcol}\textbf{HoH@3} & \cellcolor{finalcol}\textbf{73.38} & \cellcolor{finalcol}\textbf{100.00} & \cellcolor{finalcol}\textbf{72.00} & \cellcolor{finalcol}\textbf{72.92} & \cellcolor{finalcol}\textbf{63.54} \\
\addlinespace[2pt]
\rowcolor{adventurerow}\multicolumn{8}{c}{\textbf{Adventure}} \\*
\multirow{4}{0.13\textwidth}{Horror} & \multirow{4}{0.19\textwidth}{Lighthouse} & Vanilla & 51.95 & 61.67 & 65.00 & 35.00 & 42.00 \\
 &  & HoH@1 & 63.89 & 65.00 & 63.75 & 50.00 & 69.50 \\
 &  & HoH@2 & 71.47 & 78.33 & 75.00 & 57.50 & 71.00 \\
 &  & \cellcolor{finalcol}\textbf{HoH@3} & \cellcolor{finalcol}\textbf{80.90} & \cellcolor{finalcol}\textbf{86.67} & \cellcolor{finalcol}\textbf{86.25} & \cellcolor{finalcol}\textbf{70.00} & \cellcolor{finalcol}\textbf{77.75} \\
\addlinespace[2pt]
\multirow{4}{0.13\textwidth}{Open World} & \multirow{4}{0.19\textwidth}{Airship Trader} & Vanilla & 59.58 & 73.33 & 48.75 & 45.00 & 70.75 \\
 &  & HoH@1 & 73.60 & 90.00 & 67.50 & 95.00 & 63.50 \\
 &  & HoH@2 & 74.40 & 90.00 & 70.00 & 90.00 & 65.42 \\
 &  & \cellcolor{finalcol}\textbf{HoH@3} & \cellcolor{finalcol}\textbf{74.41} & \cellcolor{finalcol}\textbf{95.00} & \cellcolor{finalcol}\textbf{70.00} & \cellcolor{finalcol}\textbf{87.50} & \cellcolor{finalcol}\textbf{64.38} \\
\addlinespace[2pt]
\multirow{4}{0.13\textwidth}{Visual Novel} & \multirow{4}{0.19\textwidth}{Time Paradox} & Vanilla & 50.63 & 77.00 & 60.00 & 40.31 & 34.38 \\
 &  & HoH@1 & 63.28 & 77.00 & 64.00 & 60.62 & 57.81 \\
 &  & HoH@2 & 71.15 & 79.00 & 70.00 & 77.92 & 66.04 \\
 &  & \cellcolor{finalcol}\textbf{HoH@3} & \cellcolor{finalcol}\textbf{71.97} & \cellcolor{finalcol}\textbf{90.00} & \cellcolor{finalcol}\textbf{66.00} & \cellcolor{finalcol}\textbf{86.61} & \cellcolor{finalcol}\textbf{63.93} \\
\addlinespace[2pt]
\bottomrule
\end{longtable}
\endgroup

\end{document}